\pdfoutput=1 % arXiv: force pdfLaTeX (keep within the first 5 lines)
\documentclass{article} % For LaTeX2e
\usepackage{iclr2027_conference,times}

\usepackage{amsmath,amsfonts,bm}

\def\figref#1{figure~\ref{#1}}
\def\secref#1{section~\ref{#1}}
\def\eqref#1{equation~\ref{#1}}
\def\1{\bm{1}}

\DeclareMathAlphabet{\mathsfit}{\encodingdefault}{\sfdefault}{m}{sl}
\SetMathAlphabet{\mathsfit}{bold}{\encodingdefault}{\sfdefault}{bx}{n}

\usepackage[colorlinks=true,linkcolor=red!60!black,citecolor=blue!60!black,urlcolor=magenta!60!black]{hyperref}
\usepackage{url}
\usepackage[normalem]{ulem}  % for \sout

\usepackage{graphicx}
\graphicspath{{figure}, {images}, {example}}
\DeclareGraphicsExtensions{.png,.PNG,.jpeg,.JPEG,.jpg,.JPG,.bmp,.BMP}
\usepackage{epsfig} % for figures
\usepackage{subcaption}
\usepackage{float}
\usepackage{lscape} % Useful for wide tables or figures.
\usepackage{overpic}
\usepackage{mwe} % minimal working examples package
\usepackage{xspace}

\usepackage{algorithm,algpseudocode}
\usepackage{booktabs}  % Publication quality tables
\usepackage[table]{xcolor}
\usepackage{multirow}
\usepackage{paralist}
\usepackage[shortlabels]{enumitem}

\usepackage{mathtools}
\usepackage{amsmath}
\usepackage{amsfonts}
\usepackage{units}
\usepackage{color}
\usepackage{pifont}
\usepackage{comment}
\newlength\paramargin
\newlength\figmargin
\newlength\secmargin
\newlength\figcapmargin

\renewcommand{\secref}[1]{Section~\ref{sec:#1}}
\renewcommand{\figref}[1]{Figure~\ref{fig:#1}} 

\newcommand{\eqnref}[1]{Equation~\ref{eq:#1}}

\long\def\ignorethis#1{}
\definecolor{mygray}{gray}{0.5}

\providecommand{\errstat}[3]{%
    \ensuremath{(#1\pm#2)\times10^{#3}}%
}

\definecolor{m1st}{RGB}{255,153,153}
\definecolor{m2nd}{RGB}{255,204,153}
\definecolor{m3rd}{RGB}{254,248,173}

\newcommand{\cmf}{\cellcolor{m1st}}
\newcommand{\cms}{\cellcolor{m2nd}}

\newcommand{\ffail}[1]{\cellcolor{gray!12}\textcolor{gray!128}{#1}}

\newcommand{\NCLMCT}{NCL-MCT Solver\xspace}

\newcommand{\lossU}{\mathcal{L}^\mathrm{data}_{u}}             % Velocity fitting.
\newcommand{\lossXi}{\mathcal{L}^\mathrm{law}_{\xi}}          % Mobility residual.
\newcommand{\lossF}{\mathcal{L}^\mathrm{law}_{f}}             % Driving-force residual.
\newcommand{\lossR}{\mathcal{L}^\mathrm{law}_{r}}             % Relative reaction-rate residual.
\newcommand{\lossCurl}{\mathcal{L}_{\mathrm{curl}}}
\newcommand{\lossVel}{\mathcal{L}_{\mathrm{vel}}}
\newcommand{\lossMF}{\mathcal{L}_{\mathrm{law}}}

\newcommand{\lossError}{\ell}

\newcommand{\epsLoss}{\varepsilon}
\newcommand{\lambdaCurl}{\lambda_{\mathrm{curl}}}

\title{Neural Constitutive Learning for Generalized Reaction-Diffusion Systems}

\author{
Shang-Ke Chen, Yu-Peng Wang, Shih-Hsuan Hung \& Min-Jhe Lu \\
National Tsing Hua University \\
\AND
Wei-Fang Sun, Chao-Shun Zhan \& Simon See \\
NVIDIA
}

\iclrfinalcopy % de-anonymize: show authors, remove line-number ruler

\newcommand{\PreprintHeader}{Preprint}
\begin{document}

\maketitle
\lhead{\PreprintHeader}

\begin{abstract}
% 1. Problem and research question
Generalized reaction-diffusion systems encompass diverse transport mechanisms and coupled reaction kinetics.
A central question for neural PDE solvers is what should be learned so that a common interface can accommodate phase-field and degenerate transport, local reactions, and multispecies coupling.
%
% 2. Learning target and shared temporal evolution
We propose the Neural Constitutive Laws--Mass-Compression-Transport (NCL-MCT) Solver, which learns PDE-specific constitutive responses while retaining temporal evolution in a shared MCT integrator.
Transport is represented through mobility and thermodynamic driving force, and reaction through relative reaction rates.
%
% 3. Why the representation supports reuse
These constitutive responses depend on the current density rather than explicitly on the initial condition or elapsed time, motivating their reuse across different initial conditions and time horizons.
%
% 4. How the constitutive responses are learned
The same interface supports velocity-data supervision and known-law supervision, neither of which requires time integration during training.
When constitutive laws are known, supervision can be evaluated on independently sampled density fields, enabling trajectory-free constitutive learning without generating solution trajectories.
%
% 5. Cross-system evidence
Across seven systems, separately trained constitutive modules share the same interface and MCT integrator and achieve relative rollout $L^2$ errors of $10^{-4}$ to $10^{-2}$.
%
% 6. Beyond-training evidence and conclusion
Tests with unseen initial-condition families and an extended time horizon assess reuse
beyond training conditions, while separate experiments demonstrate trajectory-free
constitutive learning.
These results support constitutive responses as an effective learning target for a shared neural PDE framework.
\end{abstract}

\section{Introduction}
\label{sec:intro}
Generalized reaction-diffusion partial differential equations (PDEs) describe the evolution of spatial densities through transport, diffusion, and local production or depletion.
They underpin models of biological pattern formation, tumor growth, and intracellular condensation~\citep{turing1952morphogenesis,lu2022vascular,groves2026cellspecific}.
Their dynamics span linear, nonlinear, degenerate, and higher-order diffusion, together with local and coupled reactions~\citep{cahn1958free,vazquez2007porous}.
These processes may also couple with pressure, fluid flow, or mechanical deformation~\citep{lu2019elastic,lu2022vascular}.
Despite this diversity, density balance provides a common evolution structure, while constitutive laws determine the system-specific transport and reaction responses.

Most neural PDE methods learn the solution or its evolution: physics-informed neural networks (PINNs) represent solution fields directly~\citep{raissi2019pinn}, neural operators learn solution maps~\citep{li2021fno,lu2021deeponet}, and other approaches learn temporal derivatives for numerical integration~\citep{hou2026cfo}.
Constitutive laws instead close governing equations by specifying the physical responses that distinguish one system from another.
In solid mechanics, neural constitutive laws have shown that such responses can be learned while retaining the governing evolution~\citep{ma2023nclaw}.
This suggests a complementary learning target for generalized reaction-diffusion systems: learning PDE-specific constitutive responses while retaining the shared evolution structure.
The remaining challenge is to define a common constitutive interface that accommodates phase-field and degenerate transport, local reactions, and multispecies coupling, while supporting both velocity-data supervision and known-law supervision.

We address this problem with the Neural Constitutive Laws--Mass-Compression-Transport (NCL-MCT) Solver, which separates PDE-specific constitutive responses from shared temporal kinematics.
Guided by the energetic variational approach (EnVarA)~\citep{hyon2010envara}, the constitutive interface represents transport through mobility and thermodynamic driving force, and reaction through relative reaction rates.
The constitutive modules learn these responses, while the shared MCT integrator evolves the density through the decomposition $\rho=MI$, separating material mass from compression associated with local volume change.
Each PDE trains its own constitutive modules while sharing the same interface and MCT integrator, under either velocity-data supervision or known-law supervision.
When known constitutive laws are evaluated on independently sampled density fields, the same framework further enables trajectory-free constitutive learning without generating solution trajectories.
For the autonomous systems considered here, the constitutive responses depend on the current density rather than explicitly on the initial condition or elapsed time, motivating their reuse across different initial conditions and time horizons.

We evaluate \NCLMCT on seven systems spanning phase-field and degenerate transport, local reactions, and coupled species.
Across these systems, the shared interface achieves relative rollout $L^2$ errors of $10^{-4}$ to $10^{-2}$ and is competitive with or better than the evaluated baselines on most systems.
Without retraining, the learned constitutive modules achieve roughly $6\times$ lower error on unseen initial-condition families than the strongest finite baseline.
Over an extended time horizon of $1.67\times$ the training horizon, their errors increase by at most $1.5\times$, compared with up to $5\times$ for the evaluated baselines, while remaining an order of magnitude lower in absolute error.
Our contributions are threefold:
\begin{itemize}[leftmargin=*, labelindent=0pt]

\item We formulate constitutive learning for generalized reaction-diffusion systems, representing transport through mobility and thermodynamic driving force and reaction through relative reaction rates, with velocity-data or known-law supervision, including trajectory-free constitutive learning from independently sampled density fields.
\item We develop \NCLMCT, which couples PDE-specific constitutive modules to a shared MCT integrator through a common constitutive interface.
\item We demonstrate cross-system applicability across seven systems and evaluate reuse of the learned constitutive modules, without retraining, on unseen initial-condition families and an extended time horizon.

\end{itemize}
\section{Related Work}
\label{sec:prior}

\subsection{Solution Learning with Physical Constraints}
Physics-informed neural networks (PINNs) represent PDE solutions directly and impose governing equations through residual losses~\citep{raissi2019pinn, karniadakis2021physics}, while neural operators such as FNO~\citep{li2021fno}, DeepONet~\citep{lu2021deeponet}, and CNO~\citep{raonic2023cno} learn mappings between input functions and solution fields.
Physical structure can further enter through PDE residuals, conservation, hard constraints, architectural constraints, or energy-based regularization~\citep{li2021pino,hansen2023probconserv,hao2024stability,liu2024clawno,tanaka2025eno}.
These approaches constrain the solution representation or prediction, whereas constitutive learning places the trainable model in the physical response laws used by an explicit evolution procedure.

\subsection{Time-Derivative Learning with Numerical Integration}
TI-DeepONet~\citep{nayak2026tideeponet} and PITI-DeepONet~\citep{mandl2026piti} learn temporal derivative operators and advance solutions through numerical integration, while CFO~\citep{hou2026cfo} learns the complete PDE right-hand side through flow matching.
These approaches retain explicit time integration but learn the state evolution at the derivative level.
In contrast, constitutive learning separates transport and reaction into PDE-specific physical responses that are supplied to a shared numerical integrator.

\subsection{Component Learning within Numerical Solvers}
Learned numerical solvers may replace selected discretizations or physical components while retaining established PDE update procedures~\citep{barsinai2019discretizations,kochkov2021cfd}.
FINN~\citep{karlbauer2022finn}, PeRCNN~\citep{rao2023percnn}, PAPM~\citep{liu2024papm}, and FluxGNN~\citep{horie2024fluxgnn} learn fluxes, sources, nonlinear terms, or conservative exchanges within structured numerical formulations.
These approaches preserve physical organization through component definitions and numerical assembly.
In contrast, \NCLMCT focuses specifically on PDE-specific constitutive responses and couples them to a shared MCT integrator through a common constitutive interface.

\subsection{Constitutive and Variational Learning}
\label{sec:prior_constitutive}

Constitutive laws close governing equations by specifying system-specific physical
responses, with classical examples in elasticity and
plasticity~\citep{treloar1943elasticity,arruda1993three,drucker1952soil}.
Variational formulations further organize these responses through energy,
dissipation, and kinematics.
EnVarA~\citep{hyon2010envara} provides such a structure by relating energetic
and dissipative forces.
EVNN~\citep{hu2022energetic} and energy-dissipation
approaches~\citep{lu2024generalized,lu2026variational} embed related principles
into learning and numerical evolution, while VONNs~\citep{huang2022vonns}
directly learns free-energy and dissipation potentials from macroscopic observations.
Stat-PINNs~\citep{huang2025statpinns} further learns free energy and dissipative
operators from short-time particle data, addressing non-uniqueness in thermodynamic
identification.
Recent differentiable phase-field models infer free energies and kinetic laws
through PDE evolution~\citep{cohen2026differentiable}.
In mechanics, NCLaw~\citep{ma2023nclaw}, UniPhy~\citep{mittal2025uniphy},
and PC-NCLaws~\citep{xie2025pcnclaws} learn material constitutive responses
within numerical simulators.
DOOL~\citep{chang2025dool} learns dissipative fluxes or change rates by
minimizing an Onsager Rayleighian without solution labels and advances them
through explicit conservation or change laws.
Together, these methods establish physical closures as a learning target, but
differ in the learned quantities and evolution mechanisms.
For generalized reaction-diffusion systems, \NCLMCT uses a common constitutive
interface: mobility and thermodynamic driving force for transport, and relative
reaction rates for reaction.
These PDE-specific responses are coupled to a shared MCT integrator and can be
supervised directly without differentiating through time integration.
\section{Preliminaries}
\label{sec:preliminaries}

Following the energetic variational approach
(EnVarA)~\citep{hyon2010envara,wang2020field}, we organize generalized
reaction-diffusion systems into \textbf{transport and reaction constitutive laws}
coupled to \textbf{shared temporal kinematics}.
For a smooth, positive density $\rho$ on $\Omega\subset\mathbb{R}^d$ with
problem-specific boundary conditions, the density balance is
\begin{equation}
\partial_t\rho+\nabla\cdot(\rho\mathbf{u})=\rho r.
\label{eq:prelim_general_rd}
\end{equation}

\paragraph{Constitutive laws from energetic variations.}
The constitutive laws close \eqnref{prelim_general_rd} by specifying the transport
velocity $\mathbf{u}$ and relative reaction rate $r$.
For a free energy $E[\rho]$ and quadratic dissipation
$\mathcal{D}=\int_{\Omega}\eta|\mathbf{u}|^2\,d\mathbf{x}$, the maximum dissipation
principle gives $\eta\mathbf{u}=-\rho\nabla(\delta E/\delta\rho)$, hence
\begin{equation}
\mathbf{u}=\xi\mathbf{f},\qquad
\xi=\frac{\rho}{\eta}>0,\qquad
\mathbf{f}=-\nabla\frac{\delta E}{\delta\rho},\qquad
r=\frac{S(\rho)}{\rho},
\label{eq:constitutive}
\end{equation}
where $\xi$ is the transport mobility, $\mathbf{f}$ the thermodynamic driving force,
and $S(\rho)$ the reaction source.
System-specific choices are listed in Appendix~\ref{app:pde_specifications}.

Although specifying $E$ and $\eta$ determines both constitutive factors,
$(\lambda\xi,\mathbf{f}/\lambda)$ yields the same $\mathbf{u}$ for any $\lambda>0$.
Thus, velocity data constrain only their product, while separate identification of
mobility and force requires additional constitutive information; related
identifiability issues also arise in thermodynamic learning~\citep{huang2025statpinns}.
The factorization nevertheless provides structure: $\xi>0$, while
$\mathbf{f}=-\nabla(\delta E/\delta\rho)$ is curl-free for smooth fields.
Transport and reaction also enter the balance differently:
$\mathbf{u}$ acts through a divergence, whereas $r$ acts pointwise, motivating
separate constitutive representations and numerical treatment.

\paragraph{Mass-compression-transport kinematics.}
The MCT formulation separates these effects through the decomposition $\rho=MI$:
\begin{equation}
\partial_t I+\nabla\cdot(I\mathbf{u})=0,
\qquad
\partial_t M+\mathbf{u}\cdot\nabla M=Mr,
\label{eq:prelim_im}
\end{equation}
with $I(\mathbf{x},0)=1$ and $M(\mathbf{x},0)=\rho(\mathbf{x},0)$.
For a regular material flow with local volume ratio $J$, $I=J^{-1}$ records
compression or expansion, while $M=\rho J$ represents mass per unit reference
volume.
Along a material trajectory, $M$ changes through reaction according to
$M=\rho^0\exp\!\big(\int_0^t r\,ds\big)$, and the product $\rho=MI$ recovers the
original density balance.
Appendix~\ref{app:kinematic_identity} derives these identities.
The representation assumes $\rho>0$ and a regular material flow; near vanishing
density, the relative reaction rate and factorization require additional care.
\begin{figure}[t]
    \centering
    \begin{overpic}[width=\linewidth]{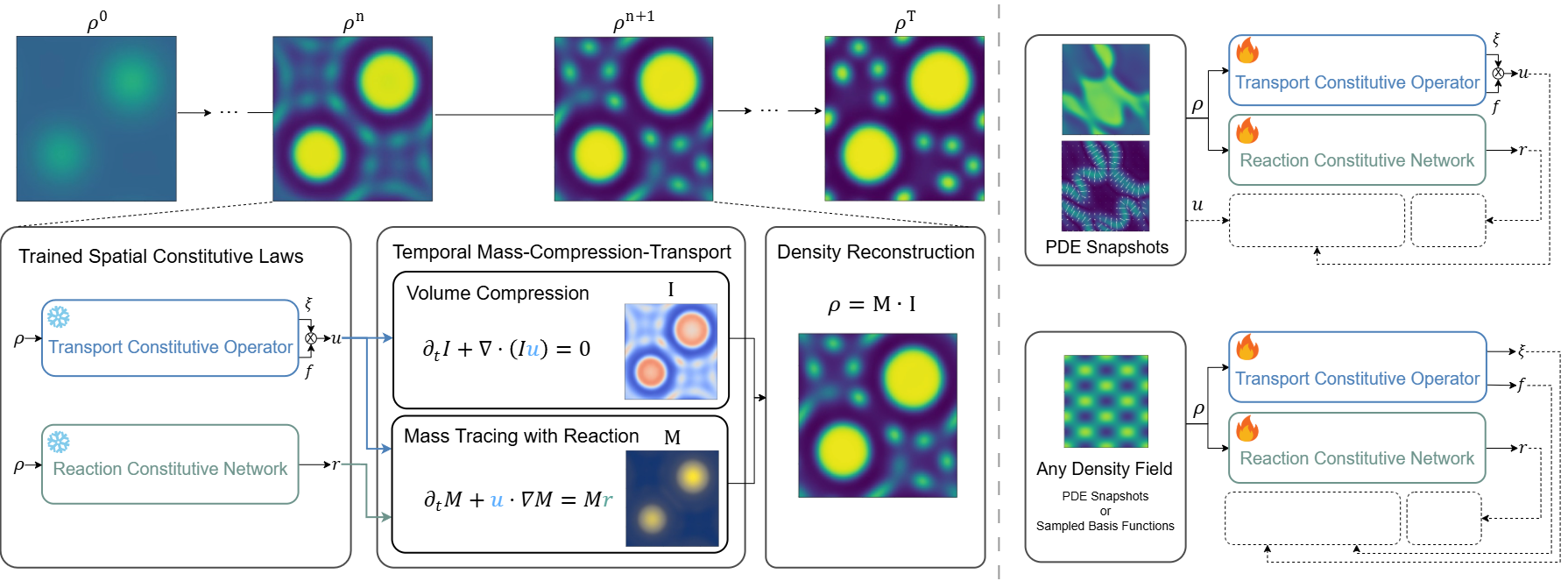}
        \put(27,-1.5){\scriptsize\textbf{(a)} Inference}
        \put(66, 17.5){\scriptsize\textbf{(b)} Learning from velocity data}
        \put(66,-1.5){\scriptsize\textbf{(c)} Learning from known constitutive laws}
        \put(66,30){\tiny{$\rho$}}
        \put(66,25){\tiny{$u$}}
        \put(78.5,22.5){\tiny{$\lossU+\lossCurl$}}
        \put(90.5,22.5){\tiny{$\lossR$}}
        \put(66,11){\tiny{$\rho$}}
        \put(78.5,3.5){\tiny{$\lossXi+\lossF$}}
        \put(90.5,3.5){\tiny{$\lossR$}}
    \end{overpic}
\caption{
Overview of \NCLMCT.
(a) At inference, frozen constitutive modules are coupled to the shared MCT integrator.
At each step, the transport velocity $\mathbf{u}=\xi\mathbf{f}$ and relative reaction rate $r$
are reevaluated from the current density to evolve $I$ and $M$ and reconstruct $\rho=MI$.
During training, transport is supervised either
(b) from reference velocities on PDE snapshots, with curl regularization, or
(c) directly from known mobility and force laws on density fields, including independently sampled density fields.
For reactive systems, $r$ is supervised by the known reaction law in both settings.
}
    \label{fig:neural_pipeline}
\end{figure}
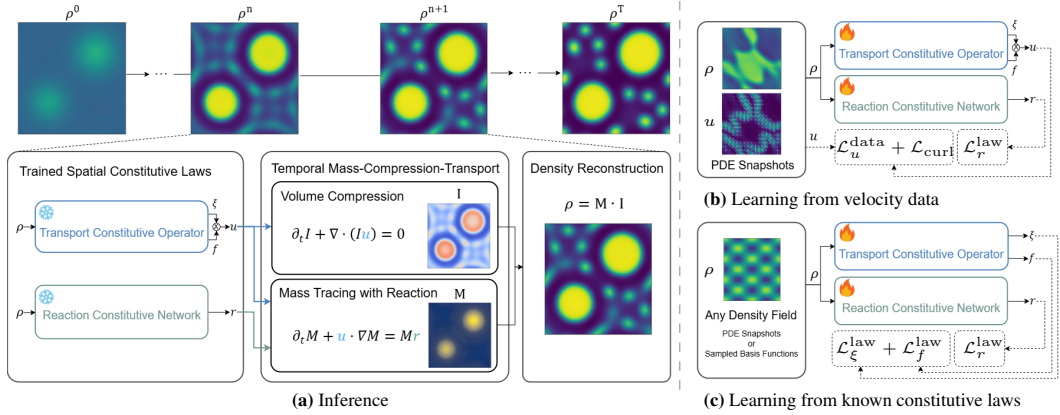

\section{Methodology}
\label{sec:method}

To share one solver interface across diffusive, phase-field, reactive, and multispecies systems, we introduce \NCLMCT, which couples PDE-specific constitutive modules to a shared mass-compression-transport (MCT) integrator.
Following the EnVarA formulation in \secref{preliminaries}, transport is represented through positive mobility and thermodynamic driving force, and reaction through relative reaction rates.
For the autonomous systems considered here, these constitutive responses depend on the current density rather than explicitly on the initial condition or elapsed time.
Figure~\ref{fig:neural_pipeline} summarizes the framework.
A \textbf{Transport Constitutive Operator} learns the mobility-force factorization, while a \textbf{Reaction Constitutive Network} learns relative reaction rates from local species densities (\secref{method_networks}).
At inference, the frozen constitutive modules are reevaluated at each MCT step to advance the compression and mass factors and reconstruct the density (\secref{method_coupling}).
For each PDE, the constitutive modules are trained separately under either velocity-data supervision or known-law supervision, without time integration during training (\secref{constitutive_learning}).

%%%%%%%%%%%%%%%%%%
\subsection{Neural Constitutive Models}
\label{sec:method_networks}

Transport constitutive laws depend on spatial variation of the density field, whereas
the reaction laws considered here are local functions of species densities.
We therefore represent transport with a field-to-field operator and reaction with a
pointwise network.

The \textbf{Transport Constitutive Operator} uses a Convolutional Neural Operator
(CNO)~\citep{raonic2023cno}.
The transport responses in \eqnref{constitutive} involve spatial differential
operators of the current density field, up to third order in the phase-field systems.
Local convolutions capture this differential structure, while the multiscale
architecture spans the spatial scales over which these responses act.
This local representation is also well suited to compactly supported densities
in degenerate transport systems.
%.
Following the EnVarA mobility-force form, we parameterize
\begin{equation}
(\widehat\xi,\widehat{\mathbf f})
=\mathcal C_{\theta_u}[\rho],
\qquad
\widehat{\mathbf u}
=\widehat\xi\,\widehat{\mathbf f},
\label{eq:method_velocity_parameterization}
\end{equation}
where $\rho$ is the density field and $\mathcal C_{\theta_u}$ is the CNO with
trainable parameters $\theta_u$.
A Softplus activation enforces $\widehat\xi>0$ by construction.
Known-law supervision directly constrains the driving force, while under
velocity-data supervision its gradient structure is encouraged by the curl penalty
(\secref{constitutive_learning}).

The \textbf{Reaction Constitutive Network} is a multilayer perceptron (MLP) applied
pointwise:
\begin{equation}
\widehat r=\mathcal R_{\theta_r}[\rho],
\label{eq:method_reaction_module}
\end{equation}
where $\mathcal R_{\theta_r}$ maps the current local species densities to relative reaction rates.
This preserves the locality of the reaction law; predicting the relative reaction
rate rather than the source term also lets reaction enter the mass-factor evolution
multiplicatively through \eqnref{prelim_im}.
Nonreactive systems set $r=0$ and omit this module.
Network architectures are detailed in Appendix~\ref{app:network_architectures}.

%%%%%%%%%%%%%%%%%%
\subsection{Shared MCT Time Integration}
\label{sec:method_coupling}

The MCT integrator contains no trainable update rule.
Given the current density $\rho^n$, the constitutive modules provide the transport velocity and relative reaction rates required by the shared MCT integrator.
Each time step applies a reaction half-step, a transport step, and a second reaction half-step.
The transport velocity is evaluated from $\rho^n$ and held fixed within the step, while the relative reaction rates are reevaluated at each reaction stage.
The compression factor $I$ is advanced with a conservative finite-volume method~\citep{eymard2000finite}, the mass factor $M$ is transported with a semi-Lagrangian method~\citep{staniforth1991semi}, and their product reconstructs $\rho^{n+1}=M^{n+1}I^{n+1}$.
This splitting mirrors the separation of transport and reaction in the continuous formulation, with $I$ tracking local compression and $M$ tracking material mass under transport and reaction.
Because the integrator only queries constitutive responses, numerically evaluated and learned constitutive responses enter through the same interface (Appendix~\ref{app:constitutive_modularity}).
The complete update does not guarantee exact discrete mass conservation of the reconstructed density or unconditional discrete energy dissipation; numerical details and empirical diagnostics are provided in Appendices~\ref{app:numerical_updates} and \ref{app:physical_diagnostics}.

%%%%%%%%%%%%%%%%%%
\subsection{Learning Constitutive Laws}
\label{sec:constitutive_learning}

We train the constitutive modules under two sources of supervision:
\textbf{velocity-data supervision} and \textbf{known-law supervision}.
The supervision source is distinct from the source of input density fields, which may
be trajectory snapshots or independently sampled density fields.
For reactive systems, both modes use known-law supervision for the relative reaction rate.
Neither mode requires time integration during training.

\textbf{Velocity-data supervision.}
When paired density-velocity observations are available, the transport response can
be learned without specifying the underlying mobility and force laws:
\begin{equation}
\lossU
=\lossError(\widehat{\mathbf u},\mathbf u^{\mathrm{data}}),
\qquad
\widehat{\mathbf u}=\widehat\xi\,\widehat{\mathbf f},
\label{eq:method_data_loss}
\end{equation}
where $\mathbf u^{\mathrm{data}}$ is the reference transport velocity and
$\lossError(\widehat{q},q)=B^{-1}\sum_{b}
\lVert\widehat{q}^{(b)}-q^{(b)}\rVert_2^2/
(\lVert q^{(b)}\rVert_2^2+\epsLoss)$
is the per-sample relative squared error over a batch of $B$ density fields, with
$\epsLoss=10^{-12}$.
This normalization prevents low-magnitude constitutive quantities from being
underweighted.

Velocity data constrain only the product
$\widehat{\xi}\widehat{\mathbf f}$; they neither resolve the factorization freedom in
\secref{preliminaries} nor enforce the expected gradient structure of the driving
force.
For smooth fields, the EnVarA force
$\mathbf f=-\nabla(\delta E/\delta\rho)$ is curl-free.
In two dimensions, we encourage this property with
\begin{equation}
\lossCurl
=
\frac{
\left\langle
(\partial_x\widehat f_y-\partial_y\widehat f_x)^2
\right\rangle
}{
\left\langle
(\partial_x\widehat f_y)^2+
(\partial_y\widehat f_x)^2
\right\rangle+\epsLoss
}.
\label{eq:method_curl_loss}
\end{equation}
Here, $\widehat f_x$ and $\widehat f_y$ are the predicted driving-force components,
and $\langle\cdot\rangle$ denotes the arithmetic mean over samples and spatial grid points in a training batch.
The penalty is applied to the driving force because spatially varying mobility may produce a transport velocity with nonzero curl; it does not guarantee an underlying energy functional.

The relative reaction rate is supervised by the known source law:
\begin{equation}
\lossR=\lossError(\widehat r,r^{\mathrm{law}}),
\qquad
r^{\mathrm{law}}=\frac{S(\rho)}{\rho},
\label{eq:method_reaction_loss}
\end{equation}
where $S$ is the prescribed local reaction source.
Because $S$ can be evaluated directly on any input density field, all reported
reaction modules use known-law supervision.
Extracting relative reaction rates from trajectory data would instead require separating reaction from transport.
The complete velocity-data objective is
\begin{equation}
\lossVel
=\lossU+\lambdaCurl\lossCurl+\lossR,
\label{eq:method_training_objectives}
\end{equation}
with $\lambdaCurl=0.01$.
The reaction term is omitted for nonreactive systems, and the curl term is omitted in one dimension.

\textbf{Known-law supervision.}
When the mobility and thermodynamic driving-force laws are known, their evaluations
on the current density provide direct constitutive targets:
\begin{equation}
\lossXi=\lossError(\widehat\xi,\xi^{\mathrm{law}}),
\qquad
\lossF=\lossError(\widehat{\mathbf f},\mathbf f^{\mathrm{law}}).
\label{eq:method_constitutive_losses}
\end{equation}
These losses constrain the two constitutive factors individually, removing the factorization ambiguity inherent in supervising only their product $\widehat{\mathbf u}=\widehat\xi\,\widehat{\mathbf f}$.
The force target also supplies the expected gradient structure.
Using the same reaction loss $\lossR$, the complete known-law objective is
\begin{equation}
\lossMF=\lossXi+\lossF+\lossR.
\label{eq:method_mf_objective}
\end{equation}
With partial constitutive knowledge, known responses can remain analytical while only the unknown constitutive modules are learned.
When known-law supervision is applied to independently sampled density fields, the framework enables \textbf{trajectory-free constitutive learning} without generating PDE solution trajectories.
%%%%%%%%%%%%%%%%%%
\section{Experiments}
\label{sec:result}

We evaluate \NCLMCT along three complementary questions.
First, \emph{cross-system applicability} asks whether a common constitutive interface
covers systems with different transport mechanisms, reactions, and species counts.
Second, \emph{generalization beyond training conditions} evaluates reuse of trained
constitutive modules on unseen initial-condition families and an extended time horizon.
Third, \emph{trajectory-free constitutive learning} tests whether known-law supervision
can train constitutive modules directly on independently sampled density fields without
generating solution trajectories.
PDE-specific constitutive modules are trained separately for each system while sharing
the same constitutive interface and MCT integrator.
Unless otherwise specified, training uses a $128\times128$ grid and ten snapshots from
each of 100 trajectories, yielding $1{,}000$ training samples.
For the trajectory-based benchmarks, we compare against methods with distinct learning
targets:
F-FNO~\citep{tran2023ffno} learns a state map,
CFO~\citep{hou2026cfo} learns a temporal derivative,
and DOOL~\citep{chang2025dool} learns dissipative fluxes through an Onsager variational
objective.
All methods use the same sampled training states where applicable, and the baselines are
trained for $100{,}000$ updates.
We report mean $\pm$ standard deviation over ten test trajectories of the space-time
relative $L^2$ rollout error $E_{\mathrm{roll}}$ and the maximum per-time relative
$L^2$ error $E_{\mathrm{max}}$.
For the extended time horizon, we additionally report
$\Delta E=\bar E(T)-\bar E(T_{\mathrm{train}})$,
where $\bar E(t)$ is the mean per-time relative $L^2$ error across test trajectories.
Vel.\ and Law denote velocity-data and known-law supervision, respectively.
Appendices~\ref{app:pde_specifications}-\ref{app:additional_experiments}
provide experimental details, additional rollouts, generalization results, and ablations.

%%%%%%%%%%%%%%%%%%%%%%%%%%%%%%%%%%%%%%%%%%%%%%%%%%%%%%%%%%%%%%%%%%%%%%%%%%%
\subsection{Applicability across Reaction-Diffusion Systems}
\label{sec:exp_generality}

\begin{table}[t]
    \centering
    \footnotesize
    \setlength{\tabcolsep}{3pt}
    \renewcommand{\arraystretch}{1.15}
\caption{
Applicability across generalized reaction-diffusion systems.
$E_{\mathrm{roll}}$ and $E_{\mathrm{max}}$ denote the space-time and maximum per-time relative $L^2$ errors, reported as mean $\pm$ standard deviation over ten test trajectories; $N_T$ is the rollout length.
Vel./Law denote velocity-data and known-law supervision, respectively.
Red/orange mark the best/second-best mean per row.
Gray$^{*}$ denotes F-FNO results computed from the finite trajectories only; $-$ denotes not evaluated.
}
    \label{tab:reaction_diffusion}

    \resizebox{\linewidth}{!}{%
    \begin{tabular}{@{}lclccccc@{}}
        \toprule
        System & $N_T$ & Metric
        & \shortstack{F-FNO\\\citep{tran2023ffno}}
        & \shortstack{CFO\\\citep{hou2026cfo}}
        & \shortstack{DOOL\\\citep{chang2025dool}}
        & \shortstack{NCL-MCT (Vel.)\\(Ours)}
        & \shortstack{NCL-MCT (Law)\\(Ours)} \\
        \midrule

        \multicolumn{8}{l}{\textbf{(a) Generalized diffusion}} \\
        \midrule

        \multirow{2}{*}{Linear Diffusion}
        & \multirow{2}{*}{$600$}
        & $E_{\mathrm{roll}}$
        & \cmf \errstat{6.52}{1.65}{-4}
        & \errstat{9.90}{4.91}{-3}
        & \errstat{5.65}{1.48}{-2}
        & \cms \errstat{9.87}{2.61}{-4}
        & \errstat{1.57}{0.46}{-3} \\
        & & $E_{\mathrm{max}}$
        & \cmf \errstat{9.76}{2.78}{-4}
        & \errstat{1.19}{0.60}{-2}
        & \errstat{7.36}{1.48}{-2}
        & \cms \errstat{1.18}{0.29}{-3}
        & \errstat{1.82}{0.59}{-3} \\
        \midrule

        \multirow{2}{*}{CH}
        & \multirow{2}{*}{$10{,}000$}
        & $E_{\mathrm{roll}}$
        & \ffail{\errstat{1.45}{0.09}{-1}}
        & \errstat{6.16}{1.47}{-2}
        & \errstat{3.32}{0.53}{-1}
        & \cmf \errstat{7.50}{1.92}{-3}
        & \cms \errstat{7.57}{1.28}{-3} \\
        & & $E_{\mathrm{max}}$
        & \ffail{\errstat{1.87}{0.14}{-1}}
        & \errstat{8.78}{2.35}{-2}
        & \errstat{4.71}{0.74}{-1}
        & \cmf \errstat{1.04}{0.30}{-2}
        & \cms \errstat{1.05}{0.25}{-2} \\
        \midrule

        \multirow{2}{*}{PM}
        & \multirow{2}{*}{$2{,}000$}
        & $E_{\mathrm{roll}}$
        & \errstat{7.87}{4.00}{-2}
        & \cmf \errstat{2.68}{0.61}{-3}
        & \errstat{7.30}{2.61}{-1}
        & \errstat{1.67}{0.17}{-2}
        & \cms \errstat{1.66}{0.25}{-2} \\
        & & $E_{\mathrm{max}}$
        & \errstat{1.46}{0.83}{-1}
        & \cmf \errstat{4.47}{1.19}{-3}
        & \errstat{3.19}{1.03}{0}
        & \cms \errstat{2.09}{0.20}{-2}
        & \errstat{2.20}{0.34}{-2} \\
        \midrule

        \multicolumn{8}{l}{\textbf{(b) Generalized reaction-diffusion}} \\
        \midrule

        \multirow{2}{*}{Fisher-KPP}
        & \multirow{2}{*}{$500$}
        & $E_{\mathrm{roll}}$
        & \cms \errstat{5.35}{1.32}{-4}
        & \errstat{5.90}{2.79}{-3}
        & $-$
        & \cmf \errstat{5.01}{1.39}{-4}
        & \errstat{6.58}{0.54}{-4} \\
        & & $E_{\mathrm{max}}$
        & \cms \errstat{7.50}{1.67}{-4}
        & \errstat{7.83}{3.73}{-3}
        & $-$
        & \cmf \errstat{5.96}{1.96}{-4}
        & \errstat{8.90}{0.79}{-4} \\
        \midrule

        \multirow{2}{*}{Reactive CH}
        & \multirow{2}{*}{$10{,}000$}
        & $E_{\mathrm{roll}}$
        & \ffail{\errstat{1.29}{0.23}{-1}}
        & \errstat{5.64}{1.11}{-2}
        & $-$
        & \cmf \errstat{8.86}{2.21}{-3}
        & \cms \errstat{9.62}{2.13}{-3} \\
        & & $E_{\mathrm{max}}$
        & \ffail{\errstat{1.63}{0.34}{-1}}
        & \errstat{7.95}{2.06}{-2}
        & $-$
        & \cms \errstat{1.40}{0.42}{-2}
        & \cmf \errstat{1.37}{0.38}{-2} \\
        \midrule

        \multirow{2}{*}{Reactive PM}
        & \multirow{2}{*}{$1{,}000$}
        & $E_{\mathrm{roll}}$
        & \errstat{3.96}{3.77}{-2}
        & \cmf \errstat{7.78}{3.98}{-3}
        & $-$
        & \errstat{3.84}{0.79}{-2}
        & \cms \errstat{1.56}{0.57}{-2} \\
        & & $E_{\mathrm{max}}$
        & \errstat{4.81}{4.01}{-2}
        & \cmf \errstat{1.52}{1.09}{-2}
        & $-$
        & \errstat{5.32}{1.16}{-2}
        & \cms \errstat{1.89}{0.55}{-2} \\
        \midrule

        \multicolumn{8}{l}{\textbf{(c) Multispecies reaction-diffusion}} \\
        \midrule

        \multirow{2}{*}{Schnakenberg ($U$)}
        & \multirow{4}{*}{$20{,}000$}
        & $E_{\mathrm{roll}}$
        & \ffail{\errstat{2.73}{0.36}{-1}}
        & \errstat{1.28}{0.16}{-1}
        & $-$
        & \cms \errstat{2.52}{1.77}{-2}
        & \cmf \errstat{2.38}{1.26}{-2} \\
        & & $E_{\mathrm{max}}$
        & \ffail{\errstat{7.06}{0.63}{-1}}
        & \errstat{2.87}{0.49}{-1}
        & $-$
        & \cms \errstat{8.28}{8.55}{-2}
        & \cmf \errstat{7.36}{6.15}{-2} \\
        \addlinespace[3pt]

        \multirow{2}{*}{Schnakenberg ($V$)}
        & & $E_{\mathrm{roll}}$
        & \ffail{\errstat{1.18}{0.18}{-1}}
        & \errstat{4.39}{0.59}{-2}
        & $-$
        & \cms \errstat{6.74}{3.37}{-3}
        & \cmf \errstat{6.57}{2.27}{-3} \\
        & & $E_{\mathrm{max}}$
        & \ffail{\errstat{2.33}{0.26}{-1}}
        & \errstat{1.04}{0.18}{-1}
        & $-$
        & \cms \errstat{2.19}{1.93}{-2}
        & \cmf \errstat{1.97}{1.30}{-2} \\
        \bottomrule
        \multicolumn{8}{l}{%
            \footnotesize $^{*}$Finite trajectory counts for gray cells:
            CH 9/10,\ Reactive CH 8/10,\ Schnakenberg 9/10.}
    \end{tabular}
    }
\end{table}
% Spatial-panel order and final time were checked against
% figures/images/fig3(s).png; the source calls Law "Ours" and Vel. "Ours (data)".
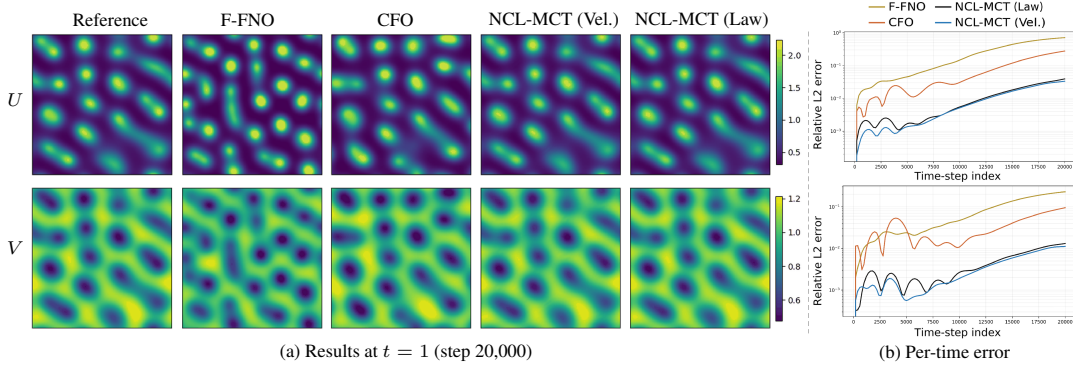
\begin{figure}[t]
	\centering
	\begin{overpic}[width=\linewidth]{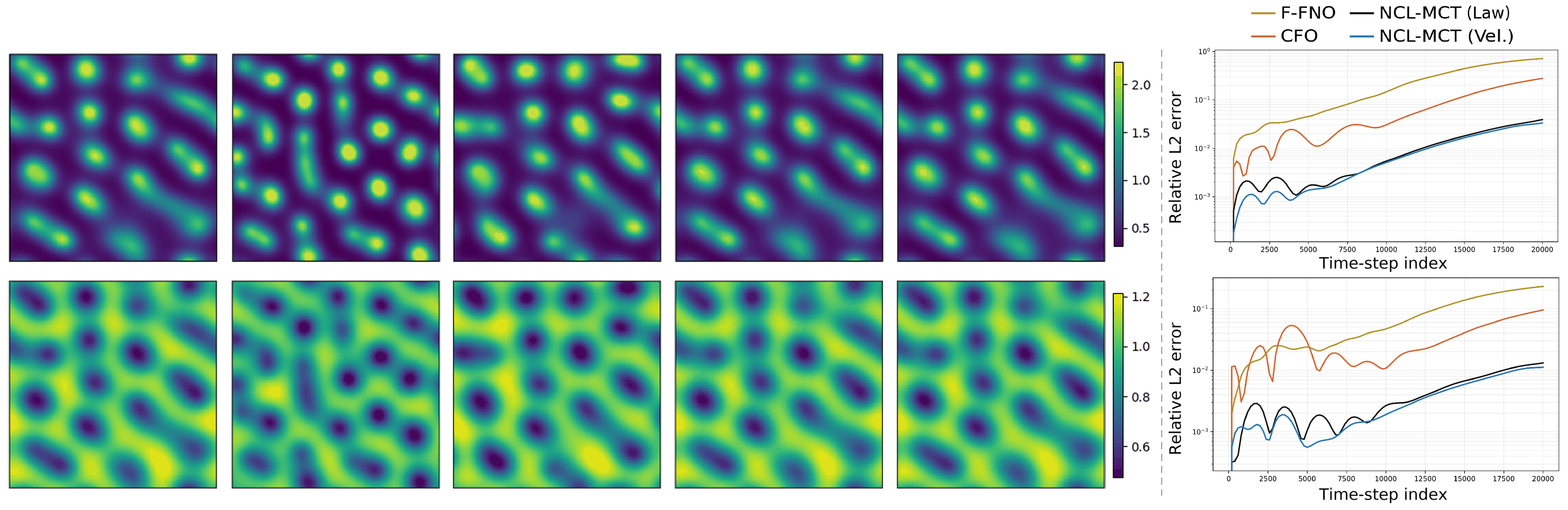}
		\put(-1,23){\makebox(0,0){\scriptsize $U$}}
		\put(-1,9){\makebox(0,0){\scriptsize $V$}}
		\put( 8,30.5){\makebox(0,0){\scriptsize Reference}}
		\put(21,30.5){\makebox(0,0){\scriptsize F-FNO}}
		\put(35,30.5){\makebox(0,0){\scriptsize CFO}}
		\put(50,30.5){\makebox(0,0){\scriptsize NCL-MCT (Vel.)}}
		\put(64,30.5){\makebox(0,0){\scriptsize NCL-MCT (Law)}}
		\put(36,-1){\makebox(0,0){\scriptsize (a) Results at $t=1$ (step 20{,}000)}}
		\put(87,-1){\makebox(0,0){\scriptsize (b) Per-time error}}
	\end{overpic}
\caption{
Schnakenberg comparison for both species.
(a) Concentrations $U$ (top) and $V$ (bottom) at $t=1$ for a test trajectory on which all methods remain finite.
(b) Per-time relative $L^2$ errors over the $20{,}000$-step rollout.
}
	\label{fig:schnakenberg_temporal}
\end{figure}

\paragraph{Generalized diffusion.}
\label{sec:exp_generalized_diffusion}
We first consider linear diffusion, Cahn-Hilliard (CH), and porous medium (PM),
representing spatial smoothing, phase separation, and degenerate diffusion with moving
low-density fronts.
The same constitutive interface and MCT integrator are used across all three systems.
On CH, both \NCLMCT configurations achieve roughly $8\times$ lower rollout errors than
the strongest evaluated baseline (Table~\ref{tab:reaction_diffusion}(a)).
On linear diffusion, F-FNO is more accurate; HC-PINN~\citep{hao2024stability} shows a similar
advantage (Appendix~\ref{app:hc_pinn_comparison}).
On PM, CFO is approximately five to six times more accurate than both \NCLMCT
configurations, which remain five to seven times more accurate than F-FNO.
PM also exposes the positive-density limitation in \secref{preliminaries}: compact
support creates vanishing-density regions where the factorization $\rho=MI$ requires
additional care near the free boundary.

\paragraph{Generalized reaction-diffusion.}
\label{sec:exp_scalar_reaction_diffusion}
Reaction is incorporated by adding a pointwise relative-reaction-rate module to the same
transport interface and MCT integrator.
The DOOL formulation evaluated here is transport-only and is therefore omitted from
these reactive benchmarks.
On Fisher-KPP, reactive CH, and reactive PM
(Table~\ref{tab:reaction_diffusion}(b)), both \NCLMCT configurations achieve roughly
$6\times$ lower errors on reactive CH than the strongest evaluated baseline.
Vel.\ attains the lowest mean error on Fisher-KPP, with F-FNO within one standard
deviation, whereas CFO is more accurate on reactive PM.
The PM ranking follows the nonreactive case, suggesting that degenerate transport remains
the dominant difficulty.
Known-law supervision reduces the reactive-PM error by roughly $2.5$-$3\times$ relative
to Vel.; the two supervision modes are otherwise comparable.

\paragraph{Multispecies reaction-diffusion.}
\label{sec:exp_multispecies_coupling}
For two species, each species uses its own transport module, while the reaction module
takes both local concentrations as input and is trained jointly with the transport
modules.
On Schnakenberg, both \NCLMCT configurations remain stable over the full
$20{,}000$-step rollout, with mean errors three to seven times below CFO and about an
order of magnitude below F-FNO across both species and metrics
(Table~\ref{tab:reaction_diffusion}(c)).
F-FNO becomes nonfinite on one trajectory near the end of the rollout, while Law has
the lower mean error for both species.
\figref{schnakenberg_temporal} shows the error histories: F-FNO errors grow steadily,
CFO accumulates larger final-state discrepancies, and both \NCLMCT configurations remain
close to the reference.

%%%%%%%%%%%%%%%%%%%%%%%%%%%%%%%%%%%%%%%%%%%%%%%%%%%%%%%%%%%%%%%%%%%%%%%%%%%
\begin{figure}[t]
\centering
\begin{overpic}[width=\linewidth]{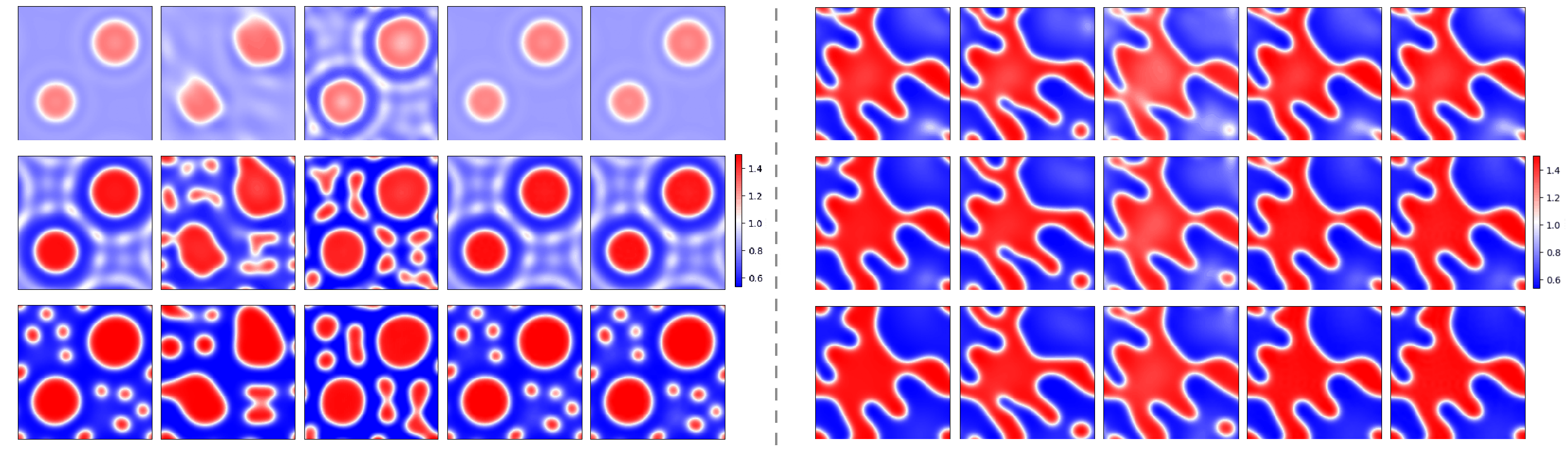}
 \put(5.75,29.2){\makebox(0,0){\tiny Reference}}
 \put(14 ,29.2){\makebox(0,0){\tiny F-FNO}}
 \put(23 ,29.2){\makebox(0,0){\tiny CFO}}
 \put(32 ,29.2){\makebox(0,0){\tiny NCL-MCT(Vel.)}}
 \put(43 ,29.2){\makebox(0,0){\tiny NCL-MCT(Law)}}
 \put(56.75,29.2){\makebox(0,0){\tiny Reference}}
 \put(65  ,29.2){\makebox(0,0){\tiny F-FNO}}
 \put(74  ,29.2){\makebox(0,0){\tiny CFO}}
 \put(83.5,29.2){\makebox(0,0){\tiny NCL-MCT(Vel.)}}
 \put(94.5,29.2){\makebox(0,0){\tiny NCL-MCT(Law)}}
 \put(0,24){\makebox(0,0){\rotatebox{90}{\tiny $t=0.02$}}}
 \put(0,14.5){\makebox(0,0){\rotatebox{90}{\tiny $t=0.05$}}}
 \put(0, 5){\makebox(0,0){\rotatebox{90}{\tiny $t=0.10$}}}
 \put(51,24){\makebox(0,0){\rotatebox{90}{\tiny $t=0.06$}}}
 \put(51,14.5){\makebox(0,0){\rotatebox{90}{\tiny $t=0.08$}}}
 \put(51, 5){\makebox(0,0){\rotatebox{90}{\tiny $t=0.10$}}}
 \put(23,-1.2){\makebox(0,0){\scriptsize (a) Unseen initial conditions}}
 \put(75,-1.2){\makebox(0,0){\scriptsize (b) Extended time horizon}}
\end{overpic}
\caption{
Generalization beyond training conditions on reactive Cahn--Hilliard.
(a) From an unseen two-nucleus initial condition, F-FNO distorts the large domains and
loses secondary droplets, while CFO merges them into elongated structures; both
\NCLMCT configurations better preserve the reference morphology.
(b) Models trained on the first $60\%$ of the interval are rolled out over the full
horizon. Beyond the training window, CFO loses interior contrast and smooths narrow
channels, while both \NCLMCT configurations better preserve the interfaces.
}
\label{fig:ood}
\end{figure}
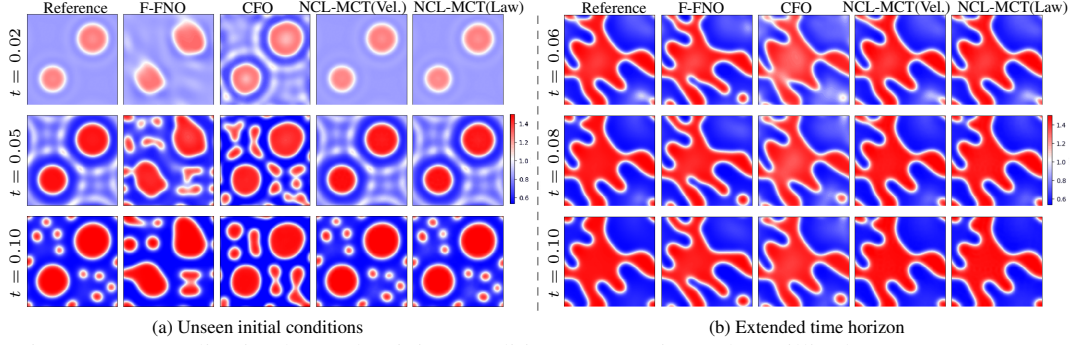
\begin{table}[t]
    \centering
    \footnotesize
    \setlength{\tabcolsep}{6pt}
    \renewcommand{\arraystretch}{1.15}
    \caption{
        Generalization beyond training conditions, aggregated across systems:
        (a) unseen initial conditions (five systems, one test trajectory per system) and
        (b) extended time horizon (four systems, ten test trajectories per system).
        Entries are averages of the per-system $E_{\mathrm{roll}}$, $E_{\mathrm{max}}$, or
        $\Delta E$ values.
        Per-system results are reported in Appendix~\ref{app:generalization}.
        DOOL is omitted because it covers too few systems for aggregation.
        Red/orange indicate the best/second-best comparable result per row.
    }
    \label{tab:ood}

    \resizebox{0.95\linewidth}{!}{%
    \begin{tabular}{@{}llcccc@{}}
        \toprule
         & Metric
        & \shortstack{F-FNO\\\citep{tran2023ffno}}
        & \shortstack{CFO\\\citep{hou2026cfo}}
        & \shortstack{NCL-MCT (Vel.)\\(Ours)}
        & \shortstack{NCL-MCT (Law)\\(Ours)} \\
        \midrule

        Unseen initial conditions
        & $E_{\mathrm{roll}}$
        & \ffail{$1.81\times10^{-1}$$^{\ast}$}
        & $1.69\times10^{-1}$
        & \cms $2.99\times10^{-2}$
        & \cmf $2.86\times10^{-2}$ \\
        (5 systems)
        & $E_{\mathrm{max}}$
        & \ffail{$2.97\times10^{-1}$$^{\ast}$}
        & $2.38\times10^{-1}$
        & \cmf $4.39\times10^{-2}$
        & \cms $4.44\times10^{-2}$ \\
        \midrule

        Extended time horizon
        & $E_{\mathrm{roll}}$
        & \ffail{$6.75\times10^{-2}$$^{\dagger}$}
        & $6.89\times10^{-2}$
        & \cms $5.85\times10^{-3}$
        & \cmf $5.49\times10^{-3}$ \\
        (4 systems)
        & $\Delta E$
        & \ffail{$1.25\times10^{-2}$$^{\dagger}$}
        & $4.82\times10^{-3}$
        & \cmf $1.97\times10^{-3}$
        & \cms $2.76\times10^{-3}$ \\
        \bottomrule
        \addlinespace[2pt]
        \multicolumn{6}{l}{\footnotesize
        $^{*}$F-FNO is nonfinite on CH and Fisher-KPP; the reported average uses only the 3/5 finite systems and is not directly comparable.}\\
        \multicolumn{6}{l}{\footnotesize
        $^{\dagger}$For reactive CH, F-FNO statistics use the 9/10 trajectories that remain finite.}
    \end{tabular}%
    }
\end{table}

\subsection{Generalization beyond Training Conditions}
\label{sec:exp_generalization}

\paragraph{Unseen initial conditions.}
Because the constitutive modules are evaluated from the current density rather than
explicitly from the initial condition, changing the initial-condition family changes
the density fields encountered during rollout.
Keeping the trained models and PDE parameters of \secref{exp_generality}, we replace only
the initial-condition family with unseen structured fields whose amplitudes remain close
to the training range, making the shift primarily spatial
(Appendix~\ref{app:unseen_ic}).
Across the five systems, both \NCLMCT configurations achieve nearly $6\times$ lower
average $E_{\mathrm{roll}}$ than CFO, the strongest baseline that remains finite across
all five systems, with per-system margins from $2.4\times$ to $13\times$
(Table~\ref{tab:ood}(a)).
\figref{ood}(a) illustrates the difference on reactive CH.
Linear diffusion shows the largest change in ranking:
F-FNO is the most accurate method in distribution, but its error increases by roughly
$260\times$ under the shift, compared with $24$-$41\times$ for \NCLMCT, and it becomes
nonfinite on CH and Fisher-KPP.
The relative degradation is not uniformly smaller for \NCLMCT; on Fisher-KPP, its error
also increases substantially but remains lower because its in-distribution error starts
from a much smaller level.

\paragraph{Extended time horizon.}
We train on the first $60\%$ of each reference trajectory and evaluate over the full
interval, so the final $40\%$ lies beyond the training horizon
(Appendix~\ref{app:extended_horizon}).
Averaged across the four systems, both \NCLMCT configurations achieve full-window
rollout errors about an order of magnitude below the evaluated baselines.
Their errors change by only $0.90$-$1.46\times$ relative to the corresponding
models trained on the full time interval (Table~\ref{tab:ood}(b)).
By comparison, CFO errors increase by about $1.6\times$ on the two Cahn-Hilliard
systems and $5.2\times$ on linear diffusion and Fisher-KPP.
F-FNO remains strongest on linear diffusion but about an order of magnitude less accurate
than \NCLMCT on the two Cahn-Hilliard systems.
Both \NCLMCT configurations also show the smallest $\Delta E$ on linear diffusion, CH,
and Fisher-KPP.
Only on reactive CH does CFO show smaller growth, despite an error at
$T_{\mathrm{train}}$ that is twelve times larger.
\figref{ood}(b) shows the corresponding reactive-CH fields.

%%%%%%%%%%%%%%%%%%%%%%%%%%%%%%%%%%%%%%%%%%%%%%%%%%%%%%%%%%%%%%%%%%%%%%%%%%%
\subsection{Trajectory-Free Constitutive Learning}
\label{sec:exp_trajectory_free}
% figures/table_func.tex -- referenced from Section 5.3 as \input{figures/table_func}
\begin{table}[t]
  \centering
  \footnotesize
  \setlength{\tabcolsep}{4pt}
  \renewcommand{\arraystretch}{1.15}
  \caption{
Trajectory-free learning from independently sampled density fields.
Both methods use the same unlabeled density fields and are evaluated on ten held-out
initial conditions.
$E_{\mathrm{roll}}$ and $E_{\mathrm{max}}$ are relative $L^2$ errors
(mean $\pm$ standard deviation), and $N_T$ is the rollout length.
Law denotes known-law supervision on independently sampled states.
Red marks the lower mean per row.
}
  \label{tab:trajectory_free}

  \resizebox{0.7\linewidth}{!}{%
  \begin{tabular}{@{}llccc@{}}
    \toprule
    System & $N_T$ & Metric
    & \shortstack{DOOL\\\citep{chang2025dool}}
    & \shortstack{NCL-MCT (Law)\\(Ours)} \\
    \midrule

      Linear Diffusion
      & \multirow{2}{*}{$4{,}000$}
      & $E_{\mathrm{roll}}$
      & \errstat{1.53}{1.07}{-3}
      & \cmf \errstat{3.13}{1.69}{-4} \\
      \scriptsize{(1D, DOOL setting)} &  
      & $E_{\mathrm{max}}$
      & \errstat{2.39}{1.67}{-3}
      & \cmf \errstat{4.05}{2.32}{-4} \\
    \midrule

      CH
      & \multirow{2}{*}{$2{,}000$}
      & $E_{\mathrm{roll}}$
      & \errstat{7.46}{2.90}{-3}
      & \cmf \errstat{6.40}{6.66}{-3} \\
      \scriptsize{(DOOL setting)} & 
      & $E_{\mathrm{max}}$
      & \errstat{1.17}{0.37}{-2}
      & \cmf \errstat{8.71}{8.33}{-3} \\
    \bottomrule
  \end{tabular}%
  }
\end{table}

Although the experiments above use trajectory snapshots as training inputs, known-law
supervision does not require them: the targets
$\xi^{\mathrm{law}}$ and $\mathbf f^{\mathrm{law}}$ in
\eqnref{method_constitutive_losses} depend only on the current density field and can be
evaluated on independently sampled density fields.
We test this trajectory-free setting on one-dimensional linear diffusion and
two-dimensional Cahn-Hilliard using the sampled training distributions of
DOOL~\citep{chang2025dool}.
For each system, DOOL and \NCLMCT (Law) use the same sampled density fields; solution
trajectories are reserved for held-out rollout evaluation
(Appendix~\ref{app:trajectory_free}).
\NCLMCT (Law) achieves lower mean errors on both systems
(Table~\ref{tab:trajectory_free}).
On linear diffusion, $E_{\mathrm{roll}}$ and $E_{\mathrm{max}}$ are lower than those of
DOOL by factors of $4.9$ and $5.9$, respectively.
On Cahn-Hilliard, the corresponding factors are $1.2$ and $1.4$, although the two
methods remain within one standard deviation.
These results show that, on these benchmarks, constitutive modules trained directly on
independently sampled density fields can support $2{,}000$-$4{,}000$-step rollouts
without solution trajectories during training.
Because the methods use different time integrators, the quantitative comparison reflects
the complete solvers rather than the learning objectives alone.
\section{Conclusion}
\label{sec:conclusion}

We presented \NCLMCT, which places neural learning in PDE-specific constitutive responses while retaining a shared MCT integrator for density evolution.
Transport is represented through mobility and thermodynamic driving force, and reaction through relative reaction rates.
Across seven generalized diffusion and reaction-diffusion systems, the shared constitutive interface achieves relative rollout $L^2$ errors of $10^{-4}$ to $10^{-2}$ and is competitive with or better than the evaluated baselines on most systems.
Tests on unseen initial-condition families and an extended time horizon further assess reuse of the learned constitutive modules without retraining.
With known-law supervision on independently sampled density fields, a separate experiment demonstrates trajectory-free constitutive learning without generating solution trajectories.
Together, these results support learning PDE-specific constitutive responses, rather than the full solution evolution, as an effective target for a shared neural PDE framework.

The current formulation assumes positive densities, with porous-medium systems exposing its limitation near vanishing density.
Future work will extend the factor evolution to such regimes, investigate constitutive identification under partially known laws, broaden the treatment of boundary conditions~\citep{liu2024opno}, incorporate parameter-conditioned constitutive models for inverse problems~\citep{cho2025pidion}, and couple the framework to multiphysics models involving pressure, flow, and tissue mechanics~\citep{lu2019elastic,lu2022vascular}.

\subsection*{Disclosure of AI Use}
The authors developed the research ideas and proposed method, performed the mathematical verification, and designed and conducted all experiments. AI tools assisted with identifying relevant literature, exploring existing neural network architectures and numerical methods, implementing and debugging code, drafting portions of the manuscript, and improving its wording and clarity. All code was reviewed and validated by the authors, who take full responsibility for the content and conclusions of this work.

% \section*{Acknowledgments}
% Funding and acknowledgements go here (not allowed in the anonymous version).

\bibliography{iclr2027_conference}
\bibliographystyle{iclr2027_conference}

\clearpage
\appendix
\section*{Appendix Overview}

\begin{itemize}[leftmargin=*, labelindent=0pt]
\item \textbf{\ref{app:kinematic_identity}. \nameref{app:kinematic_identity}}
  \begin{itemize}
  \item \ref{app:compression_volume}~\nameref{app:compression_volume}
  \item \ref{app:mass_tracing}~\nameref{app:mass_tracing}
  \item \ref{app:balance_equivalence}~\nameref{app:balance_equivalence}
  \end{itemize}

\item \textbf{\ref{app:network_architectures}. \nameref{app:network_architectures}}
  \begin{itemize}
  \item \ref{app:velocity_network}~\nameref{app:velocity_network}
  \item \ref{app:reaction_network}~\nameref{app:reaction_network}
  \end{itemize}

\item \textbf{\ref{app:numerical_updates}. \nameref{app:numerical_updates}}

\item \textbf{\ref{app:pde_specifications}. \nameref{app:pde_specifications}}

\item \textbf{\ref{app:data_and_training}. \nameref{app:data_and_training}}
  \begin{itemize}
  \item \ref{app:ic_families}~\nameref{app:ic_families}
  \item \ref{app:snapshots}~\nameref{app:snapshots}
  \item \ref{app:training_settings}~\nameref{app:training_settings}
  \end{itemize}

\item \textbf{\ref{app:rollout_results}. \nameref{app:rollout_results}}

\item \textbf{\ref{app:generalization}. \nameref{app:generalization}}
\begin{itemize}
  \item \ref{app:unseen_ic}~\nameref{app:unseen_ic}
  \item \ref{app:extended_horizon}~\nameref{app:extended_horizon}
\end{itemize}

\item \textbf{\ref{app:trajectory_free}. \nameref{app:trajectory_free}}

\item \textbf{\ref{app:additional_experiments}. \nameref{app:additional_experiments}}
  \begin{itemize}
  \item \ref{app:timing}~\nameref{app:timing}
  \item \ref{app:physical_diagnostics}~\nameref{app:physical_diagnostics}
  \item \ref{app:hc_pinn_comparison}~\nameref{app:hc_pinn_comparison}
  \item \ref{app:exp_output_ablation}~\nameref{app:exp_output_ablation}
  \item \ref{app:constitutive_modularity}~\nameref{app:constitutive_modularity}
  \item \ref{app:lambda}~\nameref{app:lambda}
  \end{itemize}

%\item \textbf{\ref{app:ai_disclosure}. \nameref{app:ai_disclosure}}
\end{itemize}

\clearpage
\section{Derivation of the Mass-Compression Formulation}
\label{app:kinematic_identity}

We verify the continuous identities used in Section~\ref{sec:preliminaries} for a single species.
Assume smooth fields, $\rho>0$, $I>0$, $M>0$, and a differentiable, invertible material flow.
Let $D_t=\partial_t+\mathbf{u}\cdot\nabla$ and
$\rho^0(\mathbf{x})=\rho(\mathbf{x},0)$.
The derivation holds on material regions where the flow is defined and does not require periodic boundaries.
At inflow boundaries, incoming factor data must be consistent with the prescribed density.

\subsection{Compression-Volume Relation}
\label{app:compression_volume}

Let $\boldsymbol{\Phi}(\mathbf{a},t)$ denote the material flow generated by $\mathbf{u}$,
with $\boldsymbol{\Phi}(\mathbf{a},0)=\mathbf{a}$, and let
\[
J(\mathbf{a},t)
=\det\nabla_{\mathbf{a}}\boldsymbol{\Phi}(\mathbf{a},t)>0
\]
be its local volume ratio.
Differentiating the flow with respect to $\mathbf{a}$ and applying Jacobi's determinant formula gives
\begin{equation}
    \frac{dJ}{dt}
    =J(\nabla\cdot\mathbf{u})(\boldsymbol{\Phi}(\mathbf{a},t),t).
\end{equation}
Meanwhile, \eqnref{prelim_im} implies
$D_t I=-I\nabla\cdot\mathbf{u}$.
Therefore,
\begin{equation}
    \frac{d}{dt}
    \bigl[I(\boldsymbol{\Phi}(\mathbf{a},t),t)J(\mathbf{a},t)\bigr]=0,
    \qquad
    I(\boldsymbol{\Phi}(\mathbf{a},t),t)=J(\mathbf{a},t)^{-1},
\end{equation}
where the second identity uses $I(\mathbf{x},0)=1$ and $J(\mathbf{a},0)=1$.
Thus, $I=J^{-1}$ measures local compression: contraction decreases $J$ and increases $I$, while expansion has the opposite effect.

\subsection{Mass Tracing with Reaction}
\label{app:mass_tracing}

Since $\rho=MI$ and $IJ=1$, a material volume element satisfies
\begin{equation}
    \rho(\boldsymbol{\Phi}(\mathbf{a},t),t)
    J(\mathbf{a},t)\,d\mathbf{a}
    =
    M(\boldsymbol{\Phi}(\mathbf{a},t),t)\,d\mathbf{a}.
\end{equation}
Hence, $M=\rho J$ represents mass per unit reference volume.
Its material equation $D_tM=Mr$ integrates to
\begin{equation}
    M(\boldsymbol{\Phi}(\mathbf{a},t),t)
    =
    \rho^0(\mathbf{a})
    \exp\!\left(
        \int_0^t r(\boldsymbol{\Phi}(\mathbf{a},s),s)\,ds
    \right).
\end{equation}
This identity remains valid when $r$ depends on $\rho$, since the integral is evaluated along the evolving material trajectory.
For any reference region $A$ whose image remains inside $\Omega$,
\begin{equation}
    \int_{\boldsymbol{\Phi}(A,t)}
    \rho(\mathbf{x},t)\,d\mathbf{x}
    =
    \int_A
    M(\boldsymbol{\Phi}(\mathbf{a},t),t)\,d\mathbf{a}.
\end{equation}
When $r=0$, $M$ is constant along material trajectories; otherwise, it records reaction-induced mass production or loss.

\subsection{Equivalence to the Density Balance}
\label{app:balance_equivalence}

For $\rho=MI$, the product rule and \eqnref{prelim_im} give
\begin{equation}
    \begin{aligned}
        \partial_t\rho+\nabla\cdot(\rho\mathbf{u})
        &=
        M\bigl[\partial_t I+\nabla\cdot(I\mathbf{u})\bigr]
        +I\bigl[\partial_t M+\mathbf{u}\cdot\nabla M\bigr] \\
        &=IMr=\rho r.
    \end{aligned}
    \label{eq:appendix_product_identity}
\end{equation}
The initialization $I(\mathbf{x},0)=1$ and
$M(\mathbf{x},0)=\rho^0(\mathbf{x})$ recovers the prescribed initial density.

Conversely, let a positive $\rho$ satisfy \eqnref{prelim_general_rd}.
Construct $M$ from its material equation with the same velocity and relative reaction rate,
with $M(\mathbf{x},0)=\rho^0(\mathbf{x})$, and set $I=\rho/M$.
Since
\[
D_t\rho=\rho(r-\nabla\cdot\mathbf{u}),
\]
we obtain
\begin{equation}
    D_t I
    =
    \frac{D_t\rho}{M}
    -\frac{\rho}{M^2}D_tM
    =
    -I\nabla\cdot\mathbf{u},
\end{equation}
which is equivalent to the conservative equation for $I$ in \eqnref{prelim_im}.
Boundary data must likewise satisfy $\rho=MI$ together with the boundary conditions of the original problem.

These identities concern the continuous formulation and do not imply discrete guarantees for accuracy, conservation, positivity, or energy dissipation.
\clearpage
\section{Network Architectures}
\label{app:network_architectures}

This appendix details the constitutive modules introduced in
Section~\ref{sec:method_networks}.
Both modules are trained without differentiating through the MCT integrator; training
settings are given in Appendix~\ref{app:training_settings}.
The implementation is built on PhysicsNeMo~\citep{PhysicsNeMo_Contributors_NVIDIA_PhysicsNeMo_An_2023},
and the code will be made publicly available.

\subsection{Transport Constitutive Operator}
\label{app:velocity_network}

The transport module uses a Convolutional Neural Operator
(CNO)~\citep{raonic2023cno} with the density field as input.
Its three output channels represent a scalar mobility $\widehat\xi$ and the two
components of the driving force
$\widehat{\mathbf f}=(\widehat f_x,\widehat f_y)$.
Softplus enforces $\widehat\xi>0$, while the force components use unrestricted linear
outputs; their product gives the transport velocity
$\widehat{\mathbf u}=\widehat\xi\,\widehat{\mathbf f}$.
The direct-velocity ablation instead outputs two unrestricted velocity components.

All two-dimensional configurations use two downsampling levels with encoder widths
$(8,16,32)$, one residual block per encoder level, and two residual blocks at the
bottleneck.
The decoder uses concatenated skip connections, with hidden width 32 for lifting and
projection.
Convolutions use $3\times3$ kernels with circular padding and LeakyReLU activations
with slope $0.01$.
Downsampling uses average pooling and upsampling uses nearest-neighbor interpolation.

\subsection{Reaction Constitutive Network}
\label{app:reaction_network}

The reaction module is a pointwise MLP with weights shared across grid points.
It has two hidden layers of width 32, $\tanh$ activations, and a residual connection
between the hidden layers.
The network maps local species densities to unrestricted relative reaction rates
without using neighboring grid values.
Weights use Xavier normal initialization, and biases are initialized to zero.
For Schnakenberg, two independent transport CNOs process $U$ and $V$, respectively,
while the reaction MLP takes both local concentrations and predicts their two relative
reaction rates using the same hidden architecture.
\section{Numerical Updates and Rollout Algorithm of MCT}
\label{app:numerical_updates}

Algorithm~\ref{alg:method_rollout} summarizes the split MCT integrator for a single species, following Section~\ref{sec:method_coupling}.
Each step applies a reaction half-step, a transport step, and a second reaction half-step.
The transport velocity $\mathbf v=\widehat{\mathbf u}[\rho^n]$ is evaluated at the beginning of the step and held fixed throughout it, while the relative reaction rates are reevaluated at each reaction stage.

\paragraph{Reaction update.}
With $I$ fixed, the reaction subproblem is $\partial_t M=Mr$.
Using $m=\log M$ gives $\partial_t m=r$.
Let $\mathcal R$ denote the relative-reaction-rate map.
An explicit midpoint update over $h=\Delta t/2$ is
\begin{equation}
    \begin{aligned}
        k_1&=\mathcal R(Ie^m),\qquad
        k_2=\mathcal R\!\left(Ie^{m+(h/2)k_1}\right),\\
        M^+&=e^{m+hk_2}.
    \end{aligned}
    \label{eq:app_reaction_midpoint}
\end{equation}
Products and exponentials act pointwise.
Nonreactive systems omit these updates.

\paragraph{Compression and mass transport.}
Let $\mathcal L_{\mathbf v}$ denote the unsplit conservative upwind finite-volume discretization of $-\nabla\cdot(I\mathbf v)$, using arithmetic averages of neighboring velocities at cell faces.
The compression factor is advanced by SSPRK2:
\begin{equation}
    \begin{aligned}
        I^{(1)}&=I^n+\Delta t\,\mathcal L_{\mathbf v}(I^n),\\
        I^{n+1}&=\tfrac12 I^n+
            \tfrac12\left[I^{(1)}+\Delta t\,\mathcal L_{\mathbf v}(I^{(1)})\right].
    \end{aligned}
    \label{eq:app_compression_ssprk2}
\end{equation}
For mass transport, let $\mathcal P[q](\mathbf x)$ denote periodic interpolation of the grid field $q$.
The midpoint semi-Lagrangian update of the post-reaction factor $M^a$ is
\begin{equation}
    \begin{aligned}
        \mathbf x_{\mathrm{mid}}
        &=\mathbf x-\tfrac{\Delta t}{2}\mathbf v(\mathbf x),\\
        \mathbf x_{\mathrm{dep}}
        &=\mathbf x-\Delta t\,
            \mathcal P[\mathbf v](\mathbf x_{\mathrm{mid}}),\\
        M^b(\mathbf x)
        &=\mathcal P[M^a](\mathbf x_{\mathrm{dep}}).
    \end{aligned}
    \label{eq:app_mass_semi_lagrangian}
\end{equation}
The two-dimensional implementation uses bilinear interpolation.
Interpolating $\mathbf v$ does not reevaluate the constitutive module.
The second reaction half-step then uses $M^b$ and $I^{n+1}$.

\begin{algorithm}[t]
    \caption{Single-species rollout with \NCLMCT}
    \label{alg:method_rollout}
    \small
    \begin{algorithmic}[1]
        \Require Initial density $\rho^0$, constitutive modules, step size $\Delta t$,
        number of steps $K$, reinitialization interval $T_{\mathrm{reinit}}$
        \Ensure Predicted densities $\{\rho^n\}_{n=1}^{K}$
        \State Initialize $I^0\gets1$, $M^0\gets\rho^0$, and $\tau\gets0$.
        \For{$n=0,\ldots,K-1$}
            \State Evaluate $\mathbf v\gets\widehat{\mathbf u}[\rho^n]$.
            \State $M^a\gets$ \Call{ReactionStep}{$M^n,I^n,\Delta t/2$}.
            \State Update $I^n\mapsto I^{n+1}$ using
            \eqnref{app_compression_ssprk2}.
            \State Update $M^a\mapsto M^b$ using
            \eqnref{app_mass_semi_lagrangian}.
            \State $M^{n+1}\gets$
            \Call{ReactionStep}{$M^b,I^{n+1},\Delta t/2$}.
            \State Reconstruct $\rho^{n+1}\gets M^{n+1}I^{n+1}$.
            \State $\tau\gets\tau+\Delta t$.
            \If{$\tau\geq T_{\mathrm{reinit}}$}
                \State $I^{n+1}\gets1$, $M^{n+1}\gets\rho^{n+1}$, and $\tau\gets0$.
            \EndIf
        \EndFor
        \Function{ReactionStep}{$M,I,h$}
            \If{the system is nonreactive}
                \State \Return $M$
            \EndIf
            \State Set $m\gets\log M$ and apply \eqnref{app_reaction_midpoint}.
            \State \Return $M^+$.
        \EndFunction
    \end{algorithmic}
\end{algorithm}

\paragraph{Boundary treatment and reinitialization.}
The MCT integrator supports periodic and Robin boundary conditions, while the constitutive modules used here assume periodic domains.
All benchmarks therefore use periodic boundaries, implemented through circular convolutional padding and periodic grid indexing and interpolation.
Reinitialization resets $I=1$ and $M=\rho$ every $T_{\mathrm{reinit}}$, limiting factor distortion while preserving the reconstructed density rather than correcting accumulated density error.

\paragraph{Scope of numerical properties.}
For the finite-volume update in \eqnref{app_compression_ssprk2}, periodic fluxes cancel in the discrete integral of $I$, while positivity requires the corresponding time-step restriction.
The logarithmic reaction update preserves $M>0$ in exact arithmetic, and bilinear interpolation preserves nonnegativity. 
These substep properties do not imply exact conservation of the total reconstructed density mass or unconditional discrete energy dissipation.
Because the transport velocity is frozen at the beginning of each step, the midpoint and SSPRK2 substeps also do not establish second-order accuracy of the complete coupled method.
Appendix~\ref{app:physical_diagnostics} therefore evaluates the complete rollout empirically.
\clearpage
\section{PDE Specifications}
\label{app:pde_specifications}

This appendix specifies each benchmark system: its governing equation and parameters,
transport constitutive pair $(\xi,\mathbf f)$, relative reaction rate $r$, and reference
solution.
All systems are posed on square periodic domains and discretized on a
$128\times128$ uniform grid; their initial-condition families are given in
Appendix~\ref{app:ic_families}.
Table~\ref{tab:system_overview} summarizes the configurations.

\begin{table}[b!]
\centering
\caption{
Per-system configurations.
``Order'' denotes the highest spatial derivative order in the transport term of the
density equation, and ``Reaction'' the local reaction source type.
$\Delta t$ and $N_T$ denote the reference time-step size and number of time steps,
respectively.
$^\dagger$\,Degenerate transport, for which the flux vanishes as $\rho\to0$ at the
free boundary.
}
\label{tab:system_overview}
\begin{tabular}{lcccccc}
\toprule
System & Domain & Order & Reaction & $t$ range & $\Delta t$ & $N_T$ \\
\midrule
Linear Diffusion     & $[0,1)^2$    & 2            & None                & $[0,0.03]$  & $5\times10^{-5}$ & 600 \\
CH                   & $[0,1)^2$    & 4            & None                & $[0,0.1]$   & $10^{-5}$        & $10^{4}$ \\
PM                   & $[-4,4)^2$   & 2$^\dagger$  & None                & $[0.1,0.3]$ & $10^{-4}$        & 2000 \\
\midrule
Fisher-KPP           & $[0,1)^2$    & 2            & Logistic            & $[0,0.015]$ & $3\times10^{-5}$ & 500 \\
Reactive CH          & $[0,1)^2$    & 4            & Double-well         & $[0,0.1]$   & $10^{-5}$        & $10^{4}$ \\
Reactive PM          & $[-4,4)^2$   & 2$^\dagger$  & Logistic            & $[0.1,0.6]$ & $5\times10^{-4}$ & 1000 \\
\midrule
Schnakenberg         & $[0,1)^2$    & 2            & Activator-inhibitor & $[0,1]$     & $5\times10^{-5}$ & $2\times10^{4}$ \\
\bottomrule
\end{tabular}
\end{table}

%%%%%%%%%%%%%%%%%%%%%%%%%%%%%%%%%%%%%%%

\subsection{Linear Diffusion}
\label{app:settings_diffusion}

\paragraph{PDE definition.}
Linear diffusion on the periodic unit square $\Omega=[0,1)^2$ is
\begin{equation}
\partial_t\rho=D\,\Delta\rho,
\qquad D=1,
\qquad t\in[0,0.03].
\label{eq:app_diffusion_pde}
\end{equation}
The system is nonreactive, with transport constitutive pair
\begin{equation}
\xi=\frac{1}{\rho},
\qquad
\mathbf f=-D\nabla\rho,
\qquad
\mathbf u=\xi\mathbf f=-D\nabla\log\rho.
\label{eq:app_diffusion_constitutive}
\end{equation}
This corresponds to the free energy
$\mathcal E[\rho]=\tfrac{D}{2}\int_\Omega\rho^2\,d\mathbf x$.

\paragraph{Reference solution.}
Because \eqnref{app_diffusion_pde} is linear and periodic, the reference is given by
exact Fourier-mode decay,
\begin{equation}
\widehat\rho(\mathbf k,t)
=
\widehat\rho(\mathbf k,0)\exp\bigl(-D\lvert\mathbf k\rvert^2t\bigr),
\label{eq:app_diffusion_reference}
\end{equation}
and therefore introduces no time-integration error.
Trajectories are stored at 601 uniformly spaced times over $[0,0.03]$, giving
$\Delta t=5\times10^{-5}$.

\subsection{Cahn-Hilliard}
\label{app:settings_ch}

\paragraph{PDE definition.}
The shifted Cahn-Hilliard equation on $\Omega=[0,1)^2$ is
\begin{equation}
\partial_t\rho=\Delta\mu,
\qquad
\mu=-\gamma_1\Delta\rho+\gamma_2W'(\rho),
\qquad
W(\rho)=\tfrac14\bigl((\rho-\rho_c)^2-h^2\bigr)^2.
\label{eq:app_ch_pde}
\end{equation}
We use $\gamma_1=10^{-4}$, $\gamma_2=1$, $\rho_c=1$, $h=0.5$, and
$t\in[0,0.1]$, so the two wells lie at $\rho=0.5$ and $\rho=1.5$.
The system is nonreactive, with transport constitutive pair
\begin{equation}
\xi=\frac{1}{\rho},
\qquad
\mathbf f=-\nabla\mu.
\label{eq:app_ch_constitutive}
\end{equation}
Here $\mu=\delta\mathcal E/\delta\rho$ for
\[
\mathcal E[\rho]
=
\int_\Omega
\left(
\tfrac{\gamma_1}{2}\lvert\nabla\rho\rvert^2
+\gamma_2W(\rho)
\right)d\mathbf x.
\]

\paragraph{Reference solution.}
We use a Fourier pseudospectral discretization with a semi-implicit update that treats
the biharmonic term implicitly and the double-well term explicitly:
\begin{equation}
\widehat\rho^{\,n+1}
=
\frac{
\widehat\rho^{\,n}
-\Delta t\,\gamma_2\lvert\mathbf k\rvert^2
\widehat{W'}(\rho^n)
}{
1+\Delta t\,\gamma_1\lvert\mathbf k\rvert^4
}.
\label{eq:app_ch_reference}
\end{equation}
Each step clamps $\rho$ to $[0.45,1.55]$ and restores the initial mean density.
Trajectories are stored at $10{,}001$ uniformly spaced times over $[0,0.1]$, giving
$\Delta t=10^{-5}$.

\subsection{Porous Medium}
\label{app:settings_pm}

\paragraph{PDE definition.}
The porous medium equation on $\Omega=[-4,4)^2$ is
\begin{equation}
\partial_t\rho=\Delta(\rho^m),
\qquad m=2,
\qquad t\in[0.1,0.3].
\label{eq:app_pm_pde}
\end{equation}
Its transport form uses the pressure
$p(\rho)=\tfrac{m}{m-1}\rho^{m-1}$ and transport constitutive pair
\begin{equation}
\xi=1,
\qquad
\mathbf f=-\nabla p(\rho)=-m\rho^{m-2}\nabla\rho.
\label{eq:app_pm_constitutive}
\end{equation}
This corresponds to the internal energy
$\mathcal E[\rho]=\tfrac{1}{m-1}\int_\Omega\rho^m\,d\mathbf x$.
For $m=2$, $\mathbf f=-2\nabla\rho$, so the transport response contains no
$1/\rho$ singularity at the free boundary.

\paragraph{Reference solution.}
The reference is the closed-form Barenblatt similarity solution of
\eqnref{app_pm_pde}, evaluated at each stored time.
It therefore introduces no time-discretization error and represents compact support
exactly through the positive-part operator.
Trajectories are stored at $2{,}001$ uniformly spaced times over $[0.1,0.3]$, giving
$\Delta t=10^{-4}$.

\subsection{Fisher-KPP}
\label{app:settings_fkpp}

\paragraph{PDE definition.}
The Fisher-KPP equation on $\Omega=[0,1)^2$ is
\begin{equation}
\partial_t\rho=D\,\Delta\rho+\lambda\rho(1-\rho),
\qquad D=1,
\qquad \lambda=5,
\qquad t\in[0,0.015].
\label{eq:app_fkpp_pde}
\end{equation}
Its transport and reaction responses are
\begin{equation}
\xi=\frac{1}{\rho},
\qquad
\mathbf f=-D\nabla\rho,
\qquad
r=\lambda(1-\rho).
\label{eq:app_fkpp_constitutive}
\end{equation}
The transport pair is identical to linear diffusion, while the logistic source
corresponds to the relative reaction rate $r$.

\paragraph{Reference solution.}
The reference uses Strang splitting, with exact Fourier-mode decay for diffusion and
the closed-form logistic update
\begin{equation}
\rho\mapsto
\frac{\rho}{\rho+(1-\rho)e^{-\lambda\Delta t}},
\label{eq:app_fkpp_reference}
\end{equation}
so splitting is the only source of time-discretization error.
Trajectories are stored at 501 uniformly spaced times over $[0,0.015]$, giving
$\Delta t=3\times10^{-5}$.

\subsection{Reactive Cahn-Hilliard}
\label{app:settings_reactive_ch}

\paragraph{PDE definition.}
We augment \eqnref{app_ch_pde} with the local reaction source
\begin{equation}
\partial_t\rho=\Delta\mu+G(\rho),
\qquad
G(\rho)=h\lambda\,\phi(1-\phi^2),
\qquad
\phi=\frac{\rho-\rho_c}{h}.
\label{eq:app_rch_pde}
\end{equation}
We retain $\gamma_1=10^{-4}$, $\gamma_2=1$, $\rho_c=1$, $h=0.5$, and
$t\in[0,0.1]$, with $\lambda=5$.
The transport pair is that of Appendix~\ref{app:settings_ch}, and the relative
reaction rate is $r=G(\rho)/\rho$.
The source drives $\phi$ toward $\pm1$, reinforcing phase separation rather than
dissipating the Cahn-Hilliard free energy.

\paragraph{Reference solution.}
The reference uses Strang splitting between the Cahn-Hilliard update
\eqnref{app_ch_reference} and a closed-form reaction update in the phase variable
$\phi$.
Each step clamps $\rho$ to $[0.45,1.55]$.
Trajectories are stored at $10{,}001$ uniformly spaced times over $[0,0.1]$, giving
$\Delta t=10^{-5}$.

\subsection{Reactive Porous Medium}
\label{app:settings_reactive_pm}

\paragraph{PDE definition.}
We augment \eqnref{app_pm_pde} with a logistic source on $\Omega=[-4,4)^2$:
\begin{equation}
\partial_t\rho=\Delta(\rho^m)+\lambda\rho(1-\rho),
\qquad m=2,
\qquad \lambda=5,
\qquad t\in[0.1,0.6].
\label{eq:app_rpm_pde}
\end{equation}
The transport pair is that of Appendix~\ref{app:settings_pm}, with relative reaction
rate $r=\lambda(1-\rho)$.

\paragraph{Reference solution.}
The reference uses Strang splitting between the closed-form logistic reaction and an
explicit nonlinear-diffusion update using a second-order five-point Laplacian on
$\rho^m$.
Physical-space differentiation avoids Gibbs oscillations from Fourier differentiation
near the free boundary.
The density is kept nonnegative after each substep.
Trajectories are stored at $1{,}001$ uniformly spaced times over $[0.1,0.6]$, giving
$\Delta t=5\times10^{-4}$.

\subsection{Schnakenberg System}
\label{app:settings_schnakenberg}

\paragraph{PDE definition.}
The two-species Schnakenberg system on $\Omega=[0,1)^2$ is
\begin{equation}
\begin{aligned}
\partial_tU&=D_U\Delta U+\gamma\bigl(a-U+U^2V\bigr),\\
\partial_tV&=D_V\Delta V+\gamma\bigl(b-U^2V\bigr).
\end{aligned}
\label{eq:app_schnakenberg_pde}
\end{equation}
We use $D_U=8\times10^{-3}$, $D_V=1.6\times10^{-1}$, $\gamma=36$,
$a=0.171$, $b=0.629$, and $t\in[0,1]$.
The homogeneous steady state
$U^\ast=a+b$, $V^\ast=b/(a+b)^2$
lies in the Turing-unstable regime.
Each species has its own transport constitutive pair and relative reaction rate:
\begin{equation}
\xi_s=\frac{1}{s},
\qquad
\mathbf f_s=-D_s\nabla s,
\qquad
r_s=\frac{S_s(U,V)}{s},
\qquad s\in\{U,V\}.
\label{eq:app_schnakenberg_constitutive}
\end{equation}
Here $S_U$ and $S_V$ denote the reaction sources in
\eqnref{app_schnakenberg_pde}.

\paragraph{Reference solution.}
The reference uses Strang splitting: exact Fourier-mode diffusion half-steps enclose a
fourth-order Runge-Kutta step for the local reaction system.
Trajectories are stored at $20{,}001$ uniformly spaced times over $[0,1]$, giving
$\Delta t=5\times10^{-5}$.
\clearpage
\section{Dataset Generation and Training Settings}
\label{app:data_and_training}

Appendix~\ref{app:pde_specifications} specifies the equations, reference solvers, and
time intervals.
This appendix defines the initial-condition distributions, retained snapshots and
supervision targets, and training settings.

\subsection{Initial-Condition Families}
\label{app:ic_families}

Four of the seven systems draw initial conditions from a Gaussian Fourier random field
(GFRF),
\begin{equation}
g(x,y)=\!\!\sum_{\substack{0\le k_x,k_y\le K\\(k_x,k_y)\neq(0,0)}}\!\!
\frac{
a_{k_xk_y}\cos\!\big(2\pi(k_xx+k_yy)\big)
+b_{k_xk_y}\sin\!\big(2\pi(k_xx+k_yy)\big)
}{
(1+k_x^2+k_y^2)^{3/2}
},
\label{eq:app_gfrf}
\end{equation}
with $K\sim\mathcal U\{3,\dots,8\}$ and
$a_{k_xk_y},b_{k_xk_y}\stackrel{\text{i.i.d.}}{\sim}\mathcal N(0,1)$.
The spectral decay produces smooth fields, which are linearly rescaled to the positive
density range of each system.
The remaining three systems use structure-specific families: compactly supported
profiles for the two porous-medium systems and perturbations of the homogeneous steady
state for Schnakenberg.
Table~\ref{tab:ic_families} summarizes all seven families.

\begin{table}[b]
\centering
\small
\setlength{\tabcolsep}{4pt}
\caption{
Initial-condition family of each system.
GFRF denotes \eqnref{app_gfrf} rescaled to the stated density range.
Random parameters are sampled independently for each trajectory.
}
\label{tab:ic_families}
\begin{tabular}{lll}
\toprule
System & Family & Range or random parameters \\
\midrule
Linear Diffusion & GFRF, \eqnref{app_gfrf} & $\rho_0\in[0.1,1.0]$ \\
CH & GFRF, \eqnref{app_gfrf} & $\rho_0\in[0.7,1.3]$ \\
Fisher-KPP & GFRF, \eqnref{app_gfrf} & $\rho_0\in[0.2,0.8]$ \\
Reactive CH & GFRF, \eqnref{app_gfrf} & $\rho_0\in[0.7,1.3]$ \\
\midrule
PM & Barenblatt, \eqnref{app_pm_ic} & $R\sim\mathcal U[0.8,1.4]$ \\
Reactive PM & Radial bump, \eqnref{app_rpm_ic}
& $R\sim\mathcal U[0.8,2.0]$, $A\sim\mathcal U[0.5,0.9]$ \\
Schnakenberg & Perturbed steady state, \eqnref{app_schnakenberg_ic}
& $\varepsilon_U\sim\mathcal U[0.04,0.10]$,
  $\varepsilon_V\sim\mathcal U[0.12,0.22]$ \\
\bottomrule
\end{tabular}
\end{table}

\paragraph{Porous medium.}
We use Barenblatt profiles centered at the origin,
\begin{equation}
\rho(\mathbf x,t)=t^{-\alpha}
\bigl[C-k\lvert\mathbf x\rvert^2t^{-2\beta}\bigr]_+^{1/(m-1)},
\qquad
\alpha=\frac{d}{d(m-1)+2},
\quad
\beta=\frac{\alpha}{d},
\quad
k=\frac{(m-1)\beta}{2m},
\label{eq:app_pm_ic}
\end{equation}
with $d=2$ and $C$ chosen so that the support radius at $t=0.3$ is
$R\sim\mathcal U[0.8,1.4]$, keeping the support away from the boundary.
Because \eqnref{app_pm_ic} is an exact solution, it also provides the reference
trajectory (Appendix~\ref{app:settings_pm}).

\paragraph{Reactive porous medium.}
Because the logistic source is incompatible with \eqnref{app_pm_ic}, we use compactly
supported radial profiles,
\begin{equation}
\rho_0(\mathbf x)
=
A\Bigl[\max\Bigl(1-\frac{\lvert\mathbf x\rvert^2}{R^2},\,0\Bigr)\Bigr]^2,
\qquad
R\sim\mathcal U[0.8,2.0],
\qquad
A\sim\mathcal U[0.5,0.9].
\label{eq:app_rpm_ic}
\end{equation}
These profiles generate finite-support spreading fronts under the reactive
porous-medium dynamics.

\paragraph{Schnakenberg.}
We perturb the homogeneous steady state
$U^\ast=a+b$, $V^\ast=b/(a+b)^2$ multiplicatively:
\begin{equation}
U_0=U^\ast(1+\varepsilon_Ug_U),
\qquad
V_0=V^\ast(1+\varepsilon_Vg_V),
\label{eq:app_schnakenberg_ic}
\end{equation}
where $g_U$ and $g_V$ are independent realizations of \eqnref{app_gfrf} with
$K\sim\mathcal U\{2,\dots,6\}$, normalized to unit maximum magnitude.
The amplitudes
$\varepsilon_U\sim\mathcal U[0.04,0.10]$ and
$\varepsilon_V\sim\mathcal U[0.12,0.22]$
are chosen so that pattern formation is driven primarily by the Turing instability.

\subsection{Trajectories, Snapshots and Supervision}
\label{app:snapshots}

For each system, we sample 100 training and 10 test initial conditions with disjoint
random seeds and integrate them over the interval in
Table~\ref{tab:system_overview}.
Ten snapshots per training trajectory are selected at approximately uniform increments
of cumulative state-space arc length, measured by successive relative $L^2$ changes and
including the initial and terminal states.
This yields $1{,}000$ training snapshots per system; for multispecies systems, all
species are retained at the same times.

For \NCLMCT, constitutive targets are evaluated on the fly at each snapshot using the
analytical laws in Appendix~\ref{app:pde_specifications}.
Each snapshot therefore supervises the map from the current density to its constitutive
responses rather than to a future density state.
Sequential held-out trajectories are used for rollout evaluation.

\subsection{Training and Baselines}
\label{app:training_settings}

\paragraph{Schedule.}
Unless otherwise stated, all models are trained from the same $1{,}000$ snapshots per
system for $100{,}000$ updates, with batch size 32 (16 for Schnakenberg) and seed 42.
Our constitutive modules (Appendix~\ref{app:network_architectures}) use Adam with
learning rate $10^{-3}$, weight decay $10^{-6}$, and exponential decay by $0.95$ every
$5{,}000$ steps.
Each baseline retains its original optimizer settings described below; we equalize the
training data and update budget rather than the optimizer schedule.

\paragraph{F-FNO~\citep{tran2023ffno}.}
F-FNO maps the current density to the next state using six factorized Fourier layers of
width 64, retaining 32 modes per dimension.
Spectral weights are shared across layers, and each layer is followed by a feed-forward
block with expansion factor 4 ($733{,}506$ parameters).
Inputs and outputs are normalized channel-wise by the training mean and standard
deviation, and Gaussian noise with standard deviation $0.01$ is added to normalized
inputs during training.
Optimization uses AdamW with learning rate $2.5\times10^{-3}$, weight decay $10^{-4}$,
500 warm-up steps, and cosine decay, minimizing per-sample relative $L^2$ error.
The target is the reference state one native time step after each sampled state, and
rollout applies the learned map autoregressively for the full $N_T$ steps in
Table~\ref{tab:system_overview}.

\paragraph{CFO~\citep{hou2026cfo}.}
CFO learns the temporal derivative through flow matching rather than a fixed-increment
state map, and therefore uses the same ten trajectory snapshots without one-step targets.
Snapshot times are normalized to $[0,1]$, and a quintic spline through the ten states
provides time-derivative targets.
A two-dimensional U-Net with channel widths $(64,128,256)$, group normalization,
Swish activations, and a 256-dimensional sinusoidal time embedding
($8.10\times10^6$ parameters) regresses the spline derivative under a stochastic
interpolant with bridge amplitude $10^{-5}$.
Optimization uses Adam with learning rate $2\times10^{-4}$ and unnormalized mean
squared error.
Rollout integrates the learned temporal-derivative field on the reference time grid
using Heun's method with two substeps per interval.

\paragraph{DOOL~\citep{chang2025dool}.}
DOOL is trained without response targets by minimizing an Onsager Rayleighian on the
same $1{,}000$ snapshots.
A Fourier DeepONet ($64{,}620$ parameters) predicts the two components of the mass flux.
Its branch network reads the lowest $8\times8$ Fourier coefficients of the density,
while the trunk network reads normalized grid coordinates; both use four $\tanh$ hidden
layers of width 70 with rank 120.
For porous medium, the predicted flux is multiplied by the density so that it vanishes
in the empty region.
Optimization uses Adam with learning rate $10^{-3}$.
Rollout integrates
$\partial_t\rho=-\nabla\cdot\mathbf j$
using spectral divergence and Heun's method on the reference time grid, with density
clamped to the training range at each stage.
We evaluate the transport-only DOOL formulation on the three nonreactive systems;
reactive benchmarks would require additional problem-specific Onsager/Rayleighian
design.

\paragraph{Common protocol.}
All methods operate on the same $128\times128$ density fields and are initialized from
the same reference state at the first evaluation time.
No additional PDE-coefficient channels are supplied.
Architecture-specific inputs, including CFO's time embedding and DOOL's trunk
coordinates, are retained as required by their formulations.
Parameter counts and training/inference costs are reported in
Table~\ref{tab:cost}.
\clearpage

\section{Per-System Rollout Results}
\label{app:rollout_results}

This appendix shows predicted fields from one representative test trajectory at three
times for each system in Table~\ref{tab:reaction_diffusion}.
Evaluation uses the ten test initial conditions of Appendix~\ref{app:snapshots}, with
rollouts advanced on the reference time grids in Table~\ref{tab:system_overview} using
the reference solutions of Appendix~\ref{app:pde_specifications}.
Each figure shows one column per method, ordered as reference, F-FNO, CFO, DOOL where
applicable, \NCLMCT{} (Vel.), and \NCLMCT{} (Law), with time increasing downward and a
shared color scale within each system.

\subsection{Linear Diffusion}
\label{app:result_diffusion}

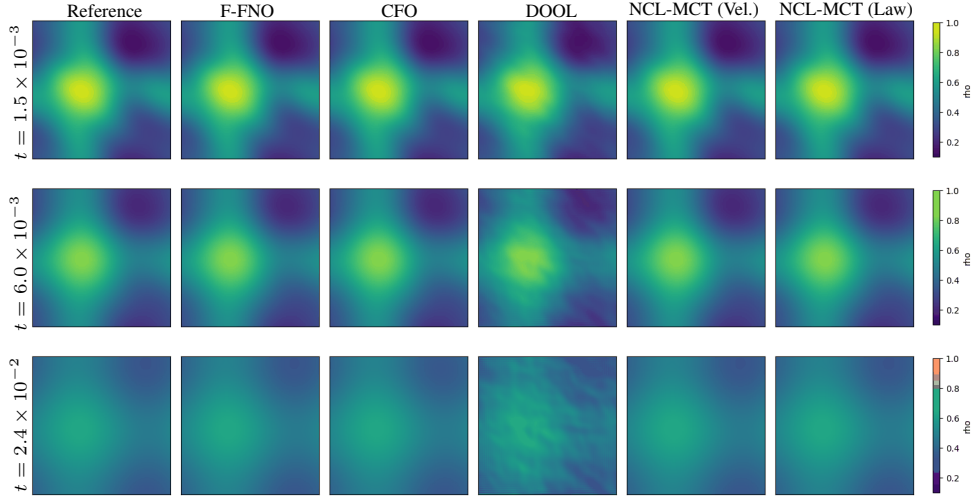
\begin{figure}[htbp]
\centering
\begin{overpic}[width=0.9\linewidth]{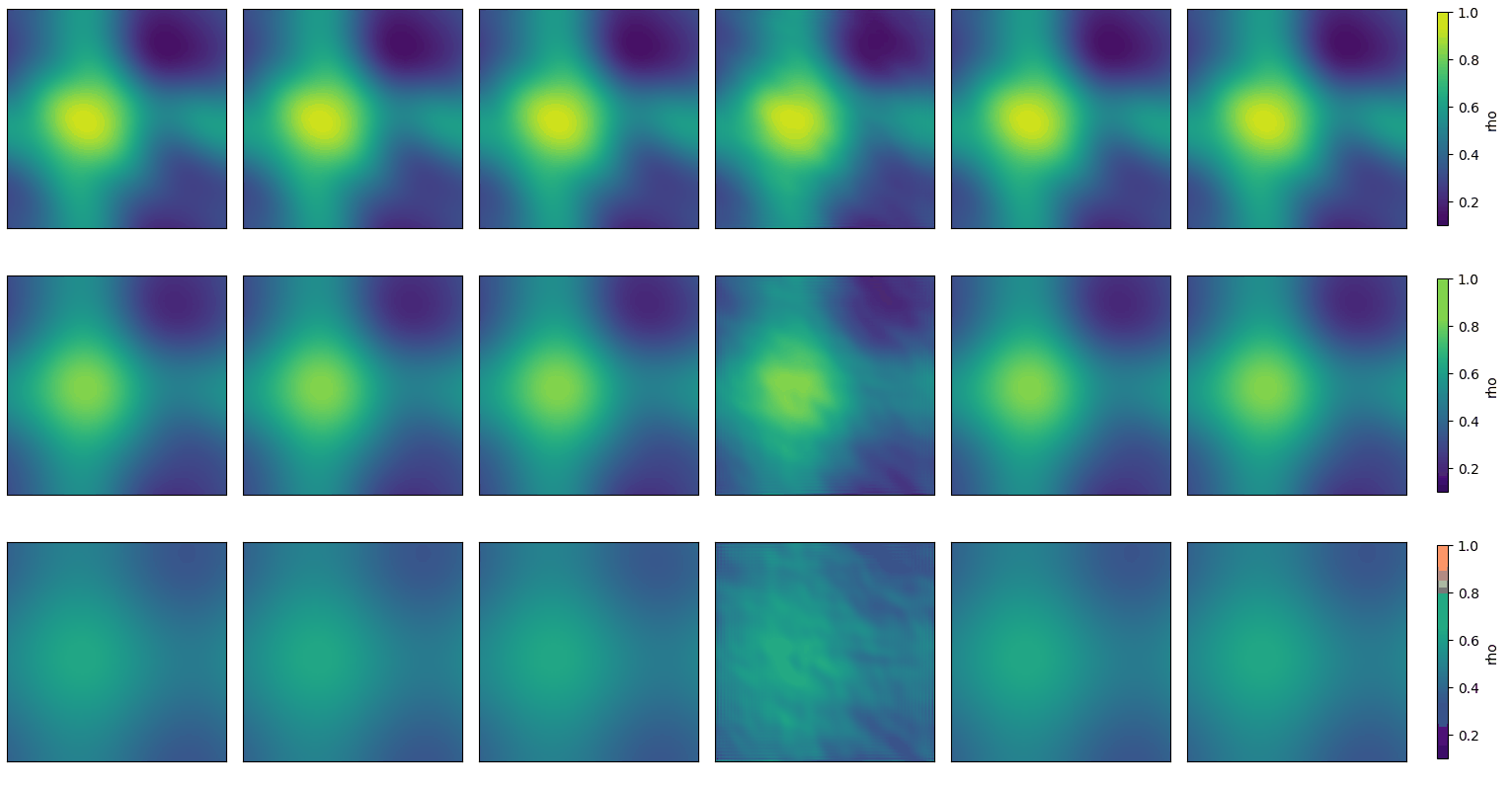}
        \put( 8,52.5){\makebox(0,0){\scriptsize Reference}}
        \put(23,52.5){\makebox(0,0){\scriptsize F-FNO}}
        \put(39,52.5){\makebox(0,0){\scriptsize CFO}}
        \put(55,52.5){\makebox(0,0){\scriptsize DOOL}}
        \put(70,52.5){\makebox(0,0){\scriptsize NCL-MCT (Vel.)}}
        \put(86,52.5){\makebox(0,0){\scriptsize NCL-MCT (Law)}}
        \put(-1,44){\makebox(0,0){\rotatebox{90}{\scriptsize $t=1.5\times10^{-3}$}}}
        \put(-1,26){\makebox(0,0){\rotatebox{90}{\scriptsize $t=6.0\times10^{-3}$}}}
        \put(-1,9){\makebox(0,0){\rotatebox{90}{\scriptsize $t=2.4\times10^{-2}$}}}
\end{overpic}
\caption{Linear diffusion, 600-step rollout to $T=0.03$.}
\label{fig:result_diffusion}
\end{figure}

Both \NCLMCT configurations track the smoothing dynamics throughout, with Vel. achieving about two-thirds of the Law error on both metrics and F-FNO remaining more accurate (Table~\ref{tab:reaction_diffusion}(a)).
All methods are visually similar at the first two times.
By $t=2.4\times10^{-2}$, the field is nearly uniform; only DOOL retains visible grid-aligned texture, while the remaining discrepancies are diffuse and low amplitude.

\subsection{Cahn-Hilliard}
\label{app:result_ch}

\begin{figure}[htbp]
\centering
\begin{overpic}[width=0.9\linewidth]{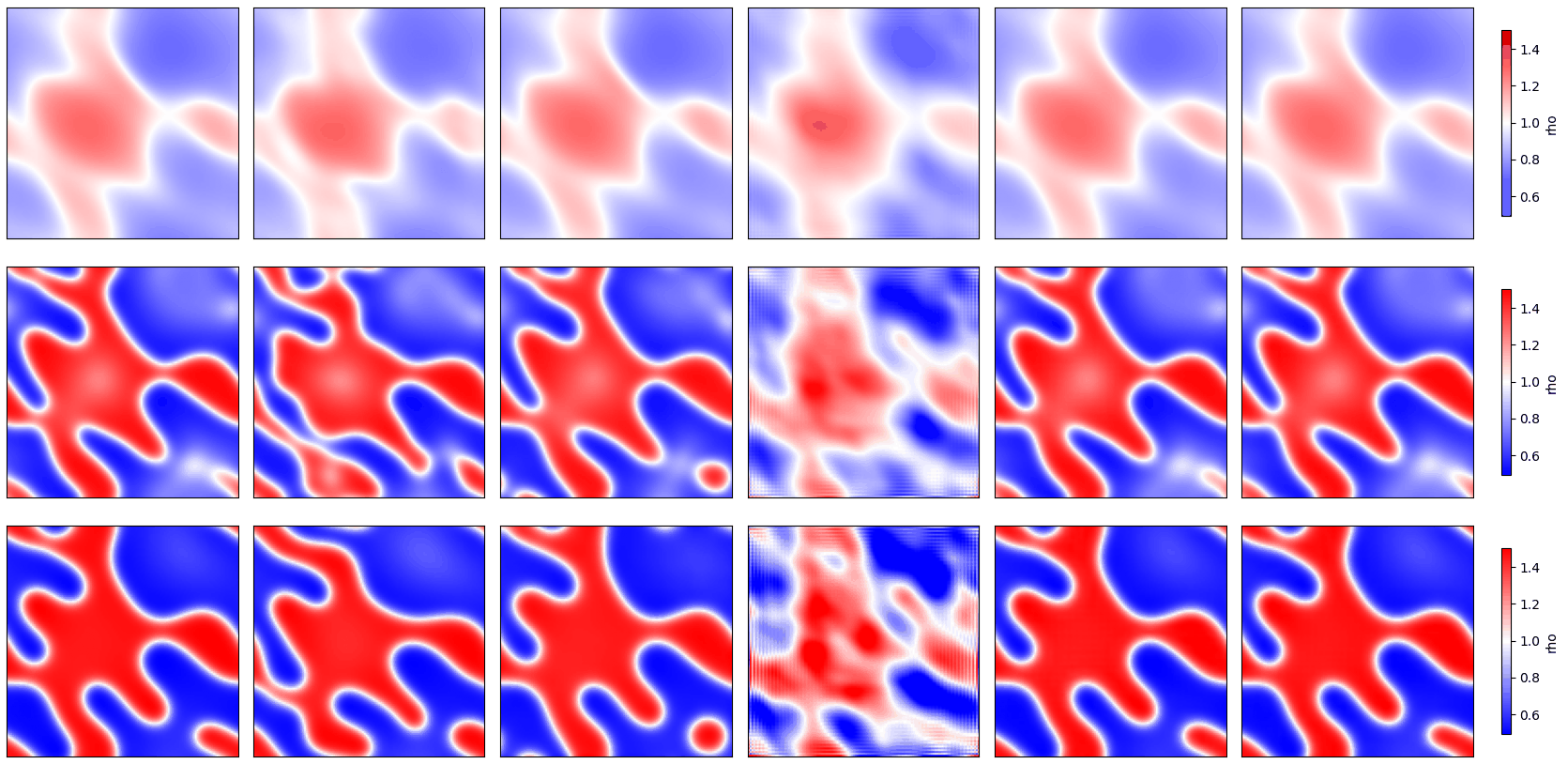}
        \put( 8,50){\makebox(0,0){\scriptsize Reference}}
        \put(23,50){\makebox(0,0){\scriptsize F-FNO}}
        \put(39,50){\makebox(0,0){\scriptsize CFO}}
        \put(55,50){\makebox(0,0){\scriptsize DOOL}}
        \put(70,50){\makebox(0,0){\scriptsize NCL-MCT (Vel.)}}
        \put(86,50){\makebox(0,0){\scriptsize NCL-MCT (Law)}}
        \put(-1,43){\makebox(0,0){\rotatebox{90}{\scriptsize $t=0.01$}}}
        \put(-1,25){\makebox(0,0){\rotatebox{90}{\scriptsize $t=0.05$}}}
        \put(-1,8){\makebox(0,0){\rotatebox{90}{\scriptsize $t=0.10$}}}
\end{overpic}
\caption{Cahn-Hilliard, 10{,}000-step rollout to $T=0.1$.}
\label{fig:result_ch}
\end{figure}
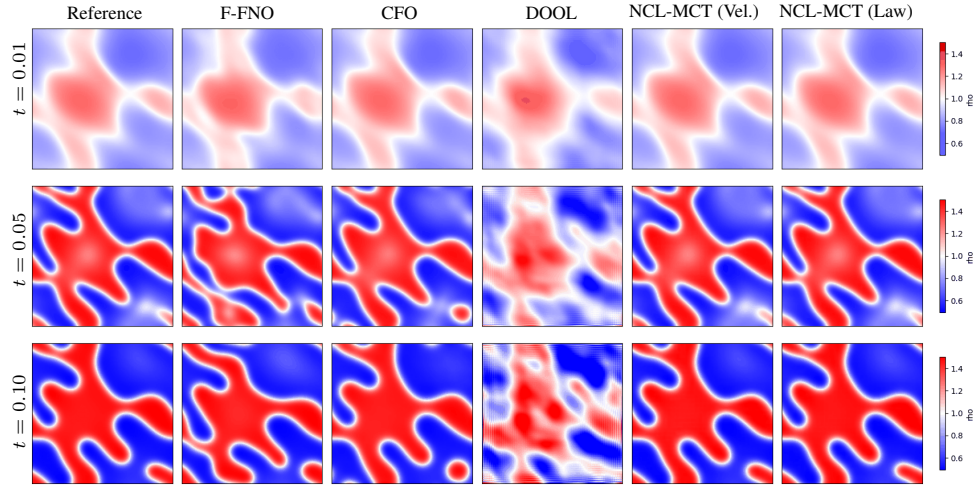

This system shows the largest margin over the evaluated baselines: both \NCLMCT configurations remain near $7.5\times10^{-3}$, roughly eight times below the strongest baseline (Table~\ref{tab:reaction_diffusion}(a)).
Both recover the domain count and locations at $T=0.1$, including the isolated droplet near the lower boundary.
F-FNO captures the two phases but distorts several interfaces, while DOOL loses the reference domain topology.

\subsection{Porous Medium}
\label{app:result_pm}

\begin{figure}[htbp]
\centering
\begin{overpic}[width=0.9\linewidth]{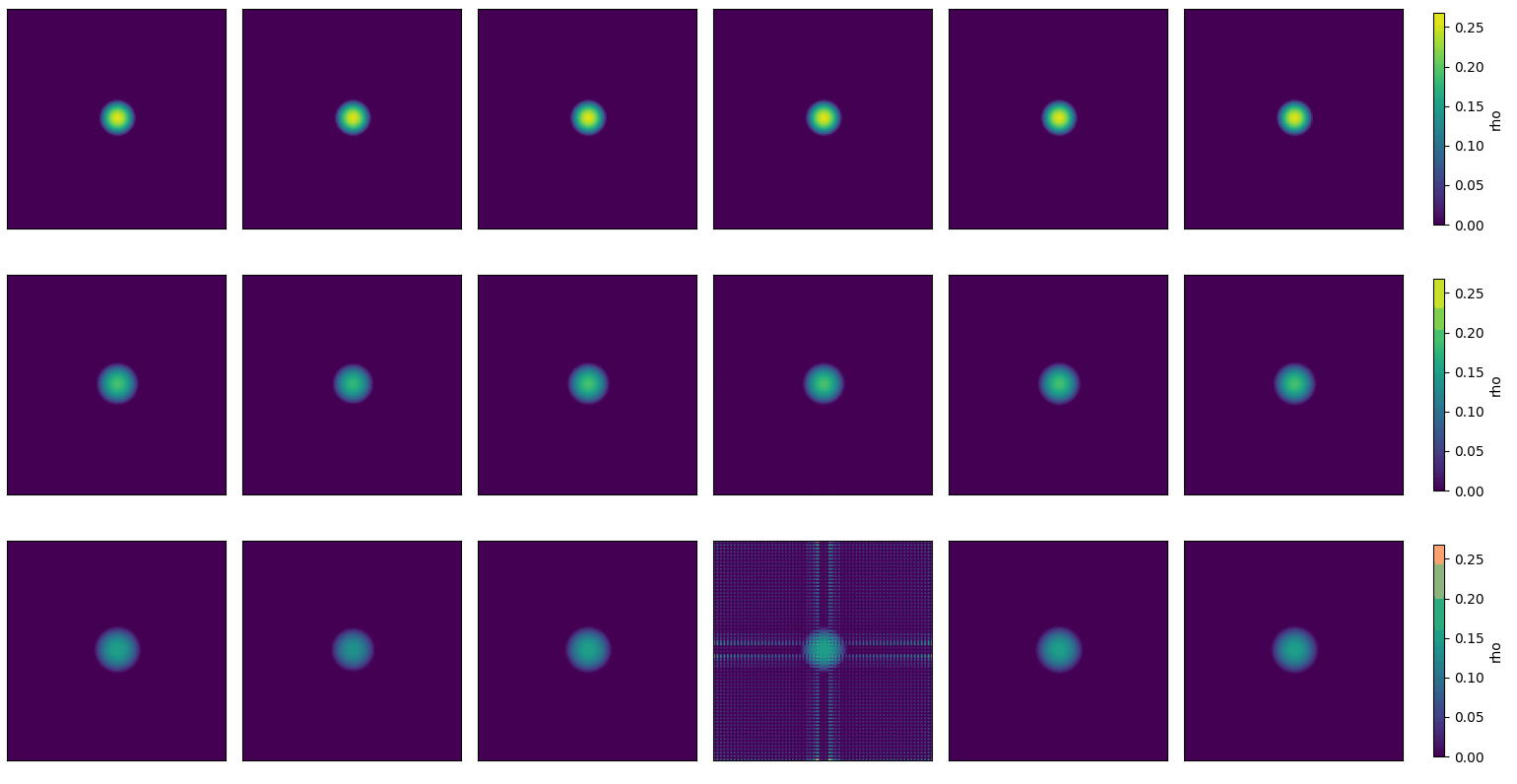}
        \put( 8,52.5){\makebox(0,0){\scriptsize Reference}}
        \put(23,52.5){\makebox(0,0){\scriptsize F-FNO}}
        \put(39,52.5){\makebox(0,0){\scriptsize CFO}}
        \put(55,52.5){\makebox(0,0){\scriptsize DOOL}}
        \put(70,52.5){\makebox(0,0){\scriptsize NCL-MCT (Vel.)}}
        \put(86,52.5){\makebox(0,0){\scriptsize NCL-MCT (Law)}}
        \put(-1,44){\makebox(0,0){\rotatebox{90}{\scriptsize $t=0.1$}}}
        \put(-1,26){\makebox(0,0){\rotatebox{90}{\scriptsize $t=0.2$}}}
        \put(-1,9){\makebox(0,0){\rotatebox{90}{\scriptsize $t=0.3$}}}
\end{overpic}
\caption{Porous medium, 2{,}000-step rollout from $t=0.1$ to $T=0.3$.}
\label{fig:result_pm}
\end{figure}
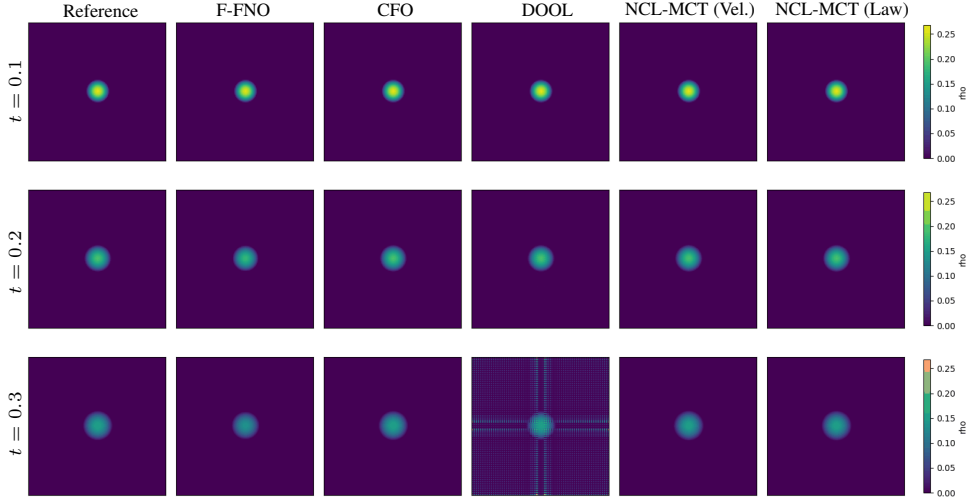

This is the only nonreactive system on which CFO is more accurate than both \NCLMCT configurations.
At this color scale, all methods except DOOL are visually close to the reference, so CFO's quantitative advantage in Table~\ref{tab:reaction_diffusion}(a) is not apparent from the fields alone.
DOOL develops grid-aligned artifacts extending from the support boundary, consistent with the difficulty of spectral differentiation across a free boundary.

\subsection{Fisher-KPP}
\label{app:result_fkpp}

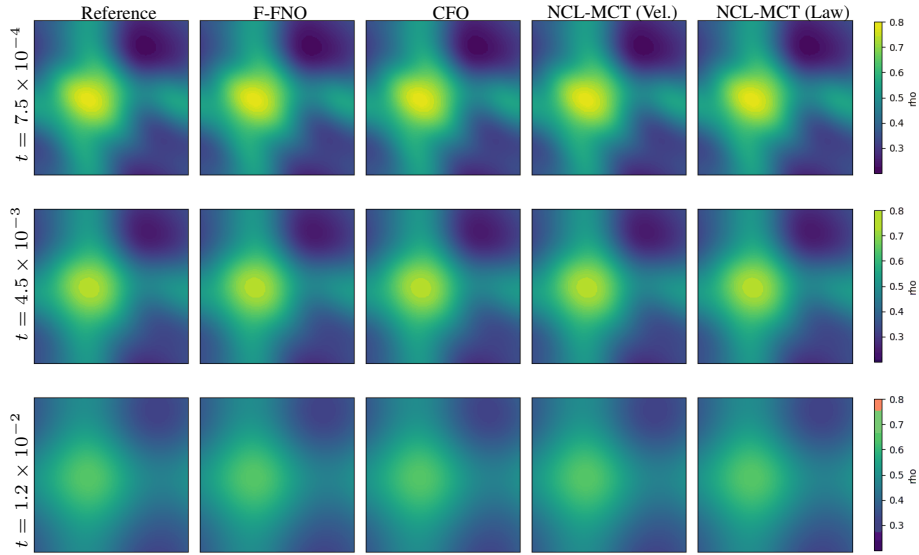
\begin{figure}[htbp]
\centering
\begin{overpic}[width=0.85\linewidth]{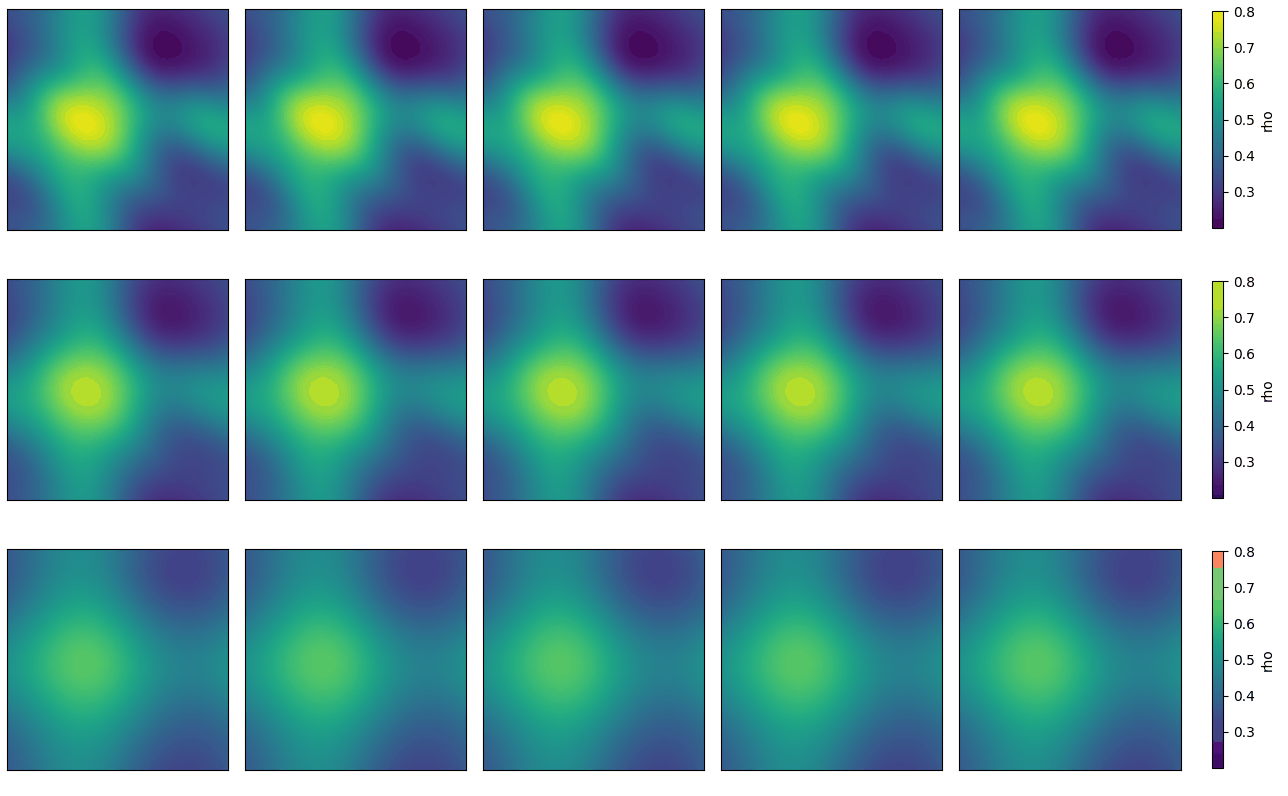}
        \put(10,62){\makebox(0,0){\scriptsize Reference}}
        \put(28,62){\makebox(0,0){\scriptsize F-FNO}}
        \put(47,62){\makebox(0,0){\scriptsize CFO}}
        \put(65,62){\makebox(0,0){\scriptsize NCL-MCT (Vel.)}}
        \put(84,62){\makebox(0,0){\scriptsize NCL-MCT (Law)}}
        \put(-1,53){\makebox(0,0){\rotatebox{90}{\scriptsize $t=7.5\times10^{-4}$}}}
        \put(-1,33){\makebox(0,0){\rotatebox{90}{\scriptsize $t=4.5\times10^{-3}$}}}
        \put(-1,10){\makebox(0,0){\rotatebox{90}{\scriptsize $t=1.2\times10^{-2}$}}}
\end{overpic}
\caption{Fisher-KPP, 500-step rollout to $T=0.015$.}
\label{fig:result_fkpp}
\end{figure}

Vel. is more accurate than Law on both metrics, giving the largest gap between the two \NCLMCT supervision modes among the scalar reactive systems, with F-FNO between them (Table~\ref{tab:reaction_diffusion}(b)).
The two \NCLMCT predictions are visually similar, indicating that the quantitative gap mainly reflects accumulated amplitude error rather than a qualitative difference in the predicted fields.

\subsection{Reactive Cahn-Hilliard}
\label{app:result_reactive_ch}

\begin{figure}[htbp]
\centering
\begin{overpic}[width=0.85\linewidth]{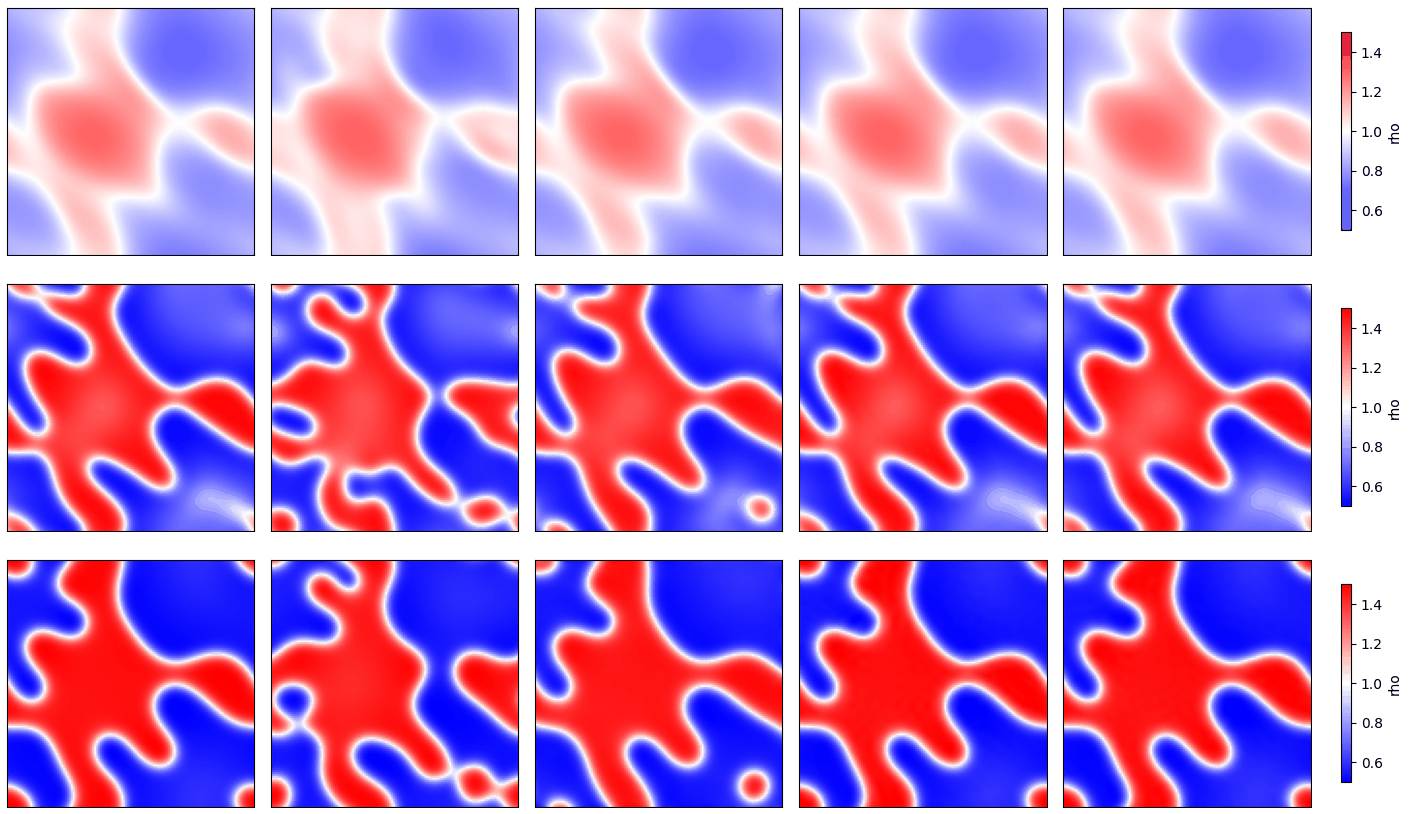}
        \put(10,58){\makebox(0,0){\scriptsize Reference}}
        \put(28,58){\makebox(0,0){\scriptsize F-FNO}}
        \put(47,58){\makebox(0,0){\scriptsize CFO}}
        \put(65,58){\makebox(0,0){\scriptsize NCL-MCT (Vel.)}}
        \put(84,58){\makebox(0,0){\scriptsize NCL-MCT (Law)}}
        \put(-1,48){\makebox(0,0){\rotatebox{90}{\scriptsize $t=0.01$}}}
        \put(-1,28){\makebox(0,0){\rotatebox{90}{\scriptsize $t=0.05$}}}
        \put(-1,10){\makebox(0,0){\rotatebox{90}{\scriptsize $t=0.10$}}}
\end{overpic}
\caption{Reactive Cahn-Hilliard, 10{,}000-step rollout to $T=0.1$.}
\label{fig:result_reactive_ch}
\end{figure}
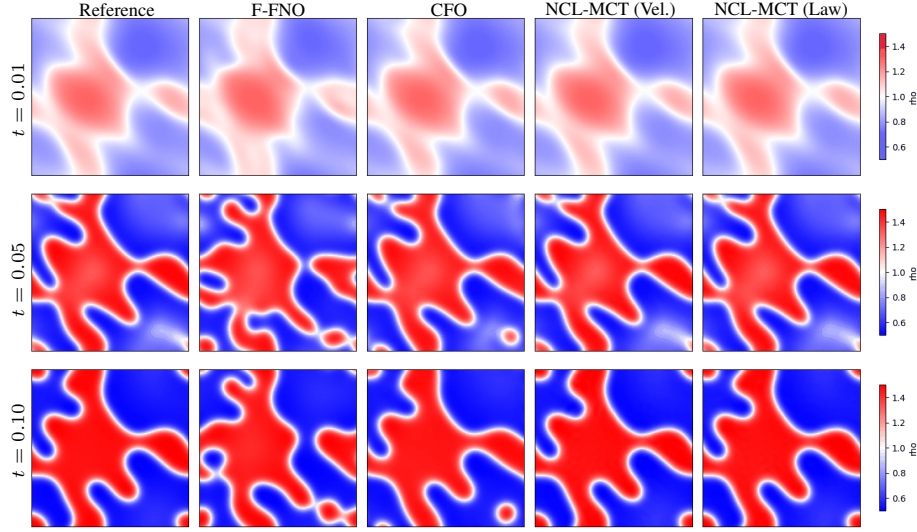

Errors for both \NCLMCT configurations remain close to those of the nonreactive CH system in Appendix~\ref{app:result_ch}, indicating that the added reaction module does not substantially degrade long-rollout accuracy.
The reaction drives both phases toward the wells, producing more saturated domain interiors than in Figure~\ref{fig:result_ch}.
Both configurations closely follow the reference, while F-FNO misplaces several smaller features.

\subsection{Reactive Porous Medium}
\label{app:result_reactive_pm}

\begin{figure}[htbp]
\centering
\begin{overpic}[width=0.85\linewidth]{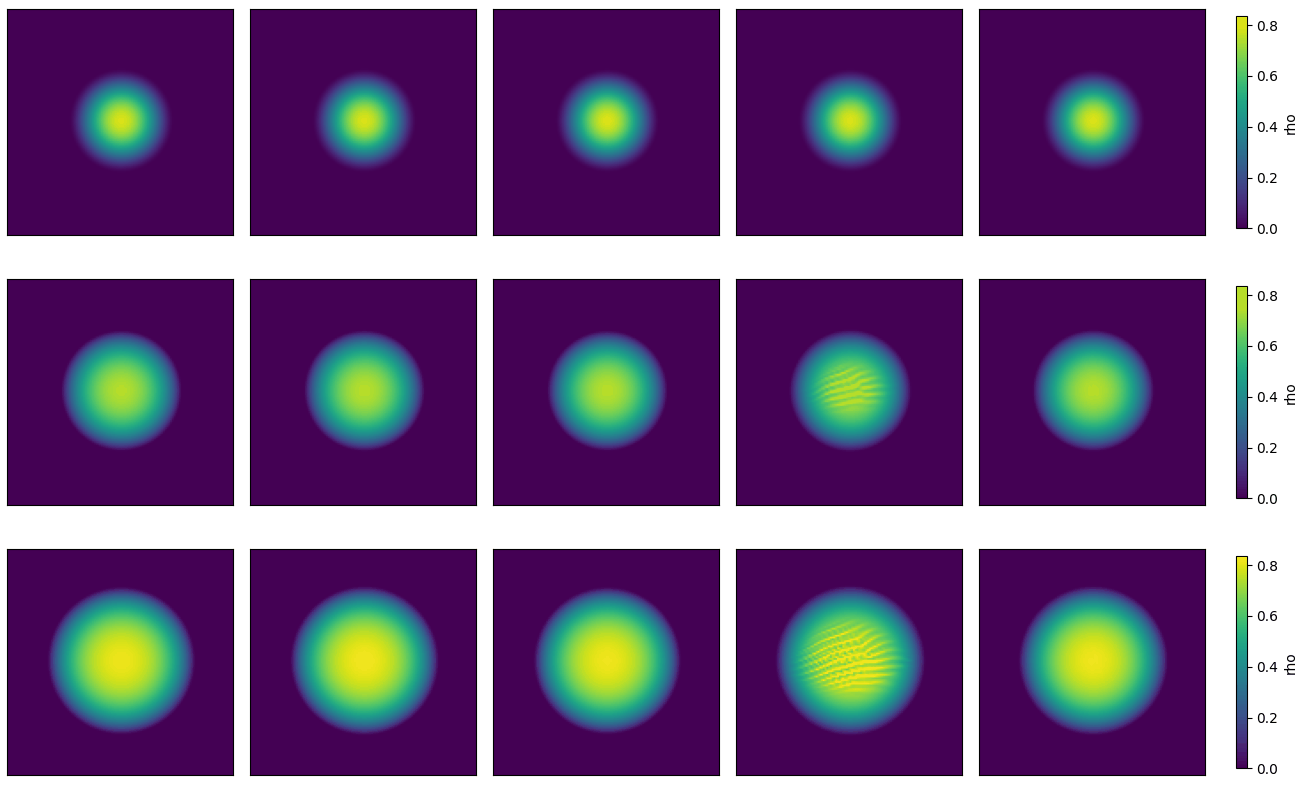}
        \put(10,62){\makebox(0,0){\scriptsize Reference}}
        \put(28,62){\makebox(0,0){\scriptsize F-FNO}}
        \put(47,62){\makebox(0,0){\scriptsize CFO}}
        \put(65,62){\makebox(0,0){\scriptsize NCL-MCT (Vel.)}}
        \put(84,62){\makebox(0,0){\scriptsize NCL-MCT (Law)}}
        \put(-1,50){\makebox(0,0){\rotatebox{90}{\scriptsize $t=0.125$}}}
        \put(-1,30){\makebox(0,0){\rotatebox{90}{\scriptsize $t=0.35$}}}
        \put(-1,10){\makebox(0,0){\rotatebox{90}{\scriptsize $t=0.60$}}}
\end{overpic}
\caption{Reactive porous medium, 1{,}000-step rollout from $t=0.1$ to $T=0.6$.}
\label{fig:result_reactive_pm}
\end{figure}
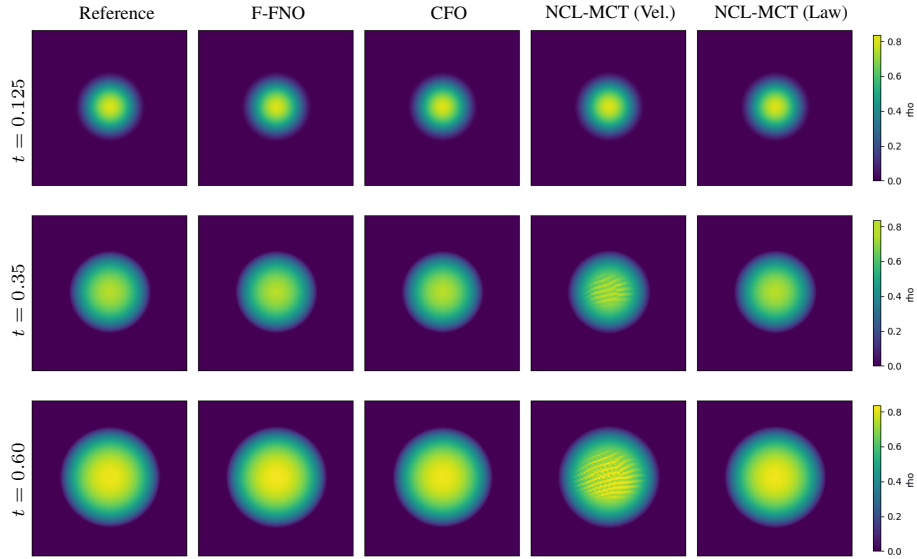

The ranking follows that of the nonreactive porous-medium system in
Appendix~\ref{app:result_pm}, consistent with degenerate transport remaining the dominant difficulty.
All methods capture the spreading and saturation of the support, but Vel. develops
grid-scale oscillations inside the support as the front advances.
These oscillations correspond to its higher error relative to Law in
Table~\ref{tab:reaction_diffusion}(b) and are absent in the nonreactive case of Figure~\ref{fig:result_pm}.

\subsection{Schnakenberg System}
\label{app:result_schnakenberg}

\begin{figure}[htbp]
\centering
\begin{overpic}[width=0.85\linewidth]{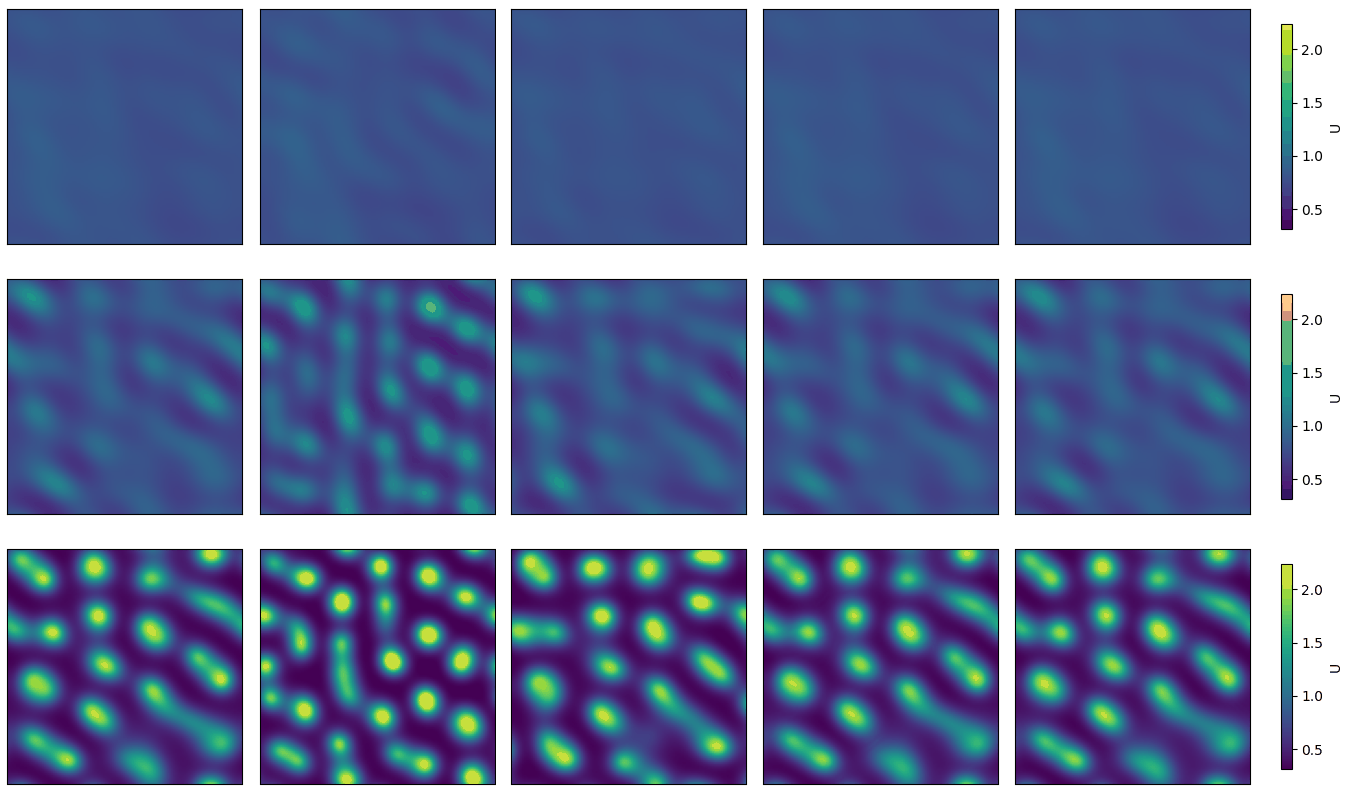}
        \put(10,60){\makebox(0,0){\scriptsize Reference}}
        \put(28,60){\makebox(0,0){\scriptsize F-FNO}}
        \put(47,60){\makebox(0,0){\scriptsize CFO}}
        \put(65,60){\makebox(0,0){\scriptsize NCL-MCT (Vel.)}}
        \put(84,60){\makebox(0,0){\scriptsize NCL-MCT (Law)}}
        \put(-1,50){\makebox(0,0){\rotatebox{90}{\scriptsize $t=0.2$}}}
        \put(-1,30){\makebox(0,0){\rotatebox{90}{\scriptsize $t=0.6$}}}
        \put(-1,10){\makebox(0,0){\rotatebox{90}{\scriptsize $t=1.0$}}}
\end{overpic}
\caption{Schnakenberg, activator $U$, 20{,}000-step rollout to $T=1$.}
\label{fig:result_schnakenberg_u}
\end{figure}

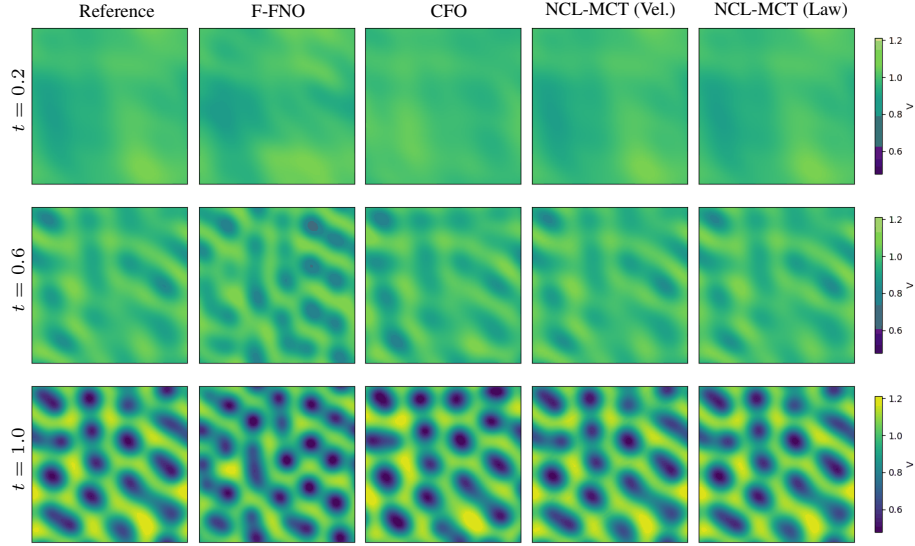
\begin{figure}[htbp]
\centering
\begin{overpic}[width=0.85\linewidth]{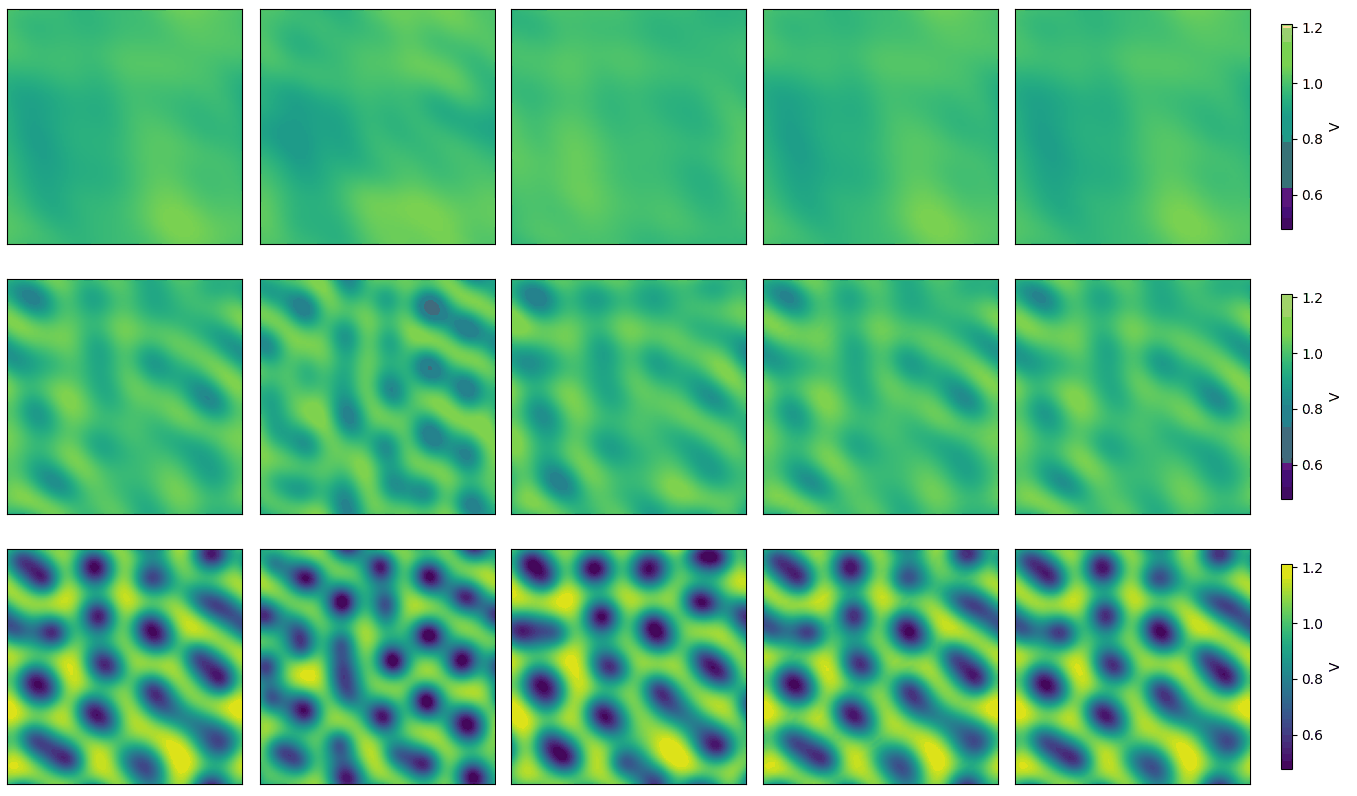}
        \put(10,60){\makebox(0,0){\scriptsize Reference}}
        \put(28,60){\makebox(0,0){\scriptsize F-FNO}}
        \put(47,60){\makebox(0,0){\scriptsize CFO}}
        \put(65,60){\makebox(0,0){\scriptsize NCL-MCT (Vel.)}}
        \put(84,60){\makebox(0,0){\scriptsize NCL-MCT (Law)}}
        \put(-1,50){\makebox(0,0){\rotatebox{90}{\scriptsize $t=0.2$}}}
        \put(-1,30){\makebox(0,0){\rotatebox{90}{\scriptsize $t=0.6$}}}
        \put(-1,10){\makebox(0,0){\rotatebox{90}{\scriptsize $t=1.0$}}}
\end{overpic}
\caption{Schnakenberg, inhibitor $V$, 20{,}000-step rollout to $T=1$.}
\label{fig:result_schnakenberg_v}
\end{figure}

This $20{,}000$-step rollout is the longest in the benchmark.
Both \NCLMCT configurations remain stable throughout, with mean errors differing by less than one standard deviation for both species.
F-FNO is about an order of magnitude less accurate and becomes nonfinite on one of ten test trajectories after roughly $18{,}000$ steps.
Both configurations reproduce the reference spot locations and elongated structures at $T=1$.
CFO captures the overall pattern type but displaces several spots, while F-FNO produces a finer, more uniform spot pattern and develops a localized circular defect from $t\approx0.6$ onward.
\clearpage

\section{Generalization Beyond Training Conditions}
\label{app:generalization}

Both experiments reuse the in-distribution checkpoints of
Appendix~\ref{app:training_settings} without modification:
no model is retrained or retuned, and no evaluation trajectory is used for training or
checkpoint selection.
Errors are reported at 101 uniformly spaced frames, following the in-distribution protocol.
We compare \NCLMCT{} (Vel.), \NCLMCT{} (Law), CFO, and F-FNO, together with
DOOL on the two nonreactive systems included in these tests.

%%%%%%%%%%%%%%%%%%%%%%%%%%%%%%%%%%%%%%%%%%%%%%%%%%%%%%%%%%%%%%%%%%%%%%%%%%%
\subsection{Unseen Initial Conditions}
\label{app:unseen_ic}

\begin{figure}[htbp]
\centering
\begin{overpic}[width=0.9\linewidth]{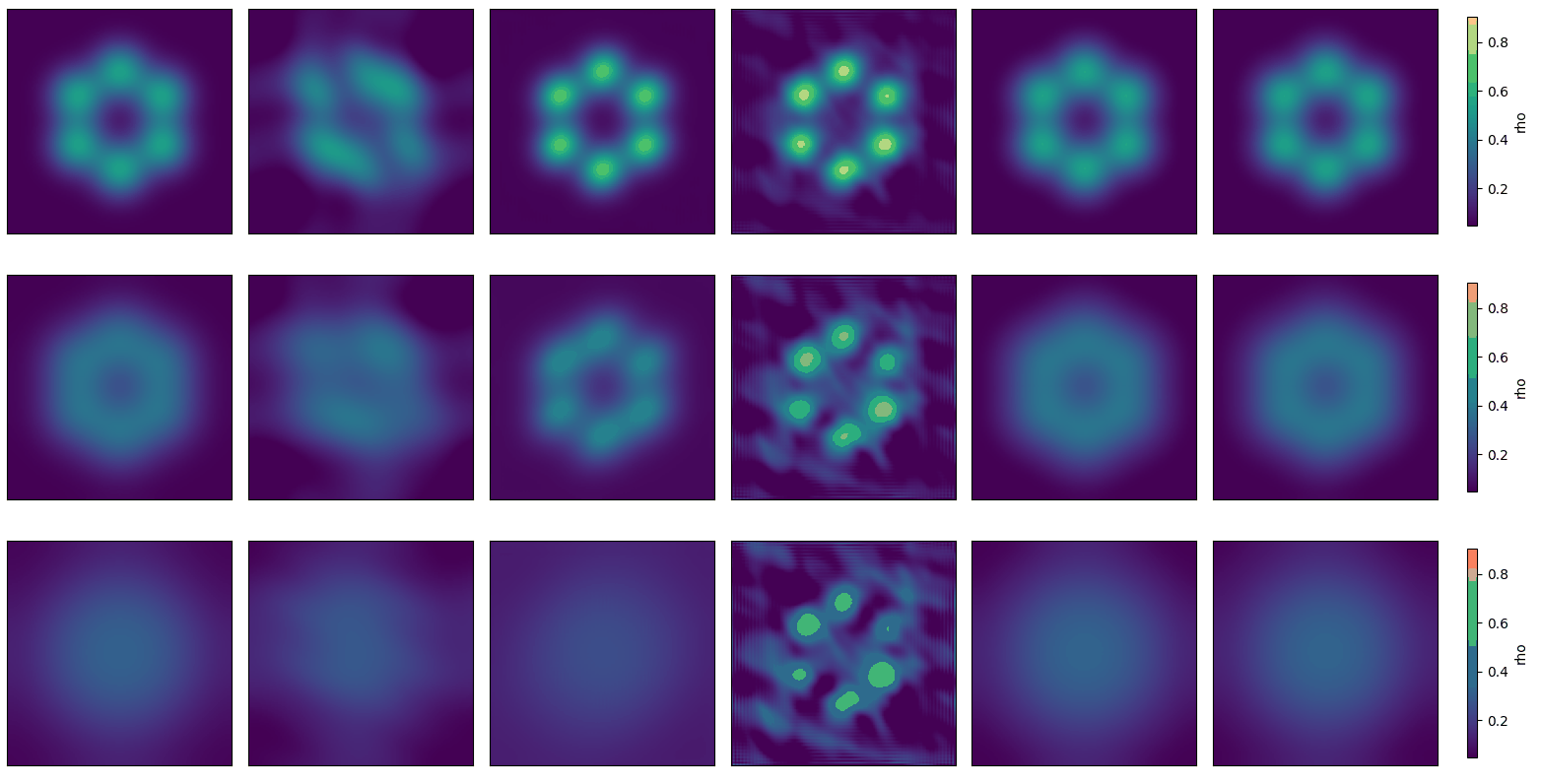}
        \put( 8,51.5){\makebox(0,0){\scriptsize Reference}}
        \put(23,51.5){\makebox(0,0){\scriptsize F-FNO}}
        \put(39,51.5){\makebox(0,0){\scriptsize CFO}}
        \put(55,51.5){\makebox(0,0){\scriptsize DOOL}}
        \put(70,51.5){\makebox(0,0){\scriptsize NCL-MCT (Vel.)}}
        \put(86,51.5){\makebox(0,0){\scriptsize NCL-MCT (Law)}}
        \put(-1,44){\makebox(0,0){\rotatebox{90}{\scriptsize $t=1.5\times10^{-3}$}}}
        \put(-1,26){\makebox(0,0){\rotatebox{90}{\scriptsize $t=4.5\times10^{-3}$}}}
        \put(-1,9){\makebox(0,0){\rotatebox{90}{\scriptsize $t=1.5\times10^{-2}$}}}
\end{overpic}
\caption{
Linear diffusion from six Gaussian blobs on a circle, \eqnref{app_ood_blobs},
600-step rollout to $T=0.03$.
F-FNO loses the six-fold symmetry early and replaces the ring by two smeared diagonal
lobes, while CFO retains separated blobs after the reference has largely merged them.
DOOL shows distinct blobs on a striped background where the reference is nearly uniform.
Both \NCLMCT configurations more closely follow the merging and flattening dynamics.
}
\label{fig:ood_diffusion}
\end{figure}

\begin{figure}[htbp]
\centering
\begin{overpic}[width=0.9\linewidth]{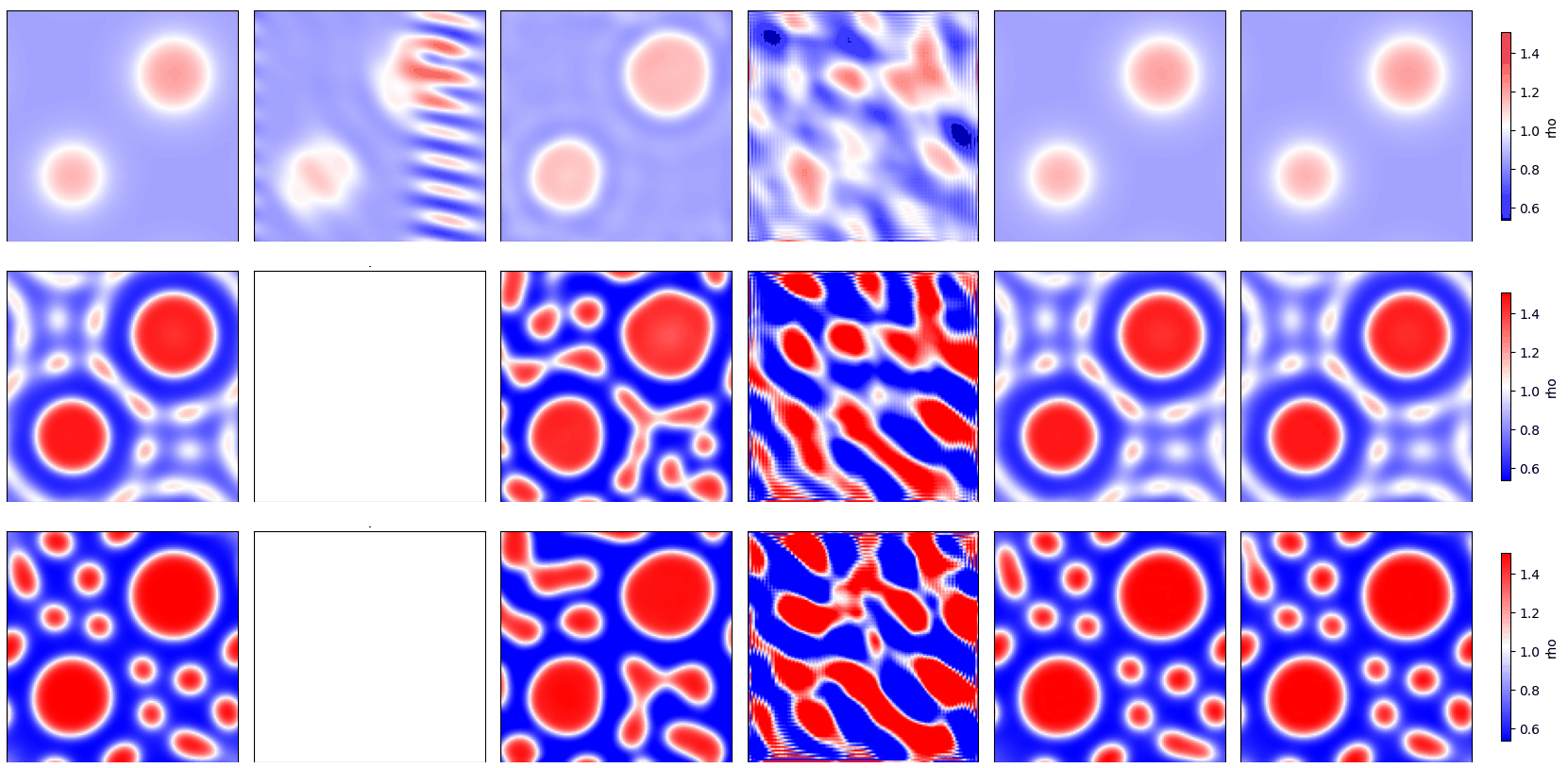}
        \put(8,50){\makebox(0,0){\scriptsize Reference}}
        \put(23,50){\makebox(0,0){\scriptsize F-FNO}}
        \put(40,50){\makebox(0,0){\scriptsize CFO}}
        \put(55,50){\makebox(0,0){\scriptsize DOOL}}
        \put(70,50){\makebox(0,0){\scriptsize NCL-MCT (Vel.)}}
        \put(86,50){\makebox(0,0){\scriptsize NCL-MCT (Law)}}
        \put(-1,43){\makebox(0,0){\rotatebox{90}{\scriptsize $t=0.01$}}}
        \put(-1,26){\makebox(0,0){\rotatebox{90}{\scriptsize $t=0.05$}}}
        \put(-1,9){\makebox(0,0){\rotatebox{90}{\scriptsize $t=0.10$}}}
\end{overpic}
\caption{
Cahn-Hilliard from two Gaussian blobs, \eqnref{app_ood_two_blobs},
10{,}000-step rollout to $T=0.1$.
F-FNO develops oscillations near the domain boundary at $t=0.01$ and becomes nonfinite
thereafter.
CFO recovers the two large domains but forms an irregular set of secondary droplets,
while DOOL produces a different labyrinthine pattern.
Both \NCLMCT configurations reproduce the two domains and the regular array of smaller
droplets between them.
}
\label{fig:ood_ch}
\end{figure}

\begin{figure}[htbp]
\centering
\begin{overpic}[width=0.9\linewidth]{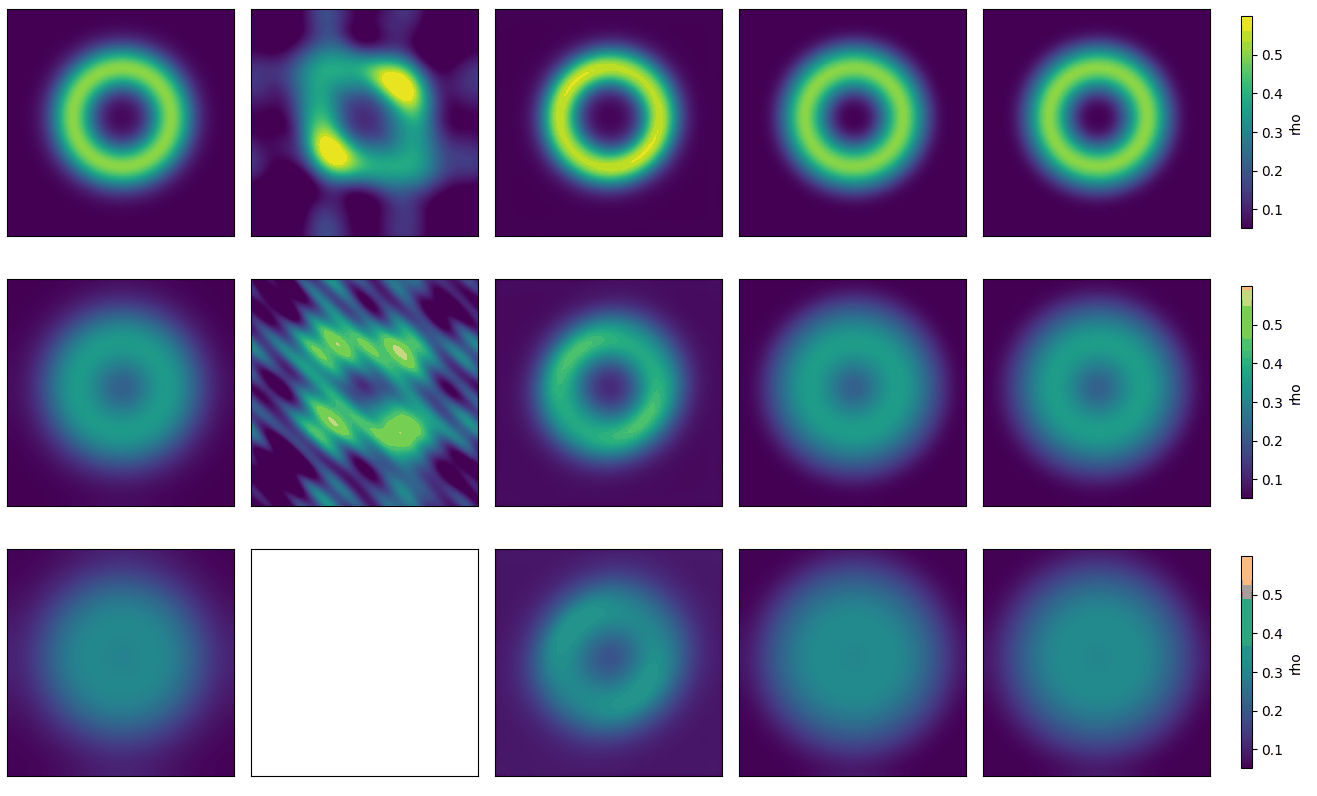}
        \put( 9,61){\makebox(0,0){\scriptsize Reference}}
        \put(28,61){\makebox(0,0){\scriptsize F-FNO}}
        \put(45,61){\makebox(0,0){\scriptsize CFO}}
        \put(64,61){\makebox(0,0){\scriptsize NCL-MCT (Vel.)}}
        \put(83,61){\makebox(0,0){\scriptsize NCL-MCT (Law)}}
        \put(-1,50){\makebox(0,0){\rotatebox{90}{\scriptsize $t=7.5\times10^{-4}$}}}
        \put(-1,30){\makebox(0,0){\rotatebox{90}{\scriptsize $t=4.5\times10^{-3}$}}}
        \put(-1,10){\makebox(0,0){\rotatebox{90}{\scriptsize $t=9\times10^{-3}$}}}
\end{overpic}
\caption{
Fisher-KPP from a Gaussian ring, \eqnref{app_ood_ring},
500-step rollout to $T=0.015$.
The blank panel denotes a nonfinite state.
F-FNO breaks the ring into two bright lobes and develops diagonal striping before
becoming nonfinite, while CFO retains a pronounced central depression after the
reference has filled in.
Both \NCLMCT configurations more closely follow the filling of the ring.
}
\label{fig:ood_fkpp}
\end{figure}

\begin{figure}[htbp]
\centering
\begin{overpic}[width=0.9\linewidth]{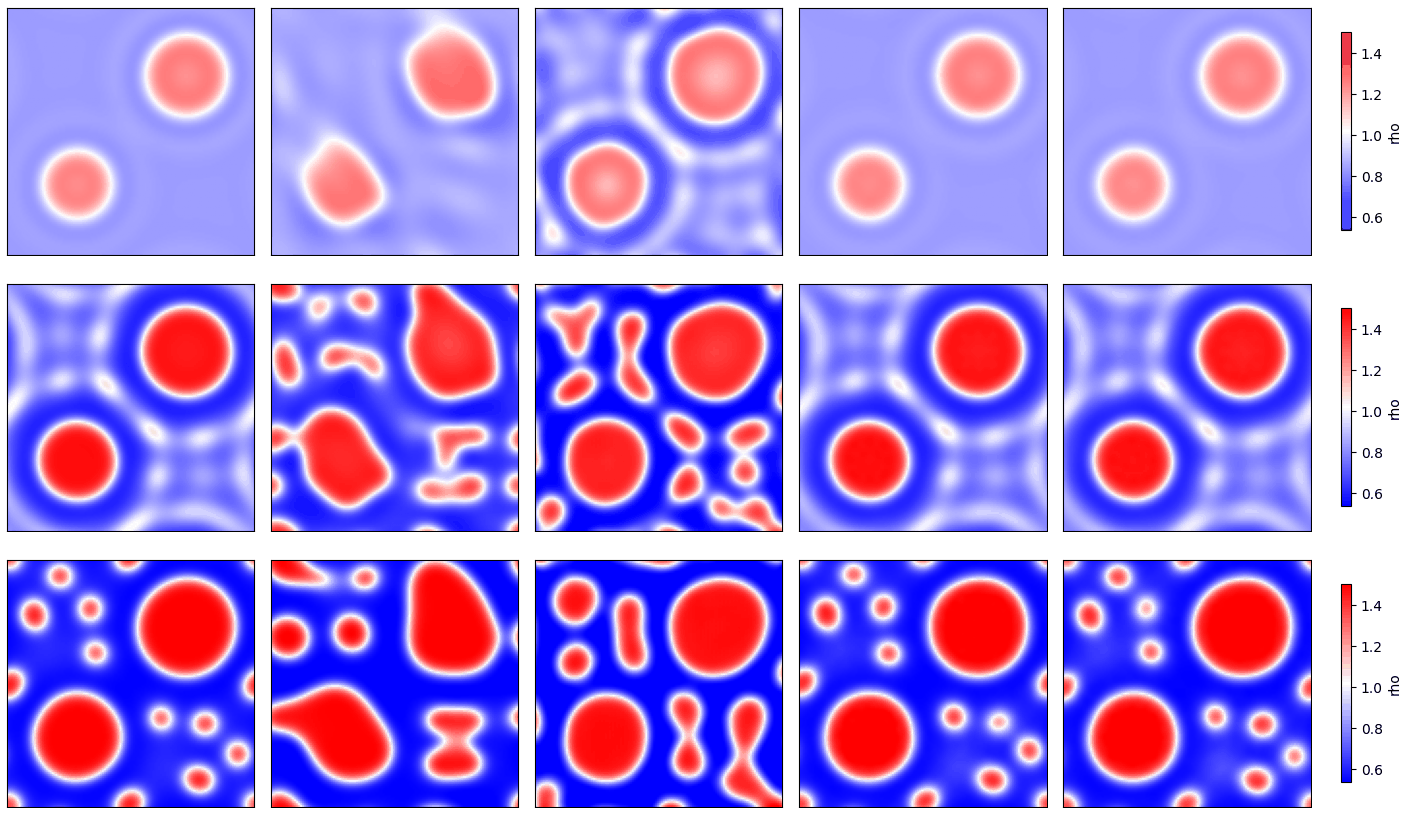}
        \put( 9,58){\makebox(0,0){\scriptsize Reference}}
        \put(28,58){\makebox(0,0){\scriptsize F-FNO}}
        \put(45,58){\makebox(0,0){\scriptsize CFO}}
        \put(64,58){\makebox(0,0){\scriptsize NCL-MCT (Vel.)}}
        \put(83,58){\makebox(0,0){\scriptsize NCL-MCT (Law)}}
        \put(-1,50){\makebox(0,0){\rotatebox{90}{\scriptsize $t=0.02$}}}
        \put(-1,30){\makebox(0,0){\rotatebox{90}{\scriptsize $t=0.05$}}}
        \put(-1,10){\makebox(0,0){\rotatebox{90}{\scriptsize $t=0.10$}}}
\end{overpic}
\caption{
Reactive Cahn-Hilliard from two Gaussian blobs,
\eqnref{app_ood_two_blobs}, 10{,}000-step rollout to $T=0.1$.
F-FNO preserves the large domains but distorts their shape and misses the
secondary-droplet array, while CFO merges the droplets into elongated structures.
Both \NCLMCT configurations recover the array more closely.
}
\label{fig:ood_rch}
\end{figure}

\begin{figure}[htbp]
\centering
\begin{overpic}[width=0.9\linewidth]{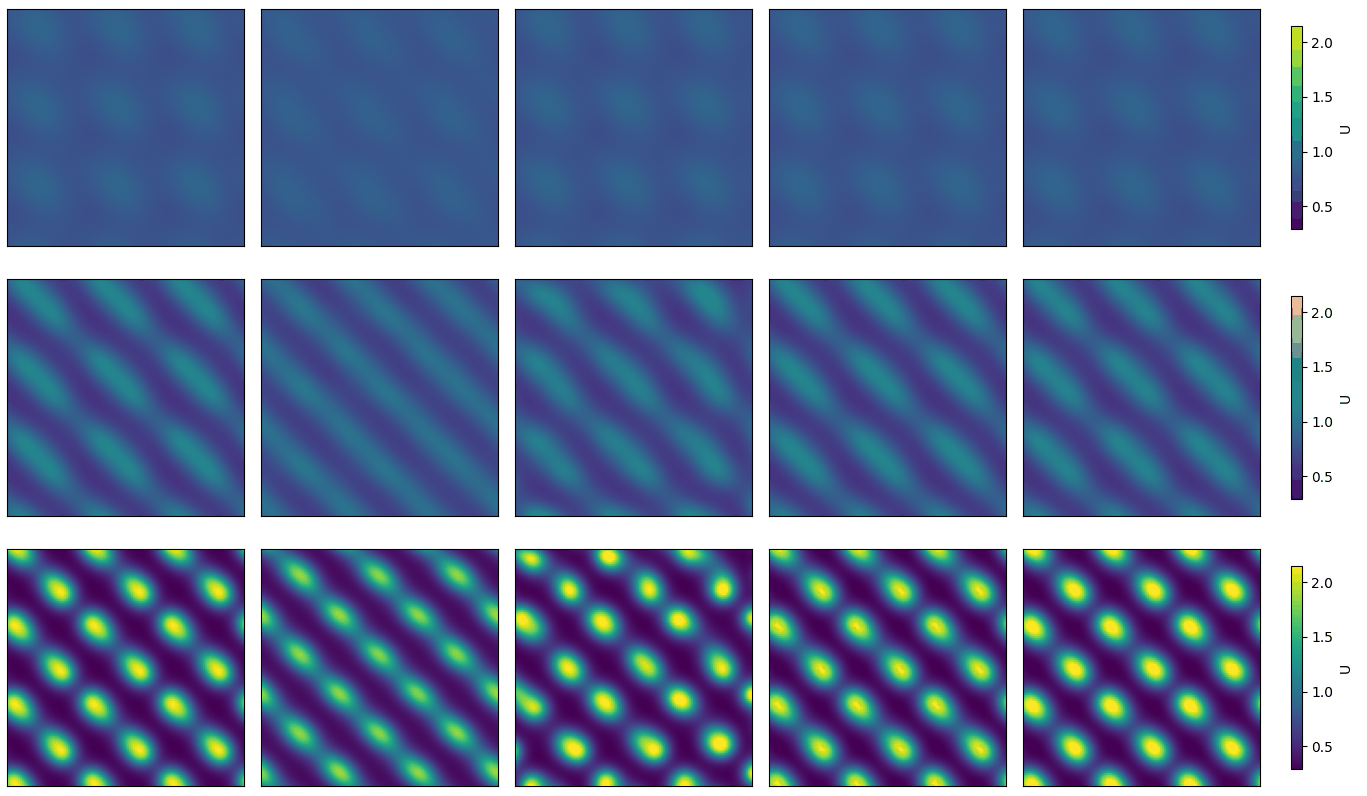}
        \put( 9,58.5){\makebox(0,0){\scriptsize Reference}}
        \put(28,58.5){\makebox(0,0){\scriptsize F-FNO}}
        \put(45,58.5){\makebox(0,0){\scriptsize CFO}}
        \put(64,58.5){\makebox(0,0){\scriptsize NCL-MCT (Vel.)}}
        \put(83,58.5){\makebox(0,0){\scriptsize NCL-MCT (Law)}}
        \put(-1,50){\makebox(0,0){\rotatebox{90}{\scriptsize $t=0.1$}}}
        \put(-1,30){\makebox(0,0){\rotatebox{90}{\scriptsize $t=0.5$}}}
        \put(-1,10){\makebox(0,0){\rotatebox{90}{\scriptsize $t=1.0$}}}
\end{overpic}
\caption{
Schnakenberg from a spot lattice, \eqnref{app_ood_structured}, activator $U$,
20{,}000-step rollout to $T=1$.
All methods remain finite but differ in pattern selection:
F-FNO retains smooth diagonal bands where the reference has formed a spot lattice,
while CFO and both \NCLMCT configurations resolve the spots more closely.
}
\label{fig:ood_schnakenberg_u}
\end{figure}

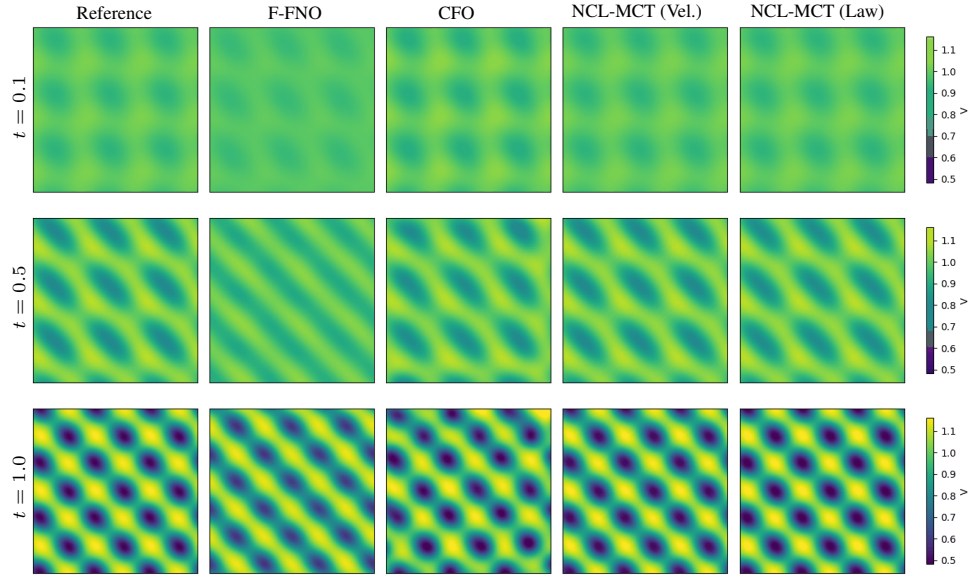
\begin{figure}[htbp]
\centering
\begin{overpic}[width=0.9\linewidth]{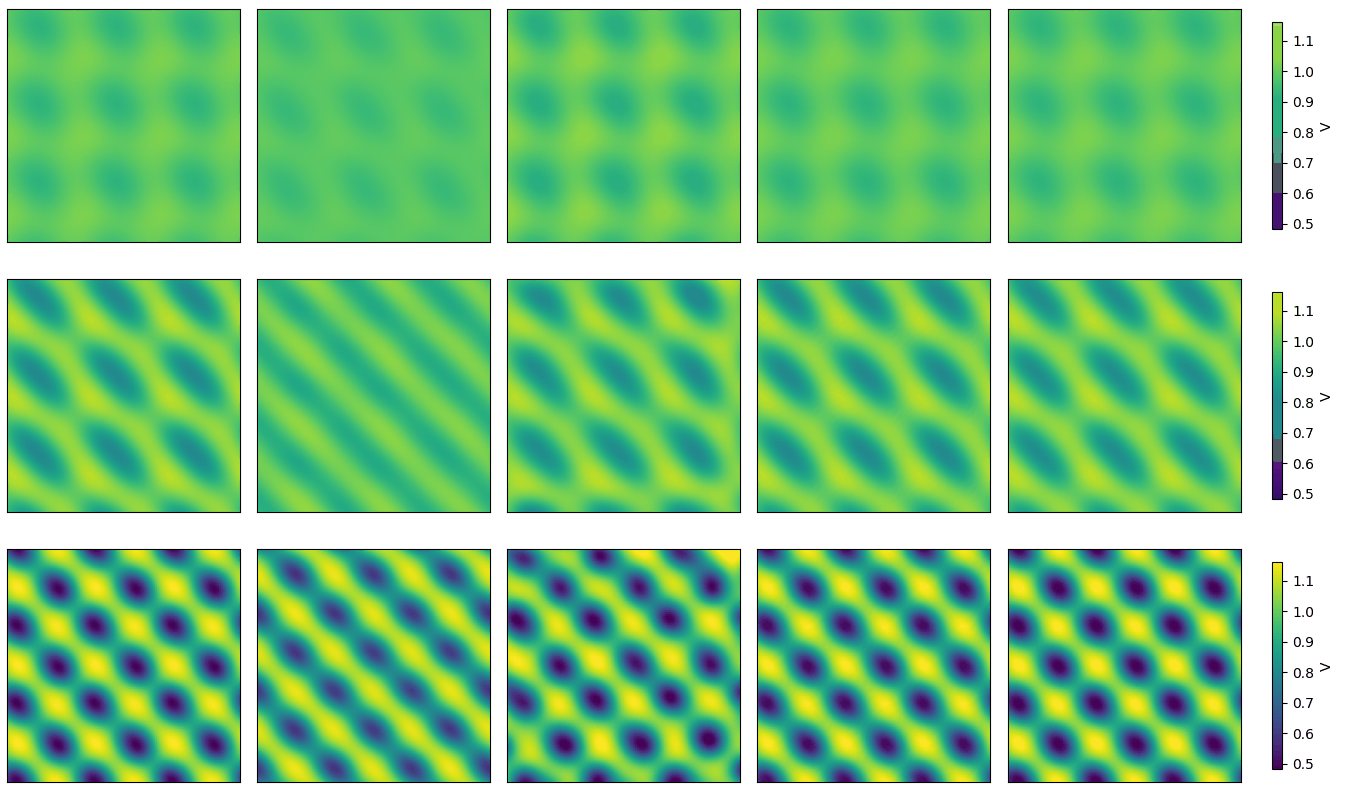}
        \put( 9,60){\makebox(0,0){\scriptsize Reference}}
        \put(28,60){\makebox(0,0){\scriptsize F-FNO}}
        \put(45,60){\makebox(0,0){\scriptsize CFO}}
        \put(64,60){\makebox(0,0){\scriptsize NCL-MCT (Vel.)}}
        \put(83,60){\makebox(0,0){\scriptsize NCL-MCT (Law)}}
        \put(-1,50){\makebox(0,0){\rotatebox{90}{\scriptsize $t=0.1$}}}
        \put(-1,30){\makebox(0,0){\rotatebox{90}{\scriptsize $t=0.5$}}}
        \put(-1,10){\makebox(0,0){\rotatebox{90}{\scriptsize $t=1.0$}}}
\end{overpic}
\caption{
Schnakenberg from a spot lattice, inhibitor $V$, with the same rollout and layout as
Figure~\ref{fig:ood_schnakenberg_u}.
The F-FNO banding is more visible for this species.
}
\label{fig:ood_schnakenberg_v}
\end{figure}

\paragraph{Experimental settings.}
We test five systems: linear diffusion, CH, Fisher-KPP, reactive CH, and Schnakenberg.
For each system, we replace the initial-condition family of
Appendix~\ref{app:ic_families} with a structured family absent from training, while
keeping the PDE parameters, domain, $128\times128$ grid, reference solver, and time
interval unchanged.
The training distributions contain smooth Gaussian Fourier random fields for four
systems and random perturbations of the homogeneous steady state for Schnakenberg,
whereas the out-of-distribution families in Table~\ref{tab:ic_ood_families} impose
explicit spatial structure, including blobs, rings, and spot lattices.
This setting tests whether constitutive responses learned from one distribution of
density fields remain accurate under a different spatial organization.
Each system uses one fixed out-of-distribution initial condition, so the reported
results are single rollouts rather than trajectory averages.
Figures~\ref{fig:ood_diffusion}-\ref{fig:ood_schnakenberg_v} show the corresponding
rollouts, with columns ordered as reference, F-FNO, CFO, DOOL where applicable,
\NCLMCT{} (Vel.), and \NCLMCT{} (Law), and time increasing downward.

\begin{table}[b]
\centering
\small
\setlength{\tabcolsep}{4pt}
\caption{
Out-of-distribution initial-condition families.
Distances are periodic on each system's domain, and all fields are evaluated on the same
$128\times128$ grid and rolled out over the same interval as the corresponding
in-distribution test.
}
\label{tab:ic_ood_families}
\begin{tabular}{lll}
\toprule
System & Family & $N_T$ \\
\midrule
Linear Diffusion & Six Gaussian blobs on a circle, \eqnref{app_ood_blobs} & 600 \\
CH & Two Gaussian blobs, \eqnref{app_ood_two_blobs} & $10^{4}$ \\
Fisher-KPP & Gaussian ring, \eqnref{app_ood_ring} & 500 \\
Reactive CH & Two Gaussian blobs, \eqnref{app_ood_two_blobs} & $10^{4}$ \\
Schnakenberg & Spot lattice, \eqnref{app_ood_structured} & $2\times10^{4}$ \\
\bottomrule
\end{tabular}
\end{table}

\paragraph{Out-of-distribution initial conditions.}
For linear diffusion, six Gaussian blobs are placed at equal angles on a circle,
\begin{equation}
\rho_0(\mathbf x)=\rho_b+A\sum_{k=0}^{5}
\exp\!\Bigl(-\tfrac{\lvert\mathbf x-\mathbf c_k\rvert^2}{2\sigma^2}\Bigr),
\qquad
\rho_b=0.05,\;A=0.85,\;\sigma=0.06,
\label{eq:app_ood_blobs}
\end{equation}
with $\mathbf c_k=(0.5,0.5)+0.22(\cos\theta_k,\sin\theta_k)$,
$\theta_k=2\pi k/6$, clipped to $[0,1]$.
The two CH systems use two blobs of unequal size on a uniform background,
\begin{equation}
\rho_0(\mathbf x)=0.85
+0.24\exp\!\Bigl(-\tfrac{\lvert\mathbf x-(0.28,0.28)\rvert^2}{2\cdot0.12^2}\Bigr)
+0.30\exp\!\Bigl(-\tfrac{\lvert\mathbf x-(0.72,0.72)\rvert^2}{2\cdot0.13^2}\Bigr),
\label{eq:app_ood_two_blobs}
\end{equation}
clipped to $[0.85,1.15]$, so phase separation starts from two prescribed nuclei rather
than a random field.
The Fisher-KPP family is a Gaussian ring with a jittered center and width,
\begin{equation}
\rho_0(\mathbf x)=\rho_b+A
\exp\!\Bigl(-\tfrac{(\lvert\mathbf x-\mathbf c\rvert-R)^2}{2\sigma^2}\Bigr),
\qquad
\rho_b=0.05,\;A=0.55,\;R=0.22,
\label{eq:app_ood_ring}
\end{equation}
with $\mathbf c=(0.5,0.5)+\boldsymbol\delta$,
$\boldsymbol\delta\sim\mathcal N(0,0.02^2I)$,
$\sigma=0.06+\eta$, and $\eta\sim\mathcal N(0,0.005^2)$, clipped to
$[10^{-6},0.95]$.
Unlike the training fields, it contains an interior minimum and a single radial length scale.
The Schnakenberg family perturbs the homogeneous steady state with a normalized
hexagonal spot lattice,
\begin{equation}
C_0=C^\ast\bigl(1+\varepsilon_C\,g\bigr),
\qquad
g\propto\cos\theta_x+\cos\theta_y+\cos(\theta_x+\theta_y),
\qquad C\in\{U,V\},
\label{eq:app_ood_structured}
\end{equation}
with $\theta_x=2\pi kx+\varphi_x$, $\theta_y=2\pi ky+\varphi_y$,
$k\sim\mathcal U\{1,2,3\}$, and independent uniform phases.
The field $g$ is made mean-free and normalized to unit maximum magnitude.
The relative amplitudes are restricted to the lower $40\%$ of the training range,
$\varepsilon_U\in[0.04,0.064]$ and $\varepsilon_V\in[0.12,0.16]$.
The candidate is accepted only if $U$, $V$, and $U^2V$ remain finite and within the
range spanned by the training snapshots throughout the rollout, restricting the shift
primarily to spatial organization rather than amplitude.

\begin{table}[t]
    \centering
    \footnotesize
    \setlength{\tabcolsep}{3pt}
    \renewcommand{\arraystretch}{1.15}
    \caption{
        Rollout errors from unseen initial conditions.
        Each system uses one fixed out-of-distribution initial condition, so entries are
        single rollouts without standard deviations.
        $E_{\mathrm{roll}}$ and $E_{\mathrm{max}}$ denote space-time and maximum per-time
        relative $L^2$ errors.
        Red/orange mark the best/second-best result per row.
        NaN denotes a nonfinite rollout; $-$ denotes a method not evaluated under the
        adopted formulation.
        Schnakenberg $U$ and $V$ count as one system.
    }
    \label{tab:ic_ood_results}

    \resizebox{\linewidth}{!}{%
    \begin{tabular}{@{}lclccccc@{}}
        \toprule
        System & $N_T$ & Metric
        & \shortstack{F-FNO\\\citep{tran2023ffno}}
        & \shortstack{CFO\\\citep{hou2026cfo}}
        & \shortstack{DOOL\\\citep{chang2025dool}}
        & \shortstack{NCL-MCT (Vel.)\\(Ours)}
        & \shortstack{NCL-MCT (Law)\\(Ours)} \\
        \midrule

        \multirow{2}{*}{Linear Diffusion}
        & \multirow{2}{*}{$600$}
        & $E_{\mathrm{roll}}$
        & $1.71\times10^{-1}$
        & $1.65\times10^{-1}$
        & $5.74\times10^{-1}$
        & \cms $4.07\times10^{-2}$
        & \cmf $3.81\times10^{-2}$ \\
        & & $E_{\mathrm{max}}$
        & $2.87\times10^{-1}$
        & $2.13\times10^{-1}$
        & $7.01\times10^{-1}$
        & \cms $4.67\times10^{-2}$
        & \cmf $4.44\times10^{-2}$ \\
        \midrule

        \multirow{2}{*}{CH}
        & \multirow{2}{*}{$10{,}000$}
        & $E_{\mathrm{roll}}$
        & NaN
        & $2.31\times10^{-1}$
        & $4.28\times10^{-1}$
        & \cms $1.96\times10^{-2}$
        & \cmf $1.83\times10^{-2}$ \\
        & & $E_{\mathrm{max}}$
        & NaN
        & $3.30\times10^{-1}$
        & $5.33\times10^{-1}$
        & \cms $3.63\times10^{-2}$
        & \cmf $3.32\times10^{-2}$ \\
        \midrule

        \multirow{2}{*}{Fisher-KPP}
        & \multirow{2}{*}{$500$}
        & $E_{\mathrm{roll}}$
        & NaN
        & $1.35\times10^{-1}$
        & $-$
        & \cms $5.71\times10^{-2}$
        & \cmf $5.36\times10^{-2}$ \\
        & & $E_{\mathrm{max}}$
        & NaN
        & $1.71\times10^{-1}$
        & $-$
        & \cms $7.02\times10^{-2}$
        & \cmf $6.57\times10^{-2}$ \\
        \midrule

        \multirow{2}{*}{Reactive CH}
        & \multirow{2}{*}{$10{,}000$}
        & $E_{\mathrm{roll}}$
        & $2.31\times10^{-1}$
        & $2.12\times10^{-1}$
        & $-$
        & \cmf $1.97\times10^{-2}$
        & \cms $2.03\times10^{-2}$ \\
        & & $E_{\mathrm{max}}$
        & $3.27\times10^{-1}$
        & $2.96\times10^{-1}$
        & $-$
        & \cms $4.30\times10^{-2}$
        & \cmf $4.07\times10^{-2}$ \\
        \midrule

        \multirow{2}{*}{Schnakenberg ($U$)}
        & \multirow{4}{*}{$20{,}000$}
        & $E_{\mathrm{roll}}$
        & $2.13\times10^{-1}$
        & $1.51\times10^{-1}$
        & $-$
        & \cmf $1.94\times10^{-2}$
        & \cms $1.96\times10^{-2}$ \\
        & & $E_{\mathrm{max}}$
        & $4.15\times10^{-1}$
        & $2.68\times10^{-1}$
        & $-$
        & \cmf $3.69\times10^{-2}$
        & \cms $6.07\times10^{-2}$ \\
        \addlinespace[3pt]

        \multirow{2}{*}{Schnakenberg ($V$)}
        & & $E_{\mathrm{roll}}$
        & $6.89\times10^{-2}$
        & $4.83\times10^{-2}$
        & $-$
        & \cmf $5.22\times10^{-3}$
        & \cms $5.37\times10^{-3}$ \\
        & & $E_{\mathrm{max}}$
        & $1.40\times10^{-1}$
        & $8.88\times10^{-2}$
        & $-$
        & \cmf $1.03\times10^{-2}$
        & \cms $1.48\times10^{-2}$ \\
        \bottomrule
    \end{tabular}%
    }
\end{table}

\paragraph{Results.}
Table~\ref{tab:ic_ood_results} reports the five rollouts.
\NCLMCT achieves the lowest error on all five systems and both metrics, with
$E_{\mathrm{roll}}$ between $2.4\times$ and $13\times$ below the strongest finite
baseline.
The two supervision modes remain within $10\%$ of each other in
$E_{\mathrm{roll}}$ across all five systems, although their $E_{\mathrm{max}}$ differs
more on Schnakenberg.
Linear diffusion shows the largest change in ranking.
F-FNO, the most accurate method in distribution, increases its error by roughly
$260\times$ to $1.71\times10^{-1}$, whereas \NCLMCT increases by
$24$-$41\times$ and remains four to six times lower.
F-FNO also becomes nonfinite on CH and Fisher-KPP.
The relative degradation is not uniformly smaller for \NCLMCT: on Fisher-KPP, its
error increases by about two orders of magnitude, compared with $23\times$ for CFO.
Thus, lower out-of-distribution error does not imply uniformly smaller relative
degradation from the in-distribution setting.
Schnakenberg behaves differently: all methods except CFO are at least as accurate as
in distribution.
The imposed spot lattice is geometrically unseen but resembles the pattern selected by
the Turing dynamics, making this primarily a structural-transfer test rather than a
uniformly harder initial condition.

%%%%%%%%%%%%%%%%%%%%%%%%%%%%%%%%%%%%%%%%%%%%%%%%%%%%%%%%%%%%%%%%%%%%%%%%%%%

\clearpage

%%%%%%%%%%%%%%%%%%%%%%%%%%%%%%%%%%%%%%%%%%%%%%%%%%%%%%%%%%%%%%%%%%%%%%%%%%%
\clearpage

\subsection{Extended Time Horizon}
\label{app:extended_horizon}

\paragraph{Experimental settings.}
For linear diffusion, CH, Fisher-KPP, and reactive CH, models are trained on
trajectories truncated to the first $60\%$ of the reference interval and evaluated
over the full interval in Table~\ref{tab:system_overview}, giving an extrapolation
factor of $1/0.6\approx1.67$.
The step size, PDE parameters, and ten in-distribution test initial conditions remain
unchanged, so the final $40\%$ of each rollout lies beyond the training horizon.
Table~\ref{tab:horizons} lists the training and evaluation horizons.

\begin{table}[b]
\centering
\small
\setlength{\tabcolsep}{6pt}
\caption{
Training and evaluation horizons.
$N_T^{\mathrm{train}}$ and $N_T$ denote the rollout lengths of the truncated training
trajectories and full evaluation, respectively.
The step size is unchanged and equals the reference $\Delta t$ in
Table~\ref{tab:system_overview}.
}
\label{tab:horizons}
\begin{tabular}{lcccc}
\toprule
System & $\Delta t$ & $N_T^{\mathrm{train}}$ ($T_{\mathrm{train}}$)
       & $N_T$ ($T$) & Extrapolated fraction \\
\midrule
Linear diffusion & $5\times10^{-5}$ & $360$ ($0.018$) & $600$ ($0.03$) & $40\%$ \\
CH               & $10^{-5}$        & $6{,}000$ ($0.06$) & $10{,}000$ ($0.1$) & $40\%$ \\
Fisher-KPP      & $3\times10^{-5}$ & $300$ ($0.009$) & $500$ ($0.015$) & $40\%$ \\
Reactive CH      & $10^{-5}$        & $6{,}000$ ($0.06$) & $10{,}000$ ($0.1$) & $40\%$ \\
\bottomrule
\end{tabular}
\end{table}

\begin{table}[b!]
    \centering
    \footnotesize
    \setlength{\tabcolsep}{3pt}
    \renewcommand{\arraystretch}{1.15}
    \caption{
        Rollout errors with the horizon extended beyond training.
        Models are trained on the first $60\%$ of each reference interval and evaluated
        over the full interval (Table~\ref{tab:horizons}).
        $E_{\mathrm{roll}}$ and $E_{\mathrm{max}}$ are mean $\pm$ standard deviation over
        ten test trajectories.
        $\bar E(T_{\mathrm{train}})$ is the trajectory-averaged per-time error at the
        training horizon, and $\Delta E$ is its change to the final time,
        \eqnref{app_delta_error}.
        Red/orange mark the best/second-best mean per row.
    }
    \label{tab:temporal_results}

    \resizebox{\linewidth}{!}{%
    \begin{tabular}{@{}llcccc@{}}
        \toprule
        System & Metric
        & \shortstack{F-FNO\\\citep{tran2023ffno}}
        & \shortstack{CFO\\\citep{hou2026cfo}}
        & \shortstack{NCL-MCT (Vel.)\\(Ours)}
        & \shortstack{NCL-MCT (Law)\\(Ours)} \\
        \midrule

        \multirow{4}{*}{Linear Diffusion}
        & $E_{\mathrm{roll}}$
        & \cmf \errstat{4.01}{1.26}{-4}
        & \errstat{5.13}{1.04}{-2}
        & \cms \errstat{1.44}{0.27}{-3}
        & \errstat{2.26}{0.37}{-3} \\
        & $E_{\mathrm{max}}$
        & \cmf \errstat{6.09}{1.87}{-4}
        & \errstat{6.25}{1.52}{-2}
        & \cms \errstat{1.62}{0.34}{-3}
        & \errstat{2.60}{0.46}{-3} \\
        & $\bar E(T_{\mathrm{train}})$
        & \cmf $3.78\times10^{-4}$
        & $5.82\times10^{-2}$
        & \cms $1.58\times10^{-3}$
        & $2.53\times10^{-3}$ \\
        & $\Delta E$
        & $1.69\times10^{-4}$
        & $2.09\times10^{-3}$
        & \cms $-5.72\times10^{-5}$
        & \cmf $-1.75\times10^{-4}$ \\
        \midrule

        \multirow{4}{*}{CH}
        & $E_{\mathrm{roll}}$
        & \errstat{1.39}{0.17}{-1}
        & \errstat{9.96}{0.86}{-2}
        & \cmf \errstat{9.27}{3.03}{-3}
        & \cms \errstat{1.03}{0.27}{-2} \\
        & $E_{\mathrm{max}}$
        & \errstat{1.85}{0.24}{-1}
        & \errstat{1.26}{0.13}{-1}
        & \cmf \errstat{1.44}{0.89}{-2}
        & \cms \errstat{1.49}{0.37}{-2} \\
        & $\bar E(T_{\mathrm{train}})$
        & $1.59\times10^{-1}$
        & $1.10\times10^{-1}$
        & \cmf $9.69\times10^{-3}$
        & \cms $1.17\times10^{-2}$ \\
        & $\Delta E$
        & $2.60\times10^{-2}$
        & $6.70\times10^{-3}$
        & \cms $4.51\times10^{-3}$
        & \cmf $2.57\times10^{-3}$ \\
        \midrule

        \multirow{4}{*}{Fisher-KPP}
        & $E_{\mathrm{roll}}$
        & \cms \errstat{5.96}{1.36}{-4}
        & \errstat{3.10}{0.41}{-2}
        & \cmf \errstat{5.02}{0.82}{-4}
        & \errstat{7.48}{0.64}{-4} \\
        & $E_{\mathrm{max}}$
        & \cms \errstat{8.02}{1.72}{-4}
        & \errstat{4.15}{0.68}{-2}
        & \cmf \errstat{5.81}{1.09}{-4}
        & \errstat{9.64}{0.96}{-4} \\
        & $\bar E(T_{\mathrm{train}})$
        & \cms $6.32\times10^{-4}$
        & $3.45\times10^{-2}$
        & \cmf $5.50\times10^{-4}$
        & $8.32\times10^{-4}$ \\
        & $\Delta E$
        & $1.67\times10^{-4}$
        & $6.88\times10^{-3}$
        & \cmf $2.37\times10^{-5}$
        & \cms $1.32\times10^{-4}$ \\
        \midrule

        \multirow{4}{*}{Reactive CH}
        & $E_{\mathrm{roll}}$
        & \errstat{1.30}{0.22}{-1}$^{*}$
        & \errstat{9.36}{0.71}{-2}
        & \cms \errstat{1.22}{0.40}{-2}
        & \cmf \errstat{8.69}{1.59}{-3} \\
        & $E_{\mathrm{max}}$
        & \errstat{1.72}{0.27}{-1}$^{*}$
        & \errstat{1.18}{0.12}{-1}
        & \cms \errstat{1.96}{0.81}{-2}
        & \cmf \errstat{1.38}{0.32}{-2} \\
        & $\bar E(T_{\mathrm{train}})$
        & $1.48\times10^{-1}$$^{*}$
        & $1.04\times10^{-1}$
        & \cms $1.21\times10^{-2}$
        & \cmf $8.32\times10^{-3}$ \\
        & $\Delta E$
        & $2.36\times10^{-2}$$^{*}$
        & \cmf $3.61\times10^{-3}$
        & $6.56\times10^{-3}$
        & \cms $5.36\times10^{-3}$ \\
        \bottomrule
        \addlinespace[2pt]
        \multicolumn{6}{l}{\footnotesize $^{*}$F-FNO becomes nonfinite on the seventh
        reactive CH trajectory after $400$ steps, within the training horizon;}\\
        \multicolumn{6}{l}{\footnotesize \phantom{$^{*}$}its reported statistics use the
        nine remaining trajectories.}
    \end{tabular}%
    }
\end{table}

\paragraph{Metrics.}
$E_{\mathrm{roll}}$ and $E_{\mathrm{max}}$ are computed over the full evaluation
window.
To separate extrapolation error from the error already present at the training horizon,
we evaluate the trajectory-averaged per-time error
$\bar E(t)=\frac{1}{N}\sum_i E_i(t)$ and report
\begin{equation}
\Delta E=\bar E(T)-\bar E(T_{\mathrm{train}}),
\label{eq:app_delta_error}
\end{equation}
which measures the additional error from the training horizon to the final time.
We report the difference rather than a ratio because the errors at
$T_{\mathrm{train}}$ differ substantially across methods.

\paragraph{Results.}
Relative to Table~\ref{tab:reaction_diffusion}, the errors of both \NCLMCT
configurations change by factors of $0.90$-$1.46$, so evaluation to
$1.67\times$ the shortened training horizon increases error by at most about $50\%$
(Table~\ref{tab:temporal_results}).
CFO errors increase by $1.6\times$ on the two Cahn-Hilliard systems and by
$5.2\times$ on linear diffusion and Fisher-KPP.
F-FNO changes little overall, remaining strongest on linear diffusion and second on
Fisher-KPP but about an order of magnitude less accurate than \NCLMCT on the two
Cahn-Hilliard systems.
At $T_{\mathrm{train}}$, \NCLMCT is already about an order of magnitude below both
baselines on the two Cahn-Hilliard systems.
Its $\Delta E$ is also smallest on linear diffusion, CH, and Fisher-KPP; only on
reactive CH does CFO show smaller growth, despite an error at
$T_{\mathrm{train}}$ that is twelve times larger.
On linear diffusion, $\Delta E<0$ for both \NCLMCT configurations, indicating that
the final relative error is lower than at the training horizon.

%%%%%%%%%%%%%%%%%%%%%%%%%%%%%%%%%%%%%%%%%%%%%%%%%%%%%%%%%%%%%%%%%%%%%%%%%%%
\begin{figure}[htbp]
\centering
\begin{overpic}[width=0.85\linewidth]{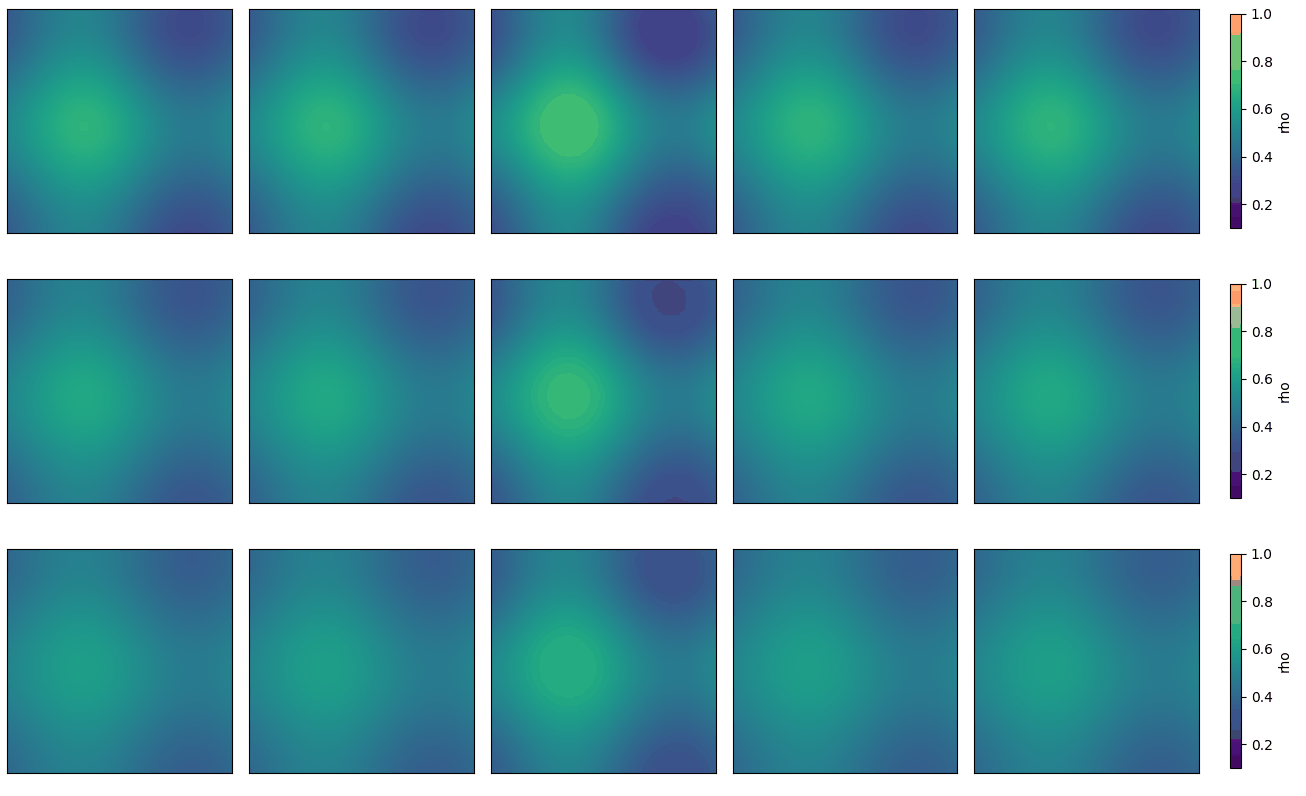}
        \put(10,62){\makebox(0,0){\scriptsize Reference}}
        \put(28,62){\makebox(0,0){\scriptsize F-FNO}}
        \put(47,62){\makebox(0,0){\scriptsize CFO}}
        \put(65,62){\makebox(0,0){\scriptsize NCL-MCT (Vel.)}}
        \put(84,62){\makebox(0,0){\scriptsize NCL-MCT (Law)}}
        \put(-1,52){\makebox(0,0){\rotatebox{90}{\scriptsize $t=0.018$}}}
        \put(-1,32){\makebox(0,0){\rotatebox{90}{\scriptsize $t=0.024$}}}
        \put(-1,10){\makebox(0,0){\rotatebox{90}{\scriptsize $t=0.030$}}}
\end{overpic}
\caption{
Linear diffusion, trained to $T_{\mathrm{train}}=0.018$ and rolled out to $T=0.03$.
Only the first row lies within the training horizon.
}
\label{fig:te_diffusion}
\end{figure}

\begin{figure}[htbp]
\centering
\begin{overpic}[width=0.85\linewidth]{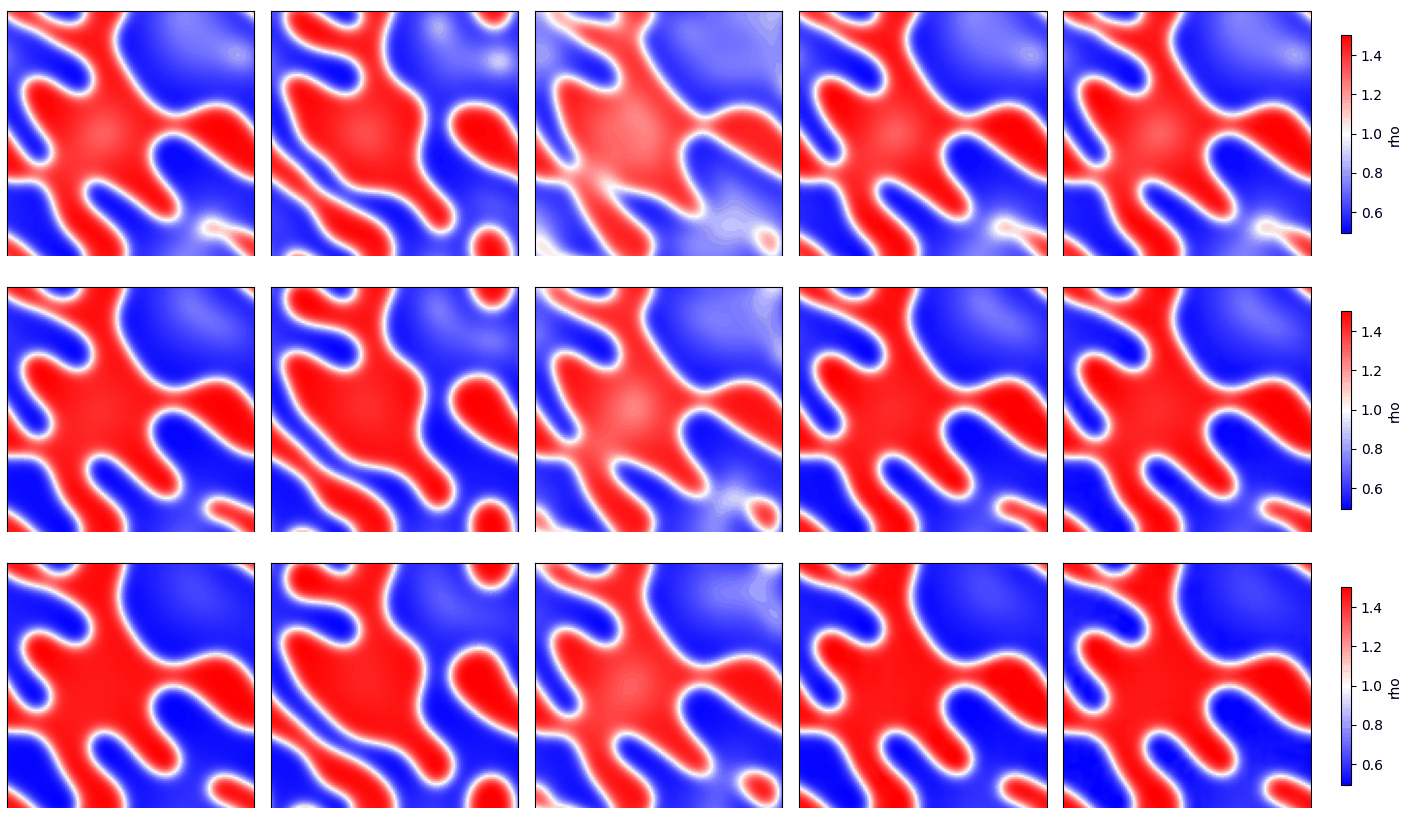}
        \put(10,58){\makebox(0,0){\scriptsize Reference}}
        \put(28,58){\makebox(0,0){\scriptsize F-FNO}}
        \put(47,58){\makebox(0,0){\scriptsize CFO}}
        \put(65,58){\makebox(0,0){\scriptsize NCL-MCT (Vel.)}}
        \put(84,58){\makebox(0,0){\scriptsize NCL-MCT (Law)}}
        \put(-1,50){\makebox(0,0){\rotatebox{90}{\scriptsize $t=0.06$}}}
        \put(-1,30){\makebox(0,0){\rotatebox{90}{\scriptsize $t=0.08$}}}
        \put(-1,10){\makebox(0,0){\rotatebox{90}{\scriptsize $t=0.1$}}}
\end{overpic}
\caption{
Cahn-Hilliard, trained to $T_{\mathrm{train}}=0.06$ and rolled out to $T=0.1$.
}
\label{fig:te_ch}
\end{figure}

\clearpage

\begin{figure}[htbp]
\centering
\begin{overpic}[width=0.85\linewidth]{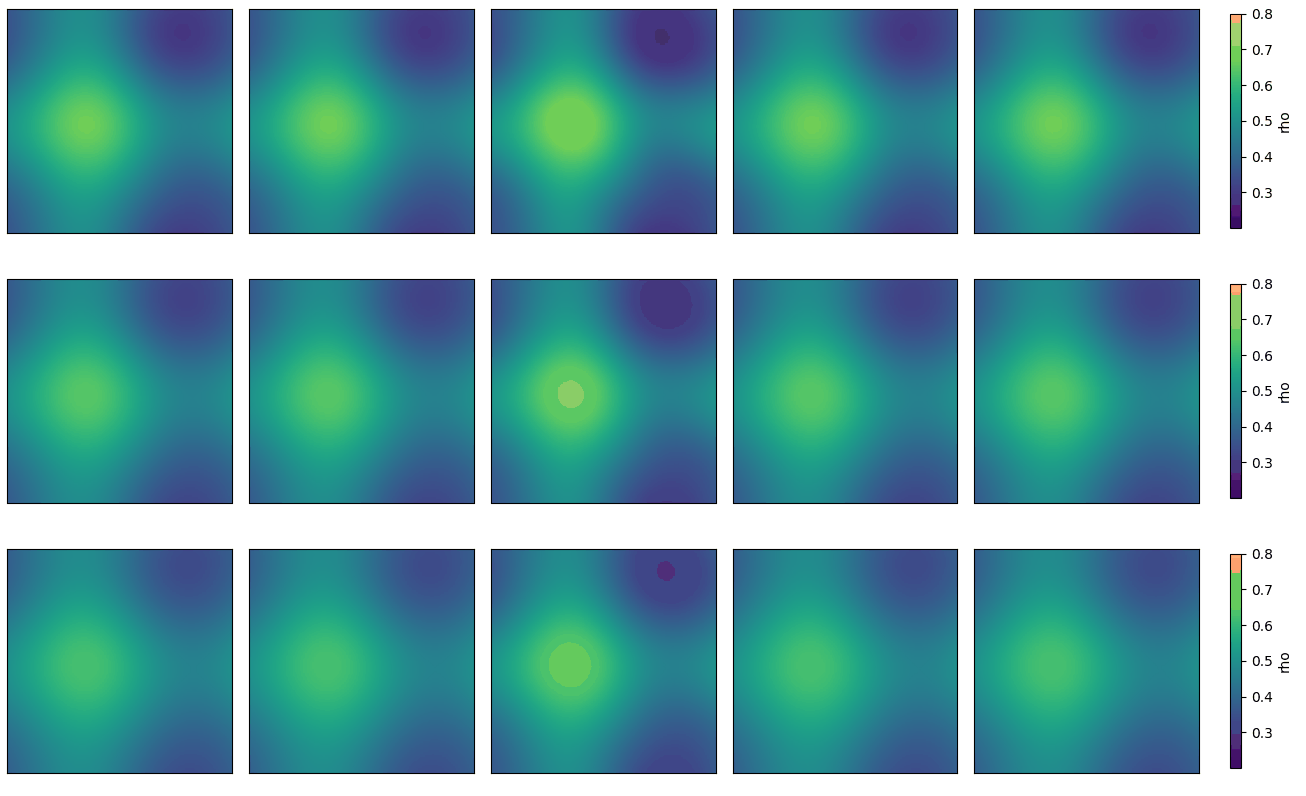}
        \put(10,62){\makebox(0,0){\scriptsize Reference}}
        \put(28,62){\makebox(0,0){\scriptsize F-FNO}}
        \put(47,62){\makebox(0,0){\scriptsize CFO}}
        \put(65,62){\makebox(0,0){\scriptsize NCL-MCT (Vel.)}}
        \put(84,62){\makebox(0,0){\scriptsize NCL-MCT (Law)}}
        \put(-1,52){\makebox(0,0){\rotatebox{90}{\scriptsize $t=0.009$}}}
        \put(-1,32){\makebox(0,0){\rotatebox{90}{\scriptsize $t=0.012$}}}
        \put(-1,10){\makebox(0,0){\rotatebox{90}{\scriptsize $t=0.015$}}}
\end{overpic}
\caption{
Fisher-KPP, trained to $T_{\mathrm{train}}=0.009$ and rolled out to $T=0.015$.
}
\label{fig:te_fkpp}
\end{figure}

\begin{figure}[htbp]
\centering
\begin{overpic}[width=0.85\linewidth]{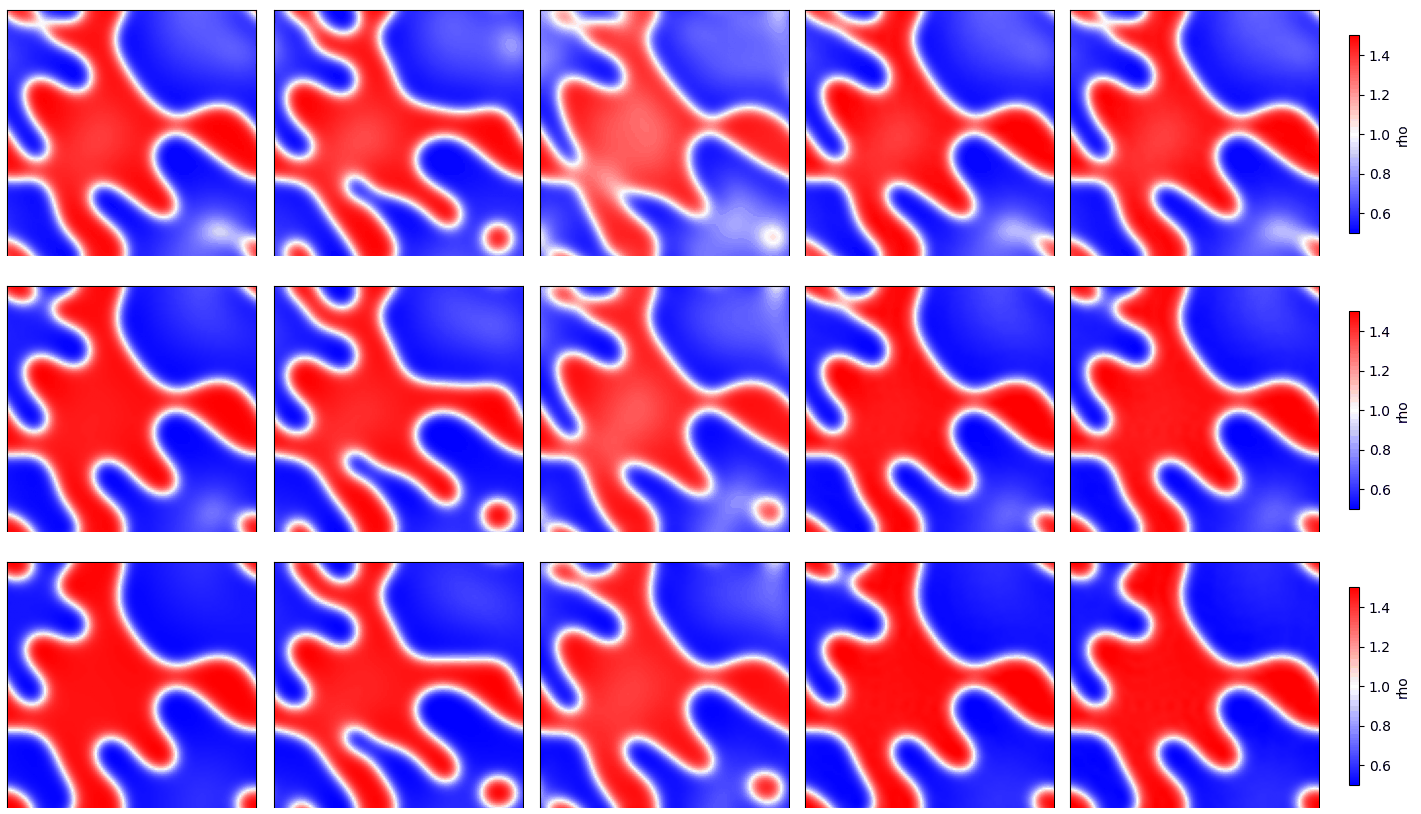}
        \put(10,58){\makebox(0,0){\scriptsize Reference}}
        \put(28,58){\makebox(0,0){\scriptsize F-FNO}}
        \put(47,58){\makebox(0,0){\scriptsize CFO}}
        \put(65,58){\makebox(0,0){\scriptsize NCL-MCT (Vel.)}}
        \put(84,58){\makebox(0,0){\scriptsize NCL-MCT (Law)}}
        \put(-1,50){\makebox(0,0){\rotatebox{90}{\scriptsize $t=0.06$}}}
        \put(-1,30){\makebox(0,0){\rotatebox{90}{\scriptsize $t=0.08$}}}
        \put(-1,10){\makebox(0,0){\rotatebox{90}{\scriptsize $t=0.10$}}}
\end{overpic}
\caption{
Reactive Cahn-Hilliard, trained to $T_{\mathrm{train}}=0.06$ and rolled out to
$T=0.1$.
}
\label{fig:te_rch}
\end{figure}
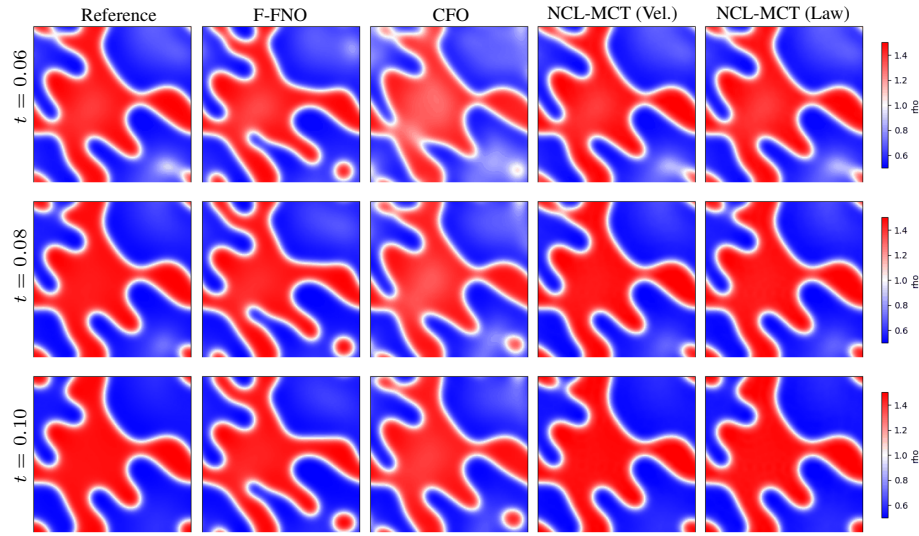
\clearpage

\section{Trajectory-Free Constitutive Learning}
\label{app:trajectory_free}

In \secref{exp_generality} and \secref{exp_generalization}, both supervision modes use
density fields drawn from precomputed solution trajectories.
Thus, even under known-law supervision, a numerical solver is still required to construct
the training set.
Here we remove this dependence.
Because the targets $\xi^{\mathrm{law}}$ and $\mathbf f^{\mathrm{law}}$ in
$\lossXi$ and $\lossF$ (\eqnref{method_constitutive_losses}) depend only on the current
density field, they can be evaluated on arbitrary positive fields without time integration.
We therefore train directly on independently sampled density fields.
We compare \NCLMCT (Law) with DOOL~\citep{chang2025dool} on the one-dimensional
linear diffusion and two-dimensional Cahn-Hilliard settings used in that work,
using the same independently sampled density fields for both methods.
Neither method uses solution trajectories, velocity fields, or time derivatives during
training; trajectories are reserved for held-out rollout evaluation.

\paragraph{One-dimensional linear diffusion (DOOL setting).}
We consider
\begin{equation}
\partial_t\rho=D\,\partial_{xx}\rho,
\qquad x\in[-\pi,\pi),\quad D=1,
\label{eq:app_tf_1d_diffusion}
\end{equation}
with periodic boundary conditions.
Both methods are trained on the same $50$ independently sampled density fields on a
$128$-point grid,
\begin{equation}
\rho(x)=2+c\sin x,
\qquad c\sim\mathcal U[0,1),
\label{eq:app_tf_single_mode}
\end{equation}
which are positive by construction.
\NCLMCT uses the factorization
$\xi=1/\rho$ and $f=-\partial_x\rho$, yielding
$u=-\partial_x\rho/\rho$ as in Appendix~\ref{app:settings_diffusion}.
This also illustrates the factorization freedom discussed in \secref{preliminaries}.
For this one-dimensional experiment, we replace the per-sample normalization of
$\lossError$ in \secref{constitutive_learning} with batch-level normalization:
the batch-summed squared error is divided by the batch-summed squared target norm,
with denominator stabilizer $10^{-12}$.
DOOL minimizes its Onsager objective on the same fields; its branch network receives the
first eight complex Fourier coefficients, including the constant mode, covering all
active modes in \eqnref{app_tf_single_mode}.
Both methods are trained for $50{,}000$ Adam updates with learning rate $10^{-3}$ and
batch size $50$.
Evaluation uses ten held-out initial conditions from the same family: $c=1$ for the
visualized example and nine independent draws.
The reference solution is given analytically by Fourier-mode decay.
Both methods roll out to $T=1$ with $\Delta t=2.5\times10^{-4}$, corresponding to
$N_T=4{,}000$ steps, with errors measured at $101$ frames.
\NCLMCT advances the one-dimensional MCT factors $I$ and $M$, whereas DOOL advances
its predicted flux using spectral differentiation and second-order Runge-Kutta.

\paragraph{Two-dimensional Cahn-Hilliard (DOOL setting).}
We reproduce Example 3.5 of \citet{chang2025dool},
\begin{equation}
\partial_t u
=
\Delta\bigl[-\gamma_1\Delta u+\gamma_2(u^3-u)\bigr],
\qquad
\mathbf x\in(0,2\pi)^2,\quad
\gamma_1=\gamma_2=1,
\label{eq:app_tf_ch2d}
\end{equation}
with periodic boundary conditions.
Because \NCLMCT uses the mobility factorization $\xi=1/\rho$, it requires a positive
density, whereas $u$ changes sign.
We therefore represent each state as $\rho=u+2$ and evaluate the constitutive law through
$u=\rho-2$.
This constant shift preserves the underlying dynamics while ensuring positivity;
$\rho$ is clamped to $[0.7,3.3]$.
The training set contains $1{,}000$ independently sampled density fields on a
$128\times128$ grid from the $K=1$ truncated Fourier coefficient box of
\citet{chang2025dool}.
We use $1{,}000$ rather than its $N_b=100$ states so that the sample budget matches the
other two-dimensional experiments in this paper.
Both methods are trained for $100{,}000$ updates with batch size $32$.
Evaluation uses ten held-out initial conditions from the same band-limited family.
The first is $u_0=\sin x\sin y$, following \citet{chang2025dool}, and the remaining nine
are independent draws.
Reference trajectories are computed with ETDRK4 at $\Delta t=10^{-5}$ and stored every
$10^{-4}$ up to $T=0.2$.
Both methods roll out with $\Delta t=10^{-4}$, corresponding to $N_T=2{,}000$ steps,
with errors measured at $101$ frames.

\begin{figure}[htbp]
\centering
\includegraphics[width=0.7\linewidth]{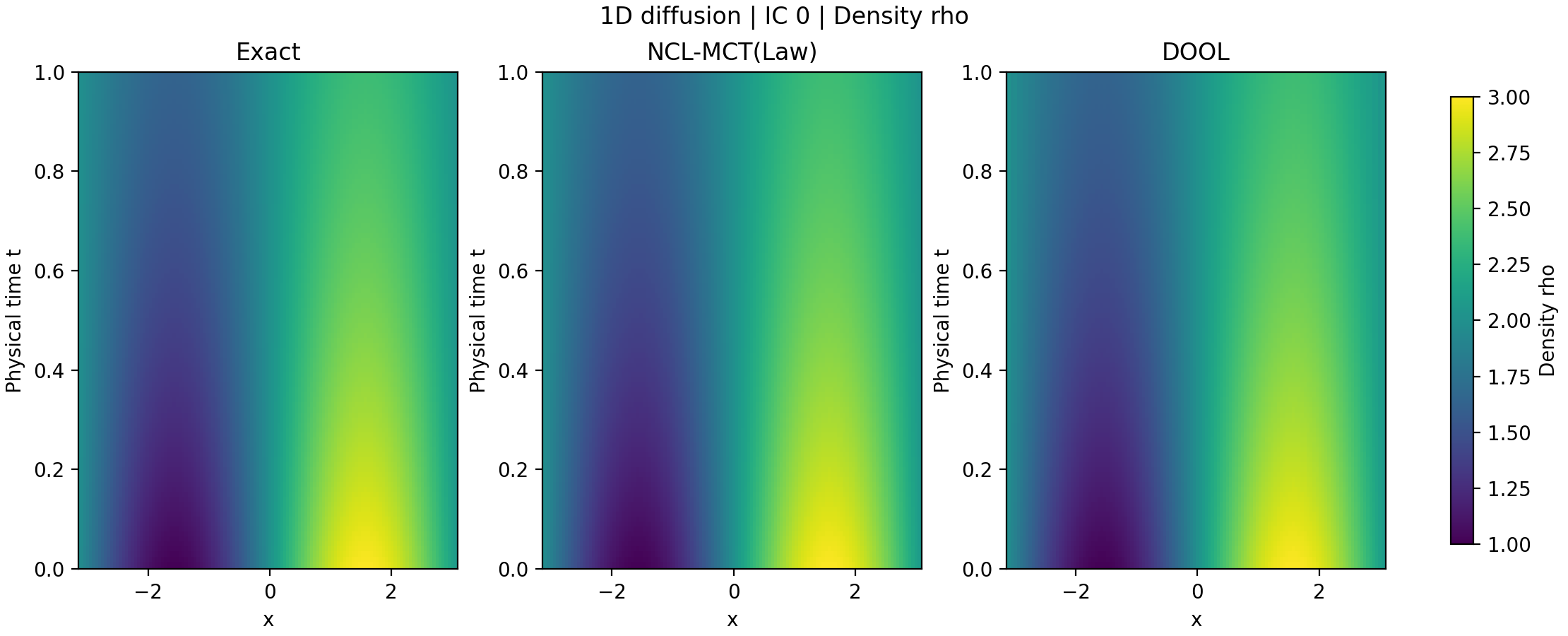}
\caption{
One-dimensional linear diffusion in the DOOL setting after trajectory-free constitutive
learning.
Space-time density for the visualized initial condition
$\rho_0(x)=2+\sin x$, rolled out to $T=1$ in $4{,}000$ steps.
Columns show the analytic solution, \NCLMCT (Law), and DOOL using a shared color scale.
}
\label{fig:tf_diffusion}
\end{figure}

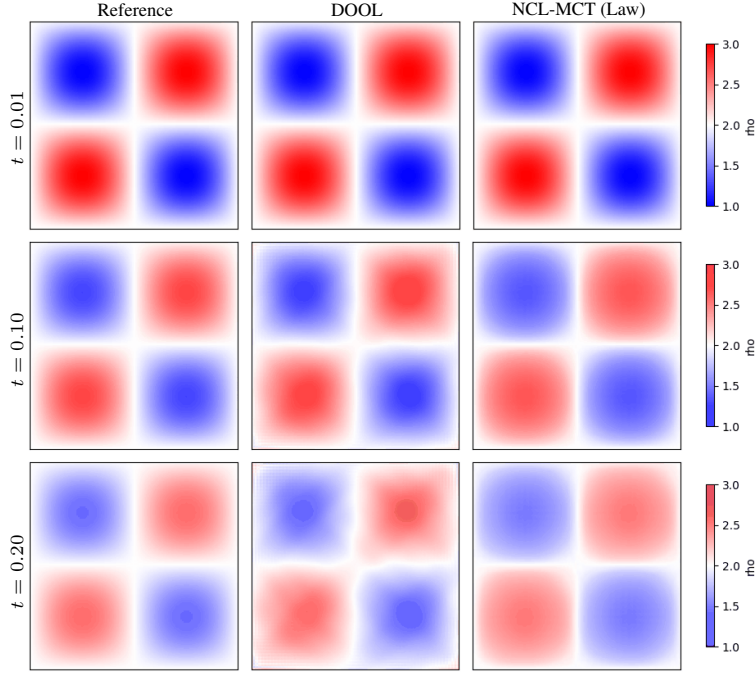
\begin{figure}[htbp]
\centering
\begin{overpic}[width=0.7\linewidth]{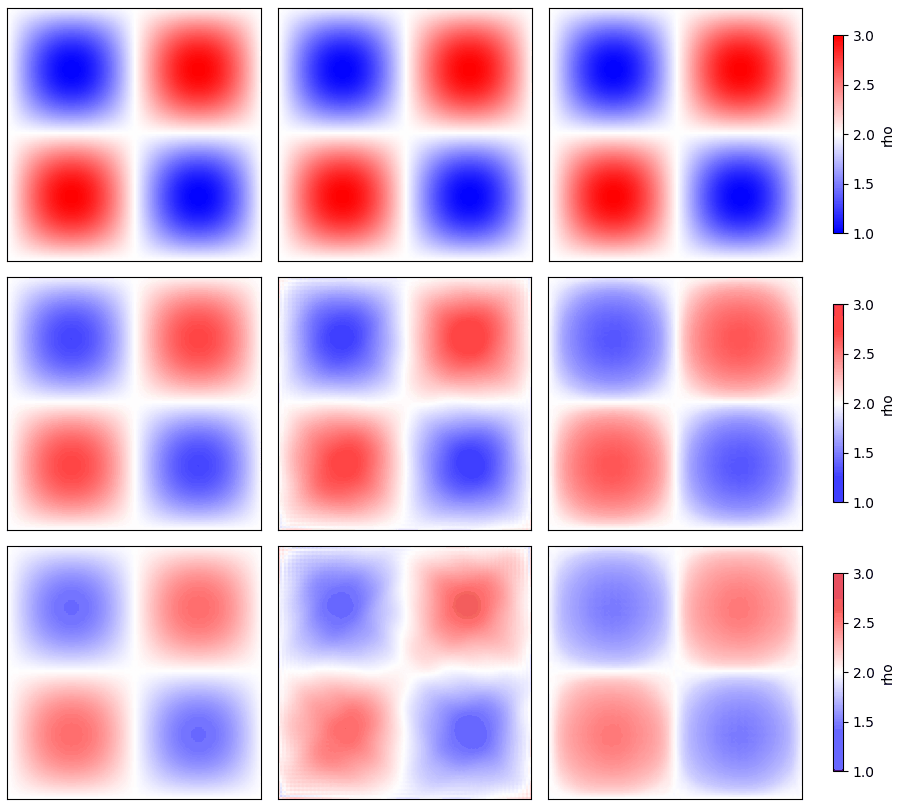}
  \put(15,90){\makebox(0,0){\scriptsize Reference}}
  \put(45,90){\makebox(0,0){\scriptsize DOOL}}
  \put(75,90){\makebox(0,0){\scriptsize NCL-MCT (Law)}}
  \put(-1,74){\makebox(0,0){\rotatebox{90}{\scriptsize $t=0.01$}}}
  \put(-1,44){\makebox(0,0){\rotatebox{90}{\scriptsize $t=0.10$}}}
  \put(-1,14){\makebox(0,0){\rotatebox{90}{\scriptsize $t=0.20$}}}
\end{overpic}
\caption{
Two-dimensional Cahn-Hilliard in the DOOL setting after trajectory-free constitutive
learning, shown using the shifted density $\rho=u+2$ for $u_0=\sin x\sin y$.
The rollout extends to $T=0.2$ over $2{,}000$ steps.
Rows show $t=0.01$, $0.10$, and $0.20$, with an independent color scale for each row.
Both methods preserve the checkerboard structure, while at the final time DOOL exhibits
grid-aligned texture and a distorted interface; \NCLMCT (Law) remains smooth with
slightly weaker interior contrast.
}
\label{fig:tf_ch2d}
\end{figure}

\paragraph{Results.}
Table~\ref{tab:trajectory_free} reports errors over the ten held-out initial conditions.
On one-dimensional linear diffusion, \NCLMCT (Law) achieves
$E_{\mathrm{roll}}=3.13\times10^{-4}$ and
$E_{\mathrm{max}}=4.05\times10^{-4}$,
respectively $4.9\times$ and $5.9\times$ lower than DOOL.
Figure~\ref{fig:tf_diffusion} shows the corresponding space-time density for the
visualized initial condition.
Both methods reproduce the single-mode decay, with the main visible discrepancy at
early times.
On two-dimensional Cahn-Hilliard, \NCLMCT (Law) also attains lower mean errors,
with $E_{\mathrm{roll}}=6.40\times10^{-3}$ and
$E_{\mathrm{max}}=8.71\times10^{-3}$, compared with
$7.46\times10^{-3}$ and $1.17\times10^{-2}$ for DOOL.
The standard deviation of \NCLMCT is larger, however, and the two methods remain within
one standard deviation; variation across initial conditions is therefore larger than the
difference in their mean errors.
Figure~\ref{fig:tf_ch2d} shows one representative rollout.
Both methods preserve the checkerboard structure over all $2{,}000$ steps, while at the
final time DOOL exhibits grid-aligned texture and a distorted interface, consistent with
the artifacts observed in Appendix~\ref{app:rollout_results}.
\NCLMCT remains smooth, with slightly reduced contrast in the domain interiors.
These experiments show that known-law supervision on independently sampled density
fields can train constitutive modules without solution trajectories and subsequently
support long-horizon rollout through the MCT integrator.
Because DOOL and \NCLMCT use different time integrators, the quantitative comparison
reflects the complete solvers rather than the training objectives alone.

\clearpage
\section{Additional Experimental Results}
\label{app:additional_experiments}

%%%%%%%%%%%%%%%%%%%%%%%%%%%%%%%%%%%%%%%%%%%%%%%%%%%%%%%%%%%
\subsection{Training Time and Inference Time}
\label{app:timing}

\begin{table}[b!]
    \centering
    \footnotesize
    \setlength{\tabcolsep}{5pt}
    \renewcommand{\arraystretch}{1.1}
    \caption{
        Computational cost of all methods.
        Params counts trainable parameters; peak VRAM is the framework-reported
        maximum live-tensor memory during training; Train is the wall-clock time for
        $100{,}000$ optimizer steps; Infer is the mean wall-clock time of one
        $N_T$-step rollout over the ten test trajectories; and ms/step is
        Infer$/N_T$.
        $-$ denotes unevaluated cases.
    }
    \label{tab:cost}

    \resizebox{\linewidth}{!}{%
    \begin{tabular}{@{}llcrrrrr@{}}
        \toprule
        System & Method & $N_T$ & Params & VRAM (GiB) & Train (s) & Infer (s) & ms/step \\
        \midrule

        \multirow{5}{*}{Linear Diffusion} & F-FNO & \multirow{5}{*}{$600$}
            & $733{,}506$ & $8.371$ & $11{,}117$ & $0.80 \pm 0.00$ & $1.34$ \\
        & CFO & & $8{,}095{,}489$ & $6.970$ & $5{,}712$ & $4.91 \pm 0.01$ & $8.18$ \\
        & DOOL & & $64{,}620$ & $0.082$ & $132$ & $0.45 \pm 0.04$ & $0.75$ \\
        & \NCLMCT (Vel.) & & $97{,}419$ & $1.061$ & $1{,}912$ & $1.22 \pm 0.00$ & $2.03$ \\
        & \NCLMCT (Law) & & $97{,}419$ & $1.049$ & $1{,}866$ & $1.22 \pm 0.00$ & $2.03$ \\
        \midrule

        \multirow{5}{*}{CH} & F-FNO & \multirow{5}{*}{$10{,}000$}
            & $733{,}506$ & $8.371$ & $11{,}117$ & $13.15 \pm 0.01$ & $1.32$ \\
        & CFO & & $8{,}095{,}489$ & $6.970$ & $5{,}716$ & $79.19 \pm 0.01$ & $7.92$ \\
        & DOOL & & $64{,}620$ & $0.082$ & $139$ & $6.25 \pm 0.01$ & $0.63$ \\
        & \NCLMCT (Vel.) & & $97{,}419$ & $1.061$ & $1{,}920$ & $19.43 \pm 0.05$ & $1.94$ \\
        & \NCLMCT (Law) & & $97{,}419$ & $1.050$ & $1{,}872$ & $19.47 \pm 0.05$ & $1.95$ \\
        \midrule

        \multirow{5}{*}{PM} & F-FNO & \multirow{5}{*}{$2{,}000$}
            & $733{,}506$ & $8.371$ & $11{,}119$ & $2.63 \pm 0.00$ & $1.31$ \\
        & CFO & & $8{,}095{,}489$ & $6.970$ & $5{,}708$ & $15.92 \pm 0.03$ & $7.96$ \\
        & DOOL & & $64{,}620$ & $0.082$ & $150$ & $1.96 \pm 0.00$ & $0.98$ \\
        & \NCLMCT (Vel.) & & $97{,}419$ & $1.062$ & $1{,}864$ & $3.98 \pm 0.01$ & $1.99$ \\
        & \NCLMCT (Law) & & $97{,}419$ & $1.050$ & $1{,}864$ & $3.96 \pm 0.01$ & $1.98$ \\
        \midrule

        \multirow{4}{*}{Fisher-KPP} & F-FNO & \multirow{4}{*}{$500$}
            & $733{,}506$ & $8.371$ & $11{,}118$ & $0.67 \pm 0.00$ & $1.33$ \\
        & CFO & & $8{,}095{,}489$ & $6.970$ & $6{,}786$ & $4.07 \pm 0.01$ & $8.15$ \\
        & \NCLMCT (Vel.) & & $98{,}572$ & $1.187$ & $2{,}012$ & $1.25 \pm 0.00$ & $2.50$ \\
        & \NCLMCT (Law) & & $98{,}572$ & $1.177$ & $2{,}041$ & $1.25 \pm 0.00$ & $2.50$ \\
        \midrule

        \multirow{4}{*}{Reactive CH} & F-FNO & \multirow{4}{*}{$10{,}000$}
            & $733{,}506$ & $8.371$ & $11{,}146$ & $13.21 \pm 0.02$ & $1.32$ \\
        & CFO & & $8{,}095{,}489$ & $6.970$ & $7{,}245$ & $79.23 \pm 0.02$ & $7.92$ \\
        & \NCLMCT (Vel.) & & $98{,}572$ & $1.187$ & $2{,}068$ & $28.03 \pm 0.05$ & $2.80$ \\
        & \NCLMCT (Law) & & $98{,}572$ & $1.177$ & $2{,}050$ & $28.17 \pm 0.02$ & $2.82$ \\
        \midrule

        \multirow{4}{*}{Reactive PM} & F-FNO & \multirow{4}{*}{$1{,}000$}
            & $733{,}506$ & $8.371$ & $11{,}111$ & $1.35 \pm 0.00$ & $1.35$ \\
        & CFO & & $8{,}095{,}489$ & $6.970$ & $8{,}281$ & $8.02 \pm 0.03$ & $8.02$ \\
        & \NCLMCT (Vel.) & & $98{,}572$ & $1.187$ & $2{,}028$ & $2.46 \pm 0.00$ & $2.46$ \\
        & \NCLMCT (Law) & & $98{,}572$ & $1.177$ & $2{,}032$ & $2.46 \pm 0.00$ & $2.46$ \\
        \midrule

        \multirow{4}{*}{Schnakenberg} & F-FNO & \multirow{4}{*}{$20{,}000$}
            & $733{,}700$ & $4.244$ & $4{,}877$ & $26.34 \pm 0.04$ & $1.32$ \\
        & CFO & & $8{,}096{,}130$ & $5.668$ & $6{,}158$ & $158.41 \pm 0.01$ & $7.92$ \\
        & \NCLMCT (Vel.) & & $194{,}520$ & $0.847$ & $1{,}656$ & $82.03 \pm 0.07$ & $4.10$ \\
        & \NCLMCT (Law) & & $194{,}520$ & $0.843$ & $1{,}618$ & $82.00 \pm 0.09$ & $4.10$ \\
        \bottomrule
    \end{tabular}%
    }
\end{table}

\paragraph{Experimental settings.}
All methods are trained for $100{,}000$ optimizer steps on the same $1{,}000$
snapshots per system and evaluated on the same ten held-out trajectories using a
single NVIDIA GeForce RTX~5090 (32~GB) and an AMD Ryzen~9 9950X3D CPU.
Peak VRAM is the framework-reported maximum live-tensor memory during training
(\texttt{torch.cuda.max\_memory\_allocated}; JAX device-memory peak for CFO),
excluding the CUDA context and allocator cache.
Inference time is averaged over the ten test trajectories after one warm-up rollout.
Because $N_T$ varies substantially across systems, we also report the normalized
per-step cost.

\paragraph{Results.}
Table~\ref{tab:cost} summarizes the measurements.
\NCLMCT uses several times fewer parameters than F-FNO and over an order of magnitude
fewer than CFO, while also requiring less training memory and wall-clock time.
This reflects the division of labor in \NCLMCT: the MCT integrator handles temporal
evolution, while the constitutive modules represent only PDE-specific responses.
Inference cost reflects the number of constitutive evaluations.
The per-step cost is lowest for nonreactive scalar systems, increases when a reaction
module is added, and is highest for Schnakenberg, which evaluates two transport
operators.
\NCLMCT remains faster than CFO on every system and at most about three times slower
than F-FNO, whose inference consists only of repeated state-map evaluations.
Vel.\ and Law have nearly identical inference costs and similar training costs because
they share the same architecture and differ only in supervision.
DOOL is the least expensive method, but its rollout errors are substantially larger on
the evaluated systems (Table~\ref{tab:reaction_diffusion}).
The transport-only DOOL formulation is evaluated on the three nonreactive systems;
reactive cases require additional problem-specific Onsager/Rayleighian design.
All models use batch size 16 for Schnakenberg
(Appendix~\ref{app:data_and_training}), explaining the lower training time and memory
of \NCLMCT and F-FNO there despite their larger multispecies models.

%%%%%%%%%%%%%%%%%%%%%%%%%%%%%%%%%%%%%%%%%%%%%%%%%%%%%%%%%%%
\subsection{Evaluation of Mass Behavior}
\label{app:physical_diagnostics}

We evaluate mass behavior on the porous-medium (PM) system to examine the effect of
retaining explicit mass-compression-transport kinematics while learning PDE-specific
constitutive responses.

\paragraph{Experimental settings.}
Relative mass error is measured against the total mass at the initial time.
The reference curve reports the corresponding drift of the reference evaluation.

\begin{figure}[ht]
    \centering
    \includegraphics[width=0.6\linewidth]{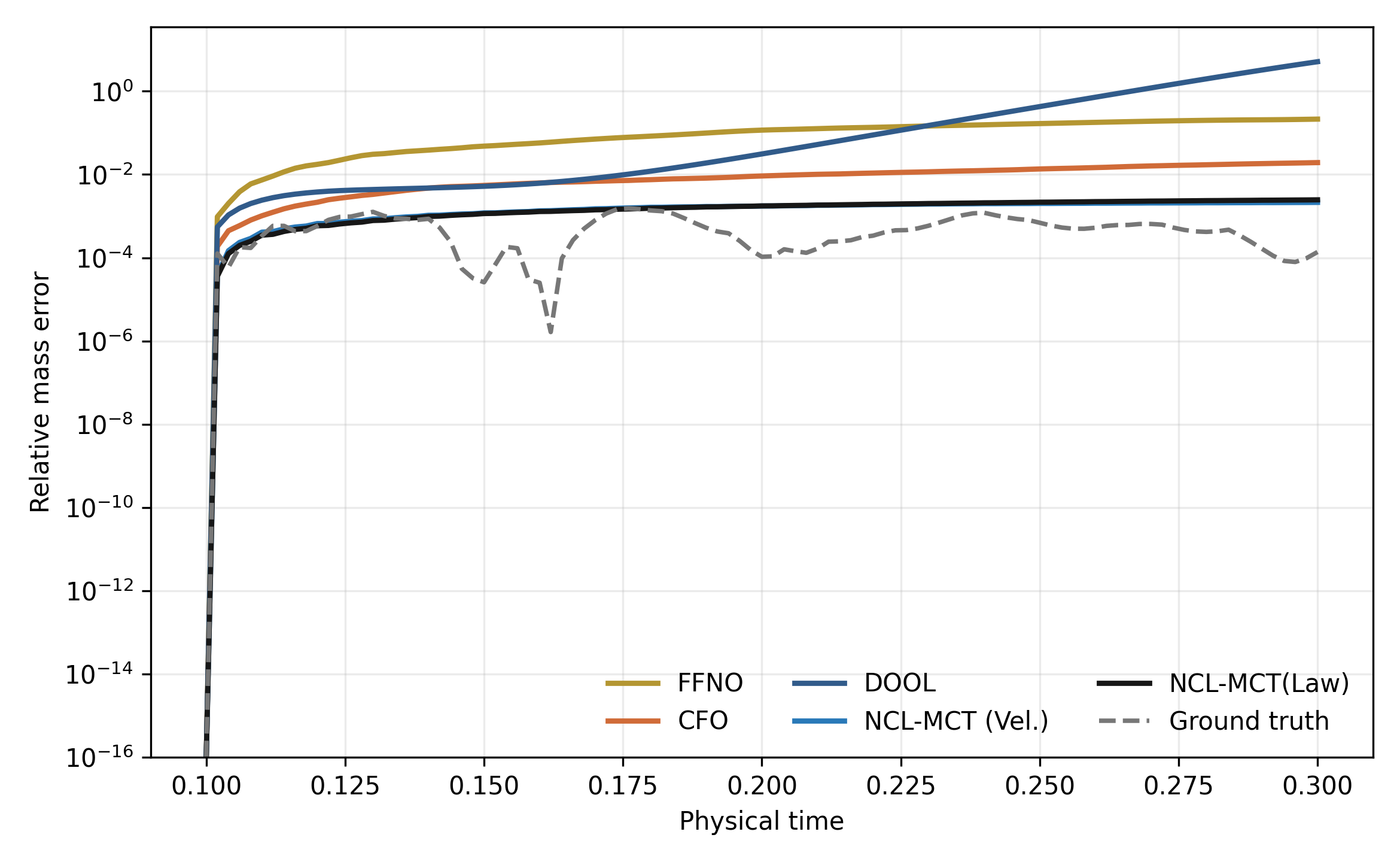}
    \caption{
        Relative mass error for porous medium, measured against the total mass at the
        initial time.
        The dashed curve shows the corresponding drift of the reference evaluation.
    }
    \label{fig:physical_diagnostics}
\end{figure}

\paragraph{Results.}
As shown in Figure~\ref{fig:physical_diagnostics}, both \NCLMCT (Vel.) and
\NCLMCT (Law) exhibit smaller relative mass drift than CFO and F-FNO, despite
having higher density rollout errors than CFO
(Table~\ref{tab:reaction_diffusion}).
This shows that lower density error does not necessarily imply better preservation of
total mass, while \NCLMCT retains favorable mass behavior over long rollouts.
This empirical behavior does not imply exact discrete conservation.
As discussed in \secref{method_coupling}, conservative evolution of the compression
factor $I$ alone does not guarantee conservation of the reconstructed density
$\rho=MI$.

%%%%%%%%%%%%%%%%%%%%%%%%%%%%%%%%%%%%%%%%%%%%%%%%%%%%%%%%%%%
\begin{table}[t!]
    \centering
    \footnotesize
    \setlength{\tabcolsep}{6pt}
    \renewcommand{\arraystretch}{1.15}
    \caption{
        Single-trajectory comparison with HC-PINN.
        HC-PINN and both \NCLMCT configurations are trained and evaluated on the same
        initial-value problem; the \NCLMCT results therefore differ from the
        ten-trajectory averages in Table~\ref{tab:reaction_diffusion}.
        Red/orange mark the best/second-best result per row.
    }
    \label{tab:hc_pinn_comparison}

    \resizebox{0.8\linewidth}{!}{%
    \begin{tabular}{@{}llccc@{}}
        \toprule
        System & Metric
        & \shortstack{HC-PINN\\\citep{hao2024stability}}
        & \shortstack{NCL-MCT (Vel.)\\(Ours)}
        & \shortstack{NCL-MCT (Law)\\(Ours)}\\
        \midrule

        \multicolumn{5}{l}{\textbf{(a) Generalized diffusion}} \\
        \midrule

        \multirow{2}{*}{Linear Diffusion}
        & $E_{\mathrm{roll}}$
        & \cmf $6.43\times10^{-4}$
        & \cms $2.17\times10^{-3}$
        & $2.23\times10^{-3}$ \\
        & $E_{\mathrm{max}}$
        & \cmf $1.55\times10^{-3}$
        & \cms $3.78\times10^{-3}$
        & $4.18\times10^{-3}$ \\
        \midrule

        \multirow{2}{*}{CH}
        & $E_{\mathrm{roll}}$
        & $2.44\times10^{-1}$
        & \cms $3.75\times10^{-2}$
        & \cmf $6.89\times10^{-3}$ \\
        & $E_{\mathrm{max}}$
        & $3.02\times10^{-1}$
        & \cms $4.93\times10^{-2}$
        & \cmf $1.11\times10^{-2}$ \\
        \midrule

        \multirow{2}{*}{PM}
        & $E_{\mathrm{roll}}$
        & $1.04\times10^{-1}$
        & \cmf $1.37\times10^{-2}$
        & \cms $1.38\times10^{-2}$ \\
        & $E_{\mathrm{max}}$
        & $1.52\times10^{-1}$
        & \cmf $1.76\times10^{-2}$
        & \cms $1.78\times10^{-2}$ \\
        \midrule

        \multicolumn{5}{l}{\textbf{(b) Generalized reaction-diffusion}} \\
        \midrule

        \multirow{2}{*}{Fisher-KPP}
        & $E_{\mathrm{roll}}$
        & \cmf $5.67\times10^{-4}$
        & \cms $1.99\times10^{-3}$
        & $2.09\times10^{-3}$ \\
        & $E_{\mathrm{max}}$
        & \cmf $1.73\times10^{-3}$
        & \cms $2.57\times10^{-3}$
        & $6.41\times10^{-3}$ \\
        \midrule

        \multirow{2}{*}{Reactive CH}
        & $E_{\mathrm{roll}}$
        & $2.56\times10^{-1}$
        & \cmf $1.37\times10^{-2}$
        & \cms $1.65\times10^{-2}$ \\
        & $E_{\mathrm{max}}$
        & $3.16\times10^{-1}$
        & \cmf $2.98\times10^{-2}$
        & \cms $3.50\times10^{-2}$ \\
        \midrule

        \multirow{2}{*}{Reactive PM}
        & $E_{\mathrm{roll}}$
        & $5.14\times10^{-2}$
        & \cms $2.44\times10^{-2}$
        & \cmf $1.42\times10^{-2}$ \\
        & $E_{\mathrm{max}}$
        & $8.25\times10^{-2}$
        & \cms $3.12\times10^{-2}$
        & \cmf $1.79\times10^{-2}$ \\
        \midrule

        \multicolumn{5}{l}{\textbf{(c) Multispecies reaction-diffusion}} \\
        \midrule

        \multirow{2}{*}{Schnakenberg ($U$)}
        & $E_{\mathrm{roll}}$
        & $2.49\times10^{-1}$
        & \cmf $3.30\times10^{-2}$
        & \cms $6.39\times10^{-2}$ \\
        & $E_{\mathrm{max}}$
        & $5.03\times10^{-1}$
        & \cmf $9.23\times10^{-2}$
        & \cms $1.74\times10^{-1}$ \\
        \addlinespace[3pt]

        \multirow{2}{*}{Schnakenberg ($V$)}
        & $E_{\mathrm{roll}}$
        & $8.17\times10^{-2}$
        & \cmf $1.01\times10^{-2}$
        & \cms $1.69\times10^{-2}$ \\
        & $E_{\mathrm{max}}$
        & $1.90\times10^{-1}$
        & \cmf $3.06\times10^{-2}$
        & \cms $4.82\times10^{-2}$ \\
        \bottomrule
    \end{tabular}%
    }
\end{table}

\subsection{Comparison with HC-PINN}
\label{app:hc_pinn_comparison}

We compare \NCLMCT with HC-PINN~\citep{hao2024stability} as two approaches to incorporating
physical structure: physics-constrained solution learning and constitutive learning
with a shared MCT integrator.
The comparison covers all seven systems spanning generalized diffusion,
generalized reaction-diffusion, and multispecies reaction-diffusion.

\paragraph{Experimental settings.}
HC-PINN fits one network to one initial-value problem, with the initial condition
embedded in the ansatz.
The resulting network represents that trajectory and must be retrained when the
initial condition changes.
\NCLMCT instead learns the constitutive responses of each PDE and subsequently
integrates different initial conditions using the shared MCT integrator.
For this comparison, we fit one HC-PINN per system to the first of the ten test initial
conditions in Appendix~\ref{app:data_and_training}, and retrain \NCLMCT using only
data from the same initial-value problem.
The reported values are therefore single-trajectory errors and differ from the
ten-trajectory averages in Table~\ref{tab:reaction_diffusion}.

We follow the architecture and hard-constraint formulation of \citet{hao2024stability}:
a $\tanh$ multilayer perceptron with seven hidden layers of width 32, applied to a
Fourier-feature embedding of the spatial coordinates truncated at 16 modes, so
periodicity is enforced structurally through the network input rather than a boundary
loss.
The prediction
$\rho=\psi+\varphi\,\rho_{\mathrm{nn}}$, with
$\psi=e^{-Ct}\rho_0$ and $\varphi=1-e^{-Ct}$,
satisfies the initial condition exactly, where $C>0$ controls the hard-constraint
ansatz.
Self-adaptive weighting is disabled.
Each optimizer step evaluates the PDE residual at $10^4$ collocation points sampled
uniformly from the same space-time domain.
Optimization uses $50{,}000$ Adam steps with learning rate $10^{-3}$, followed by
$25{,}000$ L-BFGS steps with a strong-Wolfe line search.
Because the source code of \citet{hao2024stability} is not publicly available, we reimplemented
the method from its description.
The original configuration uses $5{,}000$ L-BFGS steps after $50{,}000$ Adam steps;
we extend the L-BFGS phase to $25{,}000$ steps, by which point the loss has plateaued
and the resulting errors match or improve on the reported values.

\paragraph{Results.}
On linear diffusion and Fisher-KPP, HC-PINN is more accurate on both metrics,
consistent with the pattern in \secref{exp_generalized_diffusion}: when a
trajectory-specific model represents the solution accurately, direct solution learning
avoids repeated constitutive evaluation and numerical integration.
On the remaining five systems, both \NCLMCT configurations achieve lower errors on
both metrics, by up to $35\times$ on the two fourth-order systems and up to
$8.6\times$ on the degenerate-transport and multispecies systems, even though HC-PINN
is fitted directly to the initial-value problem on which it is evaluated.
These results support constitutive learning with a shared MCT integrator on the more
complex systems considered here, although differences in architecture, supervision,
and optimization prevent attributing the gap to any single design choice.

%%%%%%%%%%%%%%%%%%%%%%%%%%%%%%%%%%%%%%%%%%%%%%%%%%%%%%%%%%%
\clearpage
\subsection{Ablation of Constitutive Parameterization and Supervision}
\label{app:exp_output_ablation}

We compare three transport-learning configurations on Schnakenberg
(Table~\ref{tab:output_ablation}):
direct velocity prediction trained with $\lossU+\lossR$, and mobility-force
prediction under either velocity-data supervision with curl regularization
($\lossVel$, Vel.) or known-law supervision of $\xi$ and $\mathbf f$
($\lossMF$, Law).
All three use the same reaction loss $\lossR$, isolating differences in transport
parameterization and supervision.
The errors have comparable magnitude and the same ordering across both metrics and
species: Law gives the lowest mean errors, followed by Vel.\ and direct velocity
prediction.
Both mobility-force configurations therefore retain an explicit decomposition into
positive mobility and thermodynamic driving force while achieving lower mean errors
than direct velocity prediction on this system.
This comparison does not isolate the effects of factorization and curl regularization;
Appendix~\ref{app:lambda} separately examines the effect of $\lambda_{\mathrm{curl}}$
on Fisher-KPP.

% Required packages: booktabs, graphicx
% Uses the existing \cmf and \cms macros.
\providecommand{\errstat}[3]{%
    \ensuremath{(#1\pm#2)\times10^{#3}}%
}

\begin{table}[ht]
    \centering
    \small
    \setlength{\tabcolsep}{4pt}
    \renewcommand{\arraystretch}{1.15}
    \caption{
        Ablation of constitutive parameterization and supervision on Schnakenberg.
        Errors are mean $\pm$ standard deviation.
        Red/orange indicate the best/second-best unrounded mean in each column.
    }
    \label{tab:output_ablation}

    \resizebox{\linewidth}{!}{%
    \begin{tabular}{@{}llcccc@{}}
        \toprule
        & & \multicolumn{2}{c}{Schnakenberg ($U$)}
          & \multicolumn{2}{c}{Schnakenberg ($V$)} \\
        \cmidrule(lr){3-4}
        \cmidrule(lr){5-6}
        Parameterization & Objective
        & $E_{\mathrm{roll}}$ & $E_{\mathrm{max}}$
        & $E_{\mathrm{roll}}$ & $E_{\mathrm{max}}$ \\
        \midrule

        $\mathbf u,\mathbf r$
        & $\lossU+\lossR$
        & \errstat{3.40}{1.70}{-2}
        & \errstat{1.11}{0.66}{-1}
        & \errstat{7.84}{4.27}{-3}
        & \errstat{2.96}{2.18}{-2} \\
        \addlinespace[3pt]

        $(\xi,\mathbf f),\mathbf r$
        & $\lossVel$
        & \cms \errstat{2.52}{1.77}{-2}
        & \cms \errstat{8.28}{8.55}{-2}
        & \cms \errstat{6.74}{3.37}{-3}
        & \cms \errstat{2.19}{1.93}{-2} \\
        \addlinespace[3pt]

        $(\xi,\mathbf f),\mathbf r$
        & $\lossMF$
        & \cmf \errstat{2.38}{1.26}{-2}
        & \cmf \errstat{7.36}{6.15}{-2}
        & \cmf \errstat{6.57}{2.27}{-3}
        & \cmf \errstat{1.97}{1.30}{-2} \\
        \bottomrule
    \end{tabular}%
    }
\end{table}

%%%%%%%%%%%%%%%%%%%%%%%%%%%%%%%%%%%%%%%%%%%%%%%%%%%%%%%%%%%
\subsection{Ablation of Constitutive Modularity}
\label{app:constitutive_modularity}

This experiment examines whether numerically evaluated and learned constitutive
responses can be combined within the same MCT integrator.
Here, Numerical denotes responses computed directly from known constitutive laws using
the corresponding numerical discretization, while Neural denotes responses predicted
by trained constitutive modules.
We evaluate four transport-reaction combinations on Schnakenberg:
Numerical-Numerical, Neural-Numerical, Numerical-Neural, and Neural-Neural.
The Numerical-Numerical configuration serves as the numerical constitutive reference
within the same MCT integrator.

\paragraph{Experimental settings.}
All configurations use the same MCT factor evolution, test initial conditions,
$128\times128$ grid, and $20{,}000$ integration steps to $T=1$.
Neural constitutive modules are trained separately for each configuration under
known-law supervision.
Thus, this experiment evaluates module replacement rather than reuse of identical
trained modules across configurations.
Errors follow the $E_{\mathrm{roll}}$ and $E_{\mathrm{max}}$ definitions in
Section~\ref{sec:result} and are measured against the reference solution.

\paragraph{Results.}
Table~\ref{tab:module_ablation} shows that Numerical-Numerical achieves the lowest
errors for both species.
Among the mixed configurations, Neural-Numerical yields lower rollout errors than
Numerical-Neural, indicating that replacing the reaction response with a learned
approximation introduces the larger error in this experiment.
The fully learned Neural-Neural configuration achieves mean rollout errors of
$2.38\times10^{-2}$ for $U$ and $6.57\times10^{-3}$ for $V$.
All four configurations use the same MCT integrator, demonstrating that numerically
evaluated and learned transport and reaction responses can share the same constitutive
interface.

\begin{table}[t]
    \centering
    \footnotesize
    \setlength{\tabcolsep}{3pt}
    \renewcommand{\arraystretch}{1.15}
    \caption{
        Ablation of constitutive modularity on Schnakenberg.
        Numerical denotes constitutive responses computed from known laws using
        numerical discretization; Neural denotes responses predicted by trained
        constitutive modules.
        Errors are mean $\pm$ standard deviation.
        Red/orange mark the best/second-best means for each species.
        Numerical-Numerical is the numerical constitutive reference within the same
        MCT integrator.
    }
    \label{tab:module_ablation}

    \resizebox{0.7\linewidth}{!}{%
    \begin{tabular}{@{}lllcc@{}}
        \toprule
        Transport & Reaction & Species
        & $E_{\mathrm{roll}}$
        & $E_{\mathrm{max}}$ \\
        \midrule

        \multirow{2}{*}{Numerical}
        & \multirow{2}{*}{Numerical}
        & $U$
        & \cmf \errstat{6.57}{2.43}{-3}
        & \cmf \errstat{1.634}{0.480}{-2} \\
        & & $V$
        & \cmf \errstat{2.11}{0.80}{-3}
        & \cmf \errstat{6.25}{1.95}{-3} \\
        \midrule

        \multirow{2}{*}{Neural}
        & \multirow{2}{*}{Numerical}
        & $U$
        & \cms \errstat{1.059}{0.259}{-2}
        & \cms \errstat{2.361}{0.412}{-2} \\
        & & $V$
        & \cms \errstat{3.07}{0.58}{-3}
        & \cms \errstat{7.52}{1.35}{-3} \\
        \midrule

        \multirow{2}{*}{Numerical}
        & \multirow{2}{*}{Neural}
        & $U$
        & \errstat{2.191}{1.507}{-2}
        & \errstat{7.182}{6.851}{-2} \\
        & & $V$
        & \errstat{6.63}{3.40}{-3}
        & \errstat{2.178}{1.809}{-2} \\
        \midrule

        \multirow{2}{*}{Neural}
        & \multirow{2}{*}{Neural}
        & $U$
        & \errstat{2.376}{1.263}{-2}
        & \errstat{7.362}{6.152}{-2} \\
        & & $V$
        & \errstat{6.57}{2.27}{-3}
        & \errstat{1.971}{1.299}{-2} \\
        \bottomrule
    \end{tabular}%
    }
\end{table}

%%%%%%%%%%%%%%%%%%%%%%%%%%%%%%%%%%%%%%%%%%%%%%%%%%%%%%%%%%%
\subsection{Ablation of the Curl Penalty Weight}
\label{app:lambda}

The curl penalty $\lossCurl$ in \eqnref{method_curl_loss} encourages the predicted
driving force to be curl-free, reflecting the EnVarA structure
$\mathbf f=-\nabla(\delta\mathcal E/\delta\rho)$ that velocity-data supervision alone
does not enforce.
We examine whether this regularization improves rollout accuracy and its sensitivity
to the weight $\lambdaCurl$.

\paragraph{Experimental settings.}
We vary $\lambdaCurl\in\{0,\,0.01,\,0.1,\,1\}$ on Fisher-KPP under
velocity-data supervision.
Setting $\lambdaCurl=0$ removes the curl penalty, leaving transport supervised only
through the reconstructed velocity
$\widehat{\mathbf u}=\widehat\xi\,\widehat{\mathbf f}$.
All other settings are fixed, including the training samples, network architecture
(Appendix~\ref{app:network_architectures}), optimization settings and random seed
(Appendix~\ref{app:data_and_training}), and the $500$-step rollout over ten test
trajectories (Appendix~\ref{app:settings_fkpp}).
The reaction loss $\lossR$ and unit weight on $\lossU$ are retained in all runs,
isolating the effect of $\lambdaCurl$.

\paragraph{Results.}
Table~\ref{tab:curl_ablation} shows the lowest errors at the intermediate weight
$\lambdaCurl=0.01$, with the same ordering on both metrics.
Removing the penalty approximately doubles the error relative to
$\lambdaCurl=0.01$
($2.1\times$ for $E_{\mathrm{roll}}$ and $2.2\times$ for $E_{\mathrm{max}}$),
indicating that curl regularization provides information not enforced by velocity-data
supervision alone.
Increasing the weight beyond $0.01$ reduces this benefit: at $\lambdaCurl=1$, both
errors are within one standard deviation of the $\lambdaCurl=0$ run.
This suggests that overly strong curl regularization can trade velocity accuracy for
force regularity.
We therefore use $\lambdaCurl=0.01$ throughout the paper; this setting corresponds to
the \NCLMCT (Vel.) result for Fisher-KPP in
Table~\ref{tab:reaction_diffusion}.
This ablation is conducted on a single system, and $\lambdaCurl$ is not retuned per
PDE.

\begin{table}[t]
    \centering
    \footnotesize
    \setlength{\tabcolsep}{3pt}
    \renewcommand{\arraystretch}{1.15}
    \caption{
        Ablation of the curl-penalty weight $\lambdaCurl$ on Fisher-KPP under
        velocity-data supervision.
        $\lambdaCurl=0$ removes the penalty.
        Errors are mean $\pm$ standard deviation over ten test trajectories with
        $500$ rollout steps.
        Red/orange mark the best/second-best means.
        $\lambdaCurl=0.01$ is used throughout the paper.
    }
    \label{tab:curl_ablation}

    \resizebox{0.45\linewidth}{!}{%
    \begin{tabular}{@{}lcc@{}}
        \toprule
        $\lambdaCurl$
        & $E_{\mathrm{roll}}$
        & $E_{\mathrm{max}}$ \\
        \midrule
        $0$
        & \errstat{1.03}{0.23}{-3}
        & \errstat{1.32}{0.36}{-3} \\
        $0.01$
        & \cmf \errstat{5.01}{1.39}{-4}
        & \cmf \errstat{5.96}{1.96}{-4} \\
        $0.1$
        & \cms \errstat{8.61}{1.59}{-4}
        & \cms \errstat{1.17}{0.26}{-3} \\
        $1$
        & \errstat{1.13}{0.19}{-3}
        & \errstat{1.72}{0.53}{-3} \\
        \bottomrule
    \end{tabular}%
    }
\end{table}

%\section{Disclosure of AI Use}
%\label{app:ai_disclosure}
%The authors developed the research ideas and proposed method, performed the mathematical verification, and designed and conducted all experiments. AI tools assisted with identifying relevant literature, exploring existing neural network architectures and numerical methods, assisting with code implementation and debugging, and improving the manuscript's wording and clarity. All code was reviewed and validated by the authors, who take full responsibility for the content and conclusions of this work.

\end{document}